\documentclass[jair,twoside,11pt,theapa]{article}
\usepackage{jair, rawfonts}
\usepackage{theapa}
\usepackage{amssymb} 

\usepackage{amsmath,amsfonts,bm}

\def\eqref#1{equation~\ref{#1}}

\def\1{\bm{1}}

\DeclareMathAlphabet{\mathsfit}{\encodingdefault}{\sfdefault}{m}{sl}
\SetMathAlphabet{\mathsfit}{bold}{\encodingdefault}{\sfdefault}{bx}{n}

\newcommand{\R}{\mathbb{R}}

\usepackage{amsmath}
\usepackage[utf8]{inputenc} 
\usepackage[T1]{fontenc}    
\usepackage{url}            
\usepackage{booktabs}       
\usepackage{amsfonts}       
\usepackage{nicefrac}       
\usepackage{microtype}      
\usepackage{xcolor}         
\usepackage{graphicx}
\usepackage{amsthm}
\newtheorem{theorem}{Theorem}
\newtheorem{corollary}{Corollary}[theorem]
\newtheorem{lemma}{Lemma}
\newtheorem{proposition}{Proposition}
\newtheorem*{assumption}{Assumption}
\usepackage{algorithm}
\usepackage{algorithmic}
\usepackage{cleveref}
\usepackage{todonotes}
\usepackage{mathtools}
\usepackage[inline]{enumitem}
\usepackage{bbm}
\usepackage{caption}
\usepackage{subcaption}
\usepackage{multirow}
\usepackage{thm-restate}
\usepackage{cancel}

\usepackage[linalg,rl]{abaisero}
\newcommand\realset{\mathbb{R}}
\DeclareMathOperator\Exp{\mathbb{E}}
\DeclareMathOperator{\Var}{\mathbb{V}ar}
\DeclareMathOperator{\Cov}{\mathbb{C}ov}

\newcommand\noti{{\setminus i}}
\newcommand\Hs{{\bm{H}}}
\newcommand\As{{\bm{A}}}

\newcommand\hs{{\bm{h}}}
\newcommand\as{{\bm{a}}}
\newcommand\os{{\bm{o}}}
\newcommand\policies{{\bm{\pi}}}
\newcommand\rpolicies{R^\policies}
\newcommand\vpolicies{V^\policies}
\newcommand\qpolicies{Q^\policies}
\newcommand\apolicies{A^\policies}
\newcommand\ppolicies{P^\policies}

\renewcommand\v[1][]{{V^{#1}}}
\newcommand\norm[1]{\left\|#1\right\|}
\newcommand\abs[1]{\left | #1 \right |}
\newcommand\ones{\mathbbm{1}}

\newcommand{\expect}[2]{ \Exp_{#1} \left[ #2 \right]}
\newcommand{\grad}{g}
\newcommand{\samplegrad}{\hat{\grad}}

\newcommand\nablai{\nabla_{\! \theta_i}}

\jairheading{77}{2023}{295-354}{11/2022}{05/2023}
\ShortHeadings{On Centralized Critics}
{Lyu, Baisero, Xiao, Daley, \& Amato}
\firstpageno{1}

\begin{document}
\setcounter{page}{1}
\title{On Centralized Critics in Multi-Agent Reinforcement Learning}

\author{\name Xueguang Lyu \email lu.xue@northeastern.edu \\
       \name Andrea Baisero \email baisero.a@northeastern.edu \\
       \name Yuchen Xiao \email xiao.yuch@northeastern.edu \\
       \name Brett Daley \email daley.br@northeastern.edu \\
       \name Christopher Amato \email c.amato@northeastern.edu \\
       \addr Northeastern University, Khoury College of Computer Sciences,\\
      360 Huntington Avenue, Boston, MA 02115 USA}


\maketitle

\begin{abstract}
\textit{Centralized Training for Decentralized Execution},
where agents are trained offline in a centralized fashion and execute online in a decentralized manner, has become a popular approach in Multi-Agent Reinforcement Learning (MARL).
In particular, it has become popular to develop actor-critic methods that train decentralized actors with a centralized critic where the centralized critic is allowed access global information of the entire system, including the true system state.
Such centralized critics are possible given offline information and are not used for online execution. 
While these methods perform well in a number of domains and have become a de facto standard in MARL, using a centralized critic in this context has yet to be sufficiently analyzed theoretically or empirically. 
In this paper, we therefore formally analyze centralized and decentralized critic approaches, and analyze the effect of using state-based critics in partially observable environments.
We derive theories contrary to the common intuition: critic centralization is not strictly beneficial, and using state values can be harmful.
We further prove that, in particular, state-based critics can introduce unexpected bias and variance compared to history-based critics. 
Finally, we demonstrate how the theory applies in practice by comparing different forms of critics on a wide range of common multi-agent benchmarks.
The experiments show practical issues such as the difficulty of representation learning with partial observability, which highlights why the theoretical problems are often overlooked in the literature.

\end{abstract}

\section{Introduction}
\label{Introduction}

Centralized Training for Decentralized Execution (CTDE)~\cite{oliehoek2008optimal}, where agents are trained offline in a centralized manner but execute in a decentralized manner with only local information, has been widely adopted in multi-agent reinforcement learning (MARL).
Compared to independent learning, CTDE has great potential for more stable and optimal learning since agents can coordinate offline on how they will behave online.
Actor-Critic (AC) methods are popular for CTDE because a centralized critic can be used to train decentralized actors, exploiting the centralized training paradigm; since the critic is only needed to train the actors, it can be discarded once the actors are fully trained without hindering decentralized execution. Because the centralized critic is trained offline in a simulator, it can be trained on the joint observations from all the agents as well as the system state.
Using the state is intuitively considered desirable as it is often more concise than the history and provides ground-truth information.
This technique of exploiting the system state has become popular after the pioneering centralized critic works of COMA~\cite{foerster2016learning} and MADDPG~\cite{lowe2017multi}.

The statements made for the effects of centralized critics are almost entirely positive.
For instance, MADDPG~\cite{lowe2017multi} notes that it eases learning and helps to learn coordinated behaviors.
Later works list similar sentiments, suspecting that critic centralization improves performance~\cite{lee2019improved}, reduces variance~\cite{TarMAC}, stabilizes training~\cite{M3DDPG} and is more robust~\cite{simoes2020multi}.
It seems reasonable to make these assumptions because training a centralized value function would solve the cooperation issues (e.g., action shadowing~\cite{claus1998dynamics}) for a centralized policy, and result in better convergence properties.
However, as we will show, a centralized critic does not have the same effect (in solving those problems) on a set of decentralized policies.

The exploitation of the system state is considered one of the significant advantages of the CTDE paradigm. Using the state has also become a selling point for centralized critics~\cite{foerster2016learning} since state value functions are usually easier to learn than observation-history value functions.
While many works use centralized critics, they often focus on other issues without carefully examining the foundation of the critic centralization techniques that they employ, e.g., improving credit assignment~\cite{SQDDPG,LIIR}, exploration~\cite{LICA}, emergent tool use~\cite{openAI}, etc.
However, even though centralized critics have become a standard mechanism in recent works, they still lack a thorough theoretical and empirical analysis.
In this paper, we give a comprehensive investigation of these claims and show that most gains with critic centralization are questionable, as they often entail hidden trade-offs, both theoretically and empirically.

This paper fills this analytical gap by providing an analysis of critics conditioned on joint observations as well as the system state; we show that the common intuitions stated in most recent works are usually unsound for both cases. In particular,
\begin{enumerate}[label=(\arabic*)]
\item we show that centralized critics are not theoretically beneficial compared to decentralized critics,
\item we show that state-based critics may result in bias, making them theoretically inferior to history-based critics,
\item we show that centralized critics (both history and state-based) result in higher variance, 
\item we advise the use of history-state critics, which use both history and state information as a method to incorporate state without introducing bias, and  
\item we provide an extensive empirical analysis showing the trade-offs of the different critic types. 
\end{enumerate}

In our analysis, we prove that critic centralization does not theoretically improve cooperation compared to decentralized critics from a policy learning perspective, even though the values themselves may be easier to learn.
For history-based critics, we show in theory that centralized and decentralized critics have the same expected gradient.
This implies that the centralized critic, like decentralized critics, can be used to train decentralized policies in an unbiased way.
Yet on the flip side, it also implies that the centralized critic cannot ameliorate the cooperation issues seen with decentralized learners.
For state-based critics, we show that when using only state information, with the commonly seen state-based centralized critic, they may incur unbounded bias in the policy gradient compared to their provably correct history-based counterparts.
The resulting bias voids any asymptotic convergence properties.
Practically, the bias may hinder learning a reasonable policy in many domains.
We detail formally and intuitively where the bias originates while analyzing its relationship to the environmental observation model, which is highly related to the bias.
Based on our theory and empirical studies, we suggest not using state-based critics in partially observable environments that require active information gathering.

We also compare the \emph{variance} of the policy gradient for different types of critics, where, again, the theoretical result counters common intuition: we show theoretically that the centralized critic adds higher variance to the policy gradient.
We also show that even when unbiased, using a state-conditioned critic further exacerbates the policy gradient variance issue.
We note that the stability of learning policies from a critic should be a consideration separate from the stability of learning value functions (the critics themselves).
We show that in practice, factoring in the effect of value function learning, we face a bias-variance trade-off. 
We also recommend the usage of a history-state-based critic~\cite{baisero2022unbiased}, although providing policy gradients with even higher variances in theory, they are unbiased and it is usually a favorable trade-off in terms of empirical performance.

Finally, through toy examples and larger empirical studies, we show that using centralized critics can be, in many tasks, harmful to the overall performance.
We compare decentralized history-based critics, centralized history-based critics, centralized state-based critics, and (centralized) history-state-based critics.
We also highlight the deficiencies of popular benchmarks, in that they often lack partial observability.
Our experiments consist of a wide range of popular benchmarks, but we do not find significant performance gaps on most tasks.
We also report performance degradation in environments preferring stable policies, in which decentralized critics (with minimal theoretical policy gradient variance) outperform others.
In addition, we find that less partially observable environments are not sensitive to the biases caused by the state-based critics; we point out that those environments are widely used and discuss how to identify those tasks.
Overall, critic selection should be a conscious decision dependent on the task.
We give general practical advice on how to make trade-offs based on different types of environments.

This paper combines our work on analyzing centralized history-based critics~\cite{lyu2021contrasting} and centralized state-based critics~\cite{lyu2022deeper}.
We unify previous works' assumptions and mathematical notations and propose discounted visitation probabilities for properly analyzing actor-critic algorithms in cooperative multi-agent reinforcement learning, which further formalizes the theoretical results given in previous works while moving our assumptions closer to standard practices.
We also emphasize the critical consideration that some tasks are not readily available for centralized training (e.g. when a simulator is not available). 
In those cases where centralized training is cumbersome or impossible, decentralized critics should be the default choice. 
In addition, we provide bias and variance analysis for the recommended alternative, the history-state-based critic.

\section{Related Work}

The CTDE training paradigm is used in a number of recent deep MARL approaches.
Value-based CTDE approaches such as QMIX, QPLEX, and others focus on how centralized values can be reasonably factorized into decentralized ones, and have shown promising results~\cite{QTRAN,MAVEN,ROMA,WeightedQMIX,peng2021facmac,de2020deep,xiao_icra_2020,NDQ,QMIX,VDN}.
On the other hand, CTDE policy gradient methods are almost entirely based on centralized critics.

One of the first methods featuring a centralized critic was COMA~\cite{foerster:aaai18}. COMA 
adopted a centralized state-based critic with a counterfactual baseline; the state-based critic then became the standard in many other approaches.
Regarding convergence properties, COMA claims that the overall effect of a centralized critic on the decentralized policy gradient may be reduced to a single-agent actor-critic approach, which ensures convergence under similar assumptions~\cite{konda2000actor}; however, the assumption only holds in fully-observable environments and is incorrect for partially observable environments.
In this paper, we clarify and expand on the theory of centralized critics by developing the convergence properties and a bias/variance analysis for centralized and decentralized critics, and their respective policies.
Due to their theoretical properties, we provide separate discussions for state-based and history-based critics.

Concurrently with COMA, MADDPG~\cite{lowe2017multi} proposed to use a dedicated centralized critic for each agent in semi-competitive domains, demonstrating compelling empirical results in continuous action environments. 

Many other agents extend the ideas of COMA and MADDPG. We discuss some of them but many more have been developed. 
M3DDPG~\cite{M3DDPG} focuses on the competitive case and extends MADDPG to learn robust policies against altering adversarial policies by optimizing a minimax objective.
On the cooperative side, SQDDPG~\cite{SQDDPG} borrows the counterfactual baseline idea from COMA and extends MADDPG to achieve credit assignment in fully cooperative domains by reasoning over each agent's marginal contribution. 
Other researchers also use critic centralization for emergent communication with decentralized execution in TarMAC~\cite{TarMAC} and ATOC~\cite{ATOC}.
There are also efforts utilizing an attention mechanism addressing scalability problems in MAAC~\cite{MAAC}.
Also, teacher-student style transfer learning LeCTR~\cite{LeCTR} builds on top of centralized critics, which does not assume expert teachers. 
Other work includes multi-agent credit assignment and exploration in LIIR and LICA~\cite{LIIR,LICA}, goal-conditioned policies with CM3~\cite{CM3}, and for temporally abstracted policies~\cite{chakravorty2020option}.
Extensive tests based on a centralized critic in a more realistic environment using self-play for hide-and-seek~\cite{openAI} have demonstrated impressive results showing emergent tool use.
Note that impressive results also use a state-based critic, which is a common practice and used in works such as SQDDPG~\cite{SQDDPG}, LIIR~\cite{LIIR}, LICA~\cite{LICA}, VDAC-mix~\cite{VDAC}, DOP~\cite{DOP} and MACKRL~\cite{schroeder2019multi}.
As mentioned above, these state-of-the-art works use centralized critics, but they do not specifically focus on the effectiveness of centralized critics, which is the main focus of this paper.

\section{Background}

This section introduces the formal problem definition of cooperative MARL with decentralized execution and partial observability.
We introduce various forms of value functions and formalize a set of commonly accepted on-policy history (and state) distributions.
We also introduce multi-agent actor-critic methods in which the value functions, both centralized and decentralized, are approximated by critic models.
In the coming definitions and the rest of this document, we use $\Delta\mathcal{X}$ to denote the set of probability distributions over a set $\mathcal{X}$.

\subsection{Dec-POMDPs}
Decentralized partially observable Markov decision processes (Dec-POMDPs)~\cite{Oliehoek} are multi-agent cooperative sequential decision making problems.  A Dec-POMDP is a tuple $\langle \mathcal{I}, \mathcal{S}, \As, \mathbf{\Omega}, \mathcal{T}, \mathcal{O}, R, \gamma \rangle$, composed of
%
    %
    a set of agents $\mathcal{I}$,
    a state space $\mathcal{S}$, with initial state $s_0\in\mathcal{S}$,
    a joint action space $\As \doteq \bigtimes_{i\in\mathcal{I}}\mathcal{A}_i$, one per agent,
    a joint observation space $\mathbf{\Omega} \doteq \bigtimes_{i\in\mathcal{I}}\Omega_i$, one per agent,
    a stochastic state transition function $\mathcal{T}\colon \mathcal{S} \times \As \to \Delta\mathcal{S}$ that determines state transitions $\Pr(s'\mid s, \as)$,
    a stochastic joint observation function $\mathcal{O}\colon \As\times\mathcal{S} \rightarrow \Delta\mathbf{\Omega}$ that determines observation emissions $\Pr(\os\mid \as, s)$,
    a joint reward function $R\colon \mathcal{S} \times \As \to \realset$ that is shared by all agents, and
    a discount factor $\gamma\in[0, 1)$.
    %

Control in Dec-POMDPs is performed by a set of decentralized agent policies $\policies = \langle \pi_1, \ldots, \pi_{|\mathcal{I}|}\rangle$, each representing a (stochastic) mapping from each agents' individual action-observation history to its next action, $\pi_i\colon \mathcal{H}_i\to\Delta\mathcal{A}_i$,
where $\mathcal{H}_i$ is the set of \emph{action-observation} histories for agent $i$.
For instance, agent $i$'s history at timestep $t$ is the sequence of all previous actions and observations $h_{i,t} = \langle o_{i,0},a_{i,0},o_{i,1},\ldots,a_{i,t-1},o_{i,t}\rangle$.  A \emph{joint} history is the set of all agent histories $\hs_t=\langle h_{1,t}, \ldots, h_{|\mathcal{I}|,t} \rangle$.
The set of actions chosen by each agent forms the \emph{joint} action $\as_t = \langle a_{1,t}, ..., a_{|\mathcal{I}|,t} \rangle$.  As feedback from the system, a scalar reward $R(s_t, \as_t)$ is shared by all agents, and each agent receives a local observation $\langle o_{1,t},\ldots,o_{|\mathcal{I}|,t} \rangle \sim \mathcal{O}(\as_{t-1},s_t)$.

The objective of all agents is to maximize the total performance, i.e., the expected discounted sum of future rewards $J\doteq \Exp\left[ \sum_{t=0}^{\infty} \gamma^t r_{t} \right]$.

%
%

\subsection{Discounted Visitations: Counts and Distributions}
\label{sec:discounted_visitations}

The pairing of a Dec-POMDP with a set of agent policies $\policies$ fully determines the (stochastic) behavior of the system, i.e., the marginal, joint, and conditional probability of the defined random variables (e.g., states and histories). However, before defining core RL concepts like value functions and describing policy gradient variants, it is helpful to define a set of functions related to the likelihood of occurrences of certain states or histories.

\subsubsection{Discounted Visitation Counts}

First, we define the \emph{discounted visitation counts} function $\eta$, which represents a discounted notion of the expected number of times that the system variables take some given value. In the case of the discounted state visitations (also defined in a different but equivalent form for single-agent fully observable control in~\cite{Sutton1998}), we define
\begin{equation}
\eta(s) \doteq \sum_{t=0}^\infty \gamma^t \Pr(S_t = s) \,.
\end{equation}

Note that $\eta$ is formally a function of the joint policies $\policies$, and should technically be denoted as $\eta^{\policies}$;  however, given the lack of ambiguity, we omit the suffix to simplify notation.  The discount factor $\gamma$ guarantees that $\eta$ is well-defined and finite for any Dec-POMDP, even those that never terminate or keep visiting the same states ad infinitum---hence the importance of this \emph{discounted} notion of visitations.  

We will also use $\eta$ for the number of times a (joint) history or (joint) history-state pairs are visited.
We use the same overloaded symbol $\eta$ due to similarity with the state visitation counts;  the distinction is immediately clear from the inputs and context. 
In the case of history visitations and history-state visitations, we define
\begin{align}
\eta(\hs) &\doteq \sum_{t=0}^\infty \gamma^t \Pr(\Hs_t = \hs) \,, &
\eta(\hs, s) &\doteq \sum_{t=0}^\infty \gamma^t \Pr(\Hs_t = \hs, S_t = s) \,,
\end{align}

We note some relevant properties that either relate or are shared by $\eta(s)$, $\eta(\hs)$, and $\eta(\hs)$.  First, we note that $\eta(s) = \sum_\hs \eta(\hs, s)$ and $\eta(\hs) = \sum_s \eta(\hs, s)$.  Then, we note that all of these discounted counts ultimately add up to the same value, as a direct consequence of the geometric series component based on $\gamma$, i.e., $\sum_s \eta(s) = \sum_\hs \eta(\hs) = \sum_{\hs,s} \eta(\hs, s) = (1-\gamma)\I$.  In the case of $\eta(\hs)$ and $\eta(\hs, s)$, we also note that the sum over timesteps $t$ is technically superfluous, since there is only one timestep where the joint history $\hs$ can possibly be collectively seen by the agents;  therefore, $\eta(\hs)$ and $\eta(\hs, s)$ are equivalently written as
\begin{align}
\eta(\hs) &= \left. \gamma^t \Pr(\Hs_t = \hs) \right|_{t=\abs{\hs}} \,, &
\eta(\hs, s) &= \left. \gamma^t \Pr(\Hs_t = \hs, S_t = s) \right|_{t=\abs{\hs}} \,.
\end{align}

Finally, we further consider visitation variants that include action counts,
\begin{align}
\eta(s, \as) &\doteq \sum_{t=0}^\infty \gamma^t \Pr(S_t = s, \As_t = \as) \,, \\
\eta(\hs, \as) &\doteq \sum_{t=0}^\infty \gamma^t \Pr(\Hs_t = \hs, \As_t = \as) \,, \\
\eta(\hs, s, \as) &\doteq \sum_{t=0}^\infty \gamma^t \Pr(\Hs_t = \hs, S_t = s, \As_t = \as) \,,
\end{align}
\noindent which share similar properties to their action-less counterparts.  Importantly, counts $\eta(\hs, \as)$ and $\eta(\hs, s, \as)$  are respectively related to $\eta(\hs)$ and $\eta(\hs, s)$ by the policy, according to $\eta(\hs, \as) = \eta(\hs) \policies(\as; \hs)$ and  $\eta(\hs, s, \as) = \eta(\hs, s) \policies(\as; \hs)$.

\subsubsection{Discounted Visitation Probabilities}
\label{sec:rho}

Next, we define the \emph{discounted visitation probability} functions $\rho$ as normalized versions of the corresponding counts $\eta$.  Given that $\eta$ adds up to $(1-\gamma)\I$, this results in
\begin{align}
\rho(s) &\doteq (1-\gamma) \eta(s) \,, &
\rho(\hs) &\doteq (1-\gamma) \eta(\hs) \,, &
\rho(\hs, s) &\doteq (1-\gamma) \eta(\hs, s) \,, \nonumber \\
\rho(s, \as) &\doteq (1-\gamma) \eta(s, \as) \,, &
\rho(\hs, \as) &\doteq (1-\gamma) \eta(\hs, \as) \,, &
\rho(\hs, s, \as) &\doteq (1-\gamma) \eta(\hs, s, \as) \,.
\end{align}

Each $\rho$ now represents a probability distribution over the space of its inputs.  Further, common marginalization properties hold, e.g., $\rho(s) = \sum_\hs \rho(\hs, s)$, and $\rho(\hs) = \sum_s \rho(\hs, s)$.  We can also further overload $\rho$ to encompass a notion of conditional probability, e.g.,
\begin{align}
\rho(s\mid \hs) &\doteq \frac{ \eta(\hs, s) }{ \eta(\hs) } \,, &
\rho(\hs\mid s) &\doteq \frac{ \eta(\hs, s) }{ \eta(s) } \,, \nonumber \\
\rho(\as\mid \hs) &\doteq \frac{ \eta(\hs, \as) }{ \eta(\hs) } \,, &
\rho(\as\mid s) &\doteq \frac{ \eta(s, \as) }{ \eta(s) } \,,
\end{align}
\noindent for which common conditional properties also hold, e.g., $\rho(s\mid \hs) = \rho(\hs, s) \mathbin{/}\rho(\hs) $, $\rho(\hs\mid s) = \rho(\hs, s) \mathbin{/} \rho(s)$, $\rho(\as\mid\hs) = \rho(\hs, \as) \mathbin{/} \rho(\hs) $, and $\rho(\as\mid s) = {\rho(s, \as) } \mathbin{/} { \rho(s) }$. We also note that $\rho(\as\mid \hs) = \rho(\as\mid \hs, s) = \policies(\as; \hs)$.
Finally, we note that some of these $\rho$-function outputs are equivalent to direct probabilities induced by the Dec-POMDP's graphical model.
Most notably, $\rho(s\mid \hs)$ is equivalent to the conditional state probability given the joint history $\Pr(s\mid \hs)$. For others, however, there is no such corresponding probability, e.g., $\rho(\hs\mid s)$ is mathematically well-defined, while $\Pr(\hs\mid s)$ is ill-defined without assuming a particular timestep for $\hs$~\cite{baisero2022unbiased}.
We use $\rho$ to primarily simplify the notation associated with policy gradients (\Cref{sec:multi_agent_actor_critic_methods}), although we will also exploit the conditional-$\rho$ functions to resolve a formal issue that appears in our definition of a specific form of centralized value function (\Cref{sec:state-values}).

\subsection{Value Functions}
\label{sec:value_functions}

In this section, we formally define various types of value functions that are used in different forms of multi-agent policy gradient:  the joint history value function $\qpolicies(\hs, \as)$, the individual-history value function $\qpolicies_i(h, a)$, the state value function $\qpolicies(s, \as)$, and the joint history-state value function $\qpolicies(\hs, s, \as)$.  These value functions all represent some notion of expected (discounted) performance obtained by the entire team of agents, that is given and indicated as a suffix (even for the individual history case $\qpolicies_i$), and they differ exclusively in terms of the information that is available to determine the expected team performance.

\subsubsection{Joint History Value Function $\qpolicies(\hs, \as)$}
\label{sec:value_functions:hs}

The joint history value function $\qpolicies(\hs, \as)$ is a form of centralized value function, and the unique solution to the following \emph{joint history} Bellman equality,
\begin{equation}
\qpolicies(\hs, \as) = R(\hs, \as) + \gamma \Exp_{\os\mid \hs,\as}\left[ \sum_{\as'} \policies(\as'; \hs\as\os) \qpolicies(\hs\as\os, \as') \right] \,, \label{eq:bellman:q_hs_as}
\end{equation}
\noindent where $R(\hs, \as) \doteq \Exp_{s\mid \hs}\left[ R(s, \as) \right]$ is the \emph{joint history} reward function.
$\qpolicies(\hs, \as)$ is the expected long-term performance of the team of agents when each individual agent policy $\pi_i$ has observed the individual history $h_i$ and has opted to perform a first action $a_i$.

\subsubsection{Individual History Value Function $\qpolicies_i(h, a)$}
\label{sec:value_functions:hi}

The individual history value function $\qpolicies_i(h, a)$ is a form of decentralized value function from the singular perspective of the $i$-th agent, and the unique solution to the following \emph{individual history} Bellman equality,
\begin{equation}
\qpolicies_i(h, a) = \rpolicies_i(h, a) + \gamma \Exp_{o\mid h, a}\left[ \sum_{a'} \policy_i(a'; hao) \qpolicies_i(hao, a') \right] \,, \label{eq:bellman:qi_h_a}
\end{equation}
\noindent where $\rpolicies_i(h, a) \doteq \Exp_{s, \as_\noti\mid h_i=h}\left[ R(s, \as) \right]$ is the \emph{individual history} reward function, which integrates out the behavior of all other agents and the resulting state.  $\qpolicies_i(h, a)$ is the expected long-term performance of the team of agents when the $i$-th agent policy $\pi_i$ has observed the individual history $h$ and has opted to perform a first action $a$.

\subsubsection{State Value Function $\qpolicies(s, \as)$}\label{sec:state-values}
\label{sec:value_functions:s}

The state value function $\qpolicies(s, \as)$ is a form of centralized value function that \emph{attempts} to measure the expected long-term performance of the team of agents when the system state happens to be $s$, and each individual agent policy $\pi_i$ has opted to perform a first action $a_i$.  A straightforward (but na\"ive, as we will see) formalization of this notion is based on defining the state value function as the unique solution to the following \emph{state} Bellman equality,
\begin{equation}
\qpolicies(s, \as) = R(s, \as) + \gamma \Exp_{s'\mid s, \as}\left[ \sum_{\as'} \Pr(\as'\mid s') \qpolicies(s', \as') \right] \,. \label{eq:bellman:q_s_as:wrong}
\end{equation}

Although this notion of state value seems reasonable at the surface level, it suffers from a subtle formality issue that causes $\Pr(\as\mid s)$, and consequently $\qpolicies(s, \as)$ itself, to be not guaranteeably well-defined for generic control problems and teams of agents;  this is an issue intrinsic to partial observability that was already analyzed for the single-agent control case in~\cite{baisero2022unbiased}, and for the multi-agent control case in~\cite{lyu2022deeper}.

To keep this introductory section brief and compact, we redirect a more thorough discussion on the issues with $\Pr(\as\mid s)$ and \Cref{eq:bellman:q_s_as:wrong} to \Cref{sec:pr_as_s}.  Broadly speaking, the issue is related to the fact that $\Pr(\as\mid s)$ denotes a time-invariant relationship between variables that is conventionally time-variant, and is therefore undefined when a time index is not available. In fact, we note that there is no issue with a \emph{timed} variant of the state value function $\qpolicies_t(s, \as)$ defined as the solution to the following Bellman equality
\begin{equation}
\qpolicies_t(s, \as) = R(s, \as) + \gamma \Exp_{s'\mid s, \as}\left[ \sum_{\as'} \Pr(\As_{t+1}=\as'\mid S_{t+1} = s') \qpolicies_{t+1}(s', \as') \right] \,. \label{eq:bellman:q_s_as:timed}
\end{equation}
However, employing timed value functions is an unsatisfactory solution as they do not generalize well across different (and potentially many) timesteps, and is not a common practice in mainstream RL.
We thus consider the following untimed alternative definition for state values $\qpolicies(s, \as)$ as the unique solution to the following \emph{state} Bellman equality,
\begin{equation}
\qpolicies(s, \as) = R(s, \as) + \gamma \Exp_{s'\mid s, \as}\left[ \sum_{\as'} \rho(\as'\mid s') \qpolicies(s', \as') \right] \,. \label{eq:bellman:q_s_as}
\end{equation}
\noindent where $\Pr(\as'\mid s')$ has been replaced with the discounted visitation conditional probability $\rho(\as'\mid s')$ defined in \Cref{sec:rho}.  We note that $\rho(\as'\mid s')$ is inherently defined in terms of geometric series of timed probabilities, and therefore does not suffer from the same issue as $\Pr(\as'\mid s')$.  For more details on why we employ $\rho$ rather than other possible notions of the conditional relationship between $s$ and $\as$, see \Cref{sec:pr_as_s}.  Finally, we also note that $\rho(\as\mid s) = \sum_{\hs} \rho(\hs\mid s) \policies(\as; \hs)$ satisfies the sum-product rule, which solidifies an intuitive understanding of $\rho$ as a reasonable concrete substitute for the concept of $\Pr(\as\mid s)$.

\subsubsection{Joint History-State Value Function $\qpolicies(\hs, s, \as)$}
\label{sec:value_functions:hs_s}

The joint history-state value function $\qpolicies(\hs, s, \as)$ is a form of centralized value function.
$\qpolicies(\hs, s, \as)$ is the expected long-term performance of the team of agents when the unobserved environment state happens to be $s$, each individual policy agent $\pi_i$ has observed the individual history $h_i$, and has opted to perform a first action $a_i$.
It is the unique solution to the following \emph{joint history-state} Bellman equation,
\begin{equation}
\qpolicies(\hs, s, \as) = R(s, \as) + \gamma \Exp_{s', \os\mid s,\as}\left[ \sum_{\as'} \policies(\as'; \hs\as\os) \qpolicies(\hs\as\os, s', \as') \right] \,. \label{eq:bellman:q_hs_s_as}
\end{equation}

Because this history-state value function has access to both history and state information, it does not suffer from the same issue as the state-only value function;  $\Pr(\as'\mid \hs\as\os, s')$ is not only well defined, but can also be trivially reduced to the joint policy probability $\Pr(\as'\mid \hs\as\os, s') = \Pr(\as'\mid \hs\as\os) = \policies(\as'; \hs\as\os)$ due to the conditional independence between actions and states given histories, 

\subsection{Multi-Agent Actor-Critic Methods}
\label{sec:multi_agent_actor_critic_methods}

Actor-Critic methods (AC)~\cite{konda2000actor,sutton2000policy} are variants of Policy Gradient (PG) approaches that involve the training of policy and critic models. 
In this section, we define a number of standard centralized and decentralized methods. 
We primarily consider centralized training of decentralized policies~\cite{lowe2017multi,bono2018cooperative,lyu2021contrasting}, where there is one policy model per agent (each separately parameterized by $\theta_i$), and a single centralized critic model (parameterized by $\phi$).
We will omit the model parameterization when clear from the context.
To clearly distinguish critic models from the value functions that they are trained to model, we denote them with a hat, e.g., $\hat V$ is a critic model trained to model $\vpolicies$. 
%
%

Policy gradients are typically expressed in a form that depends on one of the previously defined action-value functions $\qpolicies$ (or the respective advantage function $\apolicies$).
Such values can be estimated in a number of ways;  in actor-critic, it is common to use one-step returns and the critic model to estimate $\qpolicies(\hs, \as)$ as $r + \gamma\vmodel(\hs\as\os)$, and $\qpolicies(s, \as)$ as $r + \gamma\vmodel(s')$.  In advantage actor-critic, the critic model is further used as a baseline for variance reduction,
\begin{align}
    \apolicies(\hs, \as) &\doteq \qpolicies(\hs, \as) - \vpolicies(\hs) \approx r + \gamma\vmodel(\hs\as\os) - \vmodel(\hs) \,, \\
    \apolicies(s, \as) &\doteq \qpolicies(s, \as) - \vpolicies(s) \approx r + \gamma\vmodel(s') - \vmodel(s) \,.
\end{align}

\subsubsection{Centralized Policy}

In our evaluation, we also consider the case of a fully centralized policy, i.e., a policy that does not necessarily satisfy conditional independence $\policies(\as; \hs) \neq \prod_i \policy_i(a_i; h_i)$, and that fundamentally treats all agents as a single entity that shares all observations and actions.
Although this represents a profoundly different control setting from decentralized control, it is a relevant and essential baseline that also represents an upper bound on the performance achievable by decentralized control.  The fully centralized policy case is formalized as Joint Actor-Critic (JAC)~\cite{bono2018cooperative,wang2019achieving}, which employs a centralized history critic $\qmodel(\hs,\as; \phi)$ modeled after $\qpolicies(\hs,\as)$ to update the fully centralized policy $\policies$ jointly parameterized by $\theta$.  The gradient associated with JAC is
\begin{equation}
    \nabla_{\theta} J = (1-\gamma) \Exp_{\hs, \as\sim\rho(\hs,\as)} \left[ \qpolicies(\hs, \as) \nabla_\theta \log \policies(\as; \hs, \theta) \right] \,.
\end{equation}

\subsubsection{Decentralized Policy and Critic}

\paragraph{Independent Actor-Critic (IAC)}
Among the decentralized policy gradient variants, we first consider Independent Actor-Critic (IAC)~\cite{peshkin2000learning,foerster:aaai18}, which trains decentralized policy $\policy_i(a; h)$ and critic $\qpolicies_i(h, a)$ models for each agent. Pseudocode for IAC is given in~\Cref{pseudocode:iac}.
In IAC, the gradient associated with the policy parameters $\theta_i$ can be derived as
\begin{equation}
   \nablai J = (1-\gamma) \Exp_{\hs, \as\sim\rho(\hs, \as)} \left[ \qpolicies_i(h_i, a_i) \nablai \log\pi_i(a_i; h_i, \theta_i) \right] \,. \label{eq:gradient:qiha}
\end{equation}

\subsubsection{Decentralized Policy and Centralized Critic}

Next, we consider gradient approximations that make use of centralized values and critics, which is the focus of this work;  in some cases, the approximations will be perfect and equivalent to $\nablai J(\theta_i)$, while in others they will differ.  To distinguish them more clearly from the formally correct gradient $\nablai J(\theta_i)$, we use $\grad$ to denote these approximations.
\paragraph{Independent Actor with Centralized History
Critic (IACC-H)} IACC-H is  a class of centralized critic methods where the joint history critic 
$\qmodel(\hs,\as; \phi)$ modeled after $\qpolicies(\hs, \as)$ is used to update each decentralized policy $\policy_i$~\cite{foerster:aaai18,bono2018cooperative}.
Pseudocode for IACC-H is given in~\Cref{pseudocode:iacch}.
The gradient approximation associated with IACC-H is
\begin{equation}
   \grad_\hs \doteq (1-\gamma) \Exp_{\hs, \as\sim\rho(\hs,\as)} \left[ \qpolicies(\hs, \as) \nablai \log \pi_i(a_i; h_i, \theta_i) \right] \,.
 \label{eq:gradient:qha}
\end{equation}
The IACC-H approach can be found as one of the variants of COMA~\cite{foerster:aaai18}, which also employs a variance reduction baseline.
%
%
\paragraph{Independent Actor with Centralized State Critic (IACC-S)} IACC-S employs a centralized state-based critic $\qmodel(s, \as; \phi)$ modeled after $\qpolicies(s, \as)$ to update each decentralized policy $\policy_i$.  
Pseudocode for IACC-S is given in~\Cref{pseudocode:iaccs}.
The gradient approximation associated with IACC-S is
\begin{equation}
    \grad_s \doteq (1-\gamma) \Exp_{\hs,s,\as\sim\rho(\hs, s, \as)} \left[ \qpolicies(s, \as) \nablai\log\pi_i(a_i; h_i, \theta_i) \right] \,.
\end{equation}
The IACC-S approach can be found both as a variant of COMA~\cite{foerster:aaai18} and in MADDPG~\cite{lowe2017multi}.

\paragraph{Independent Actor with Centralized History-State Critic (IACC-HS)} IACC-HS employs a centralized history-state-based critic $\qmodel(\hs, s, \as; \phi)$ modeled after $\qpolicies(\hs, s, \as)$ to update each decentralized policy $\policy_i$.
Pseudocode for IACC-HS is given in~\Cref{pseudocode:iaccs}.
The gradient approximation associated with IACC-HS is 
\begin{equation}
    \grad_{\hs,s} \doteq (1-\gamma) \Exp_{\hs,s,\as\sim\rho(\hs,s,\as)} \left[ \qpolicies(\hs, s, \as) \nablai\log\pi_i(a_i; h_i, \theta_i) \right] \,.
    \label{eq:gradient:qhsa}
\end{equation}
The IACC-HS approach has been used in some value decomposition methods such as VDAC-mix~\cite{VDAC}.



In this work, we show that $\nablai J$, $\grad_\hs$, and $\grad_{\hs,s}$ are all equivalent in expectation (i.e., unbiased), while $\grad_s$ differs in expectation (i.e., biased).  We further show they all have different variance properties.
We analyze these gradients via their single-sample Monte Carlo estimates,
\begin{align}
    \samplegrad_i(\hs, \as) &\doteq (1-\gamma) \qpolicies_i(h_i, a_i) \nablai\log\pi_i(a_i; h_i, \theta_i) \,, \\
    \samplegrad_\hs(\hs, \as) &\doteq (1-\gamma) \qpolicies(\hs, \as) \nablai\log\pi_i(a_i; h_i, \theta_i) \,, \\
    \samplegrad_s(\hs, s, \as) &\doteq (1-\gamma) \qpolicies(s, \as) \nablai\log\pi_i(a_i; h_i, \theta_i) \,, \\
    \samplegrad_{\hs,s}(\hs, s, \as) &\doteq (1-\gamma) \qpolicies(\hs, s, \as) \nablai\log\pi_i(a_i; h_i, \theta_i) \,, \\
    \intertext{such that}
    \nablai J &= \Exp_{\hs,\as\sim\rho(\hs,\as)} \left[ \samplegrad_i(\hs, \as) \right] \,, \\
    \grad_\hs &= \Exp_{\hs,\as\sim\rho(\hs,\as)} \left[ \samplegrad_\hs(\hs, \as) \right] \,, \\
    \grad_s &= \Exp_{\hs,s,\as\sim\rho(\hs,s,\as)} \left[ \samplegrad_s(\hs, s, \as) \right] \,, \\
    \grad_{\hs,s} &= \Exp_{\hs,s,\as\sim\rho(\hs,s,\as)} \left[ \samplegrad_{\hs,s}(\hs, s, \as) \right] \,.
\end{align}
\noindent To simplify notation, we will occasionally refer to these sample gradients simply as $\samplegrad_i$ $\samplegrad_\hs$, $\samplegrad_s$, and $\samplegrad_{\hs,s}$, with their inputs omitted and implicitly sampled from the appropriate distributions.
Finally, all the methods described above are summarized in \Cref{tab:summary} for convenience.

\begin{table}[t]
  \centering
  \begin{tabular}{lll}
    Abbreviation & Actor & Critic \\
    \hline
    JAC & Joint Actors & Joint Critic $\qpolicies(\hs, \as)$ \\
    IAC & Independent Actors & Decentralized Critics $\qpolicies_i(h_i, a_i)$ \\
    IACC-H & Independent Actors & Centralized Critic, history-based $\qpolicies(\hs, \as)$ \\
    IACC-S & Independent Actors & Centralized Critic, state-based $\qpolicies(s, \as)$ \\
    IACC-HS & Independent Actors & Centralized Critic, history-state-based $\qpolicies(\hs, s, \as)$ \\
  \end{tabular}
  \caption{Summary of methods.}
  \label{tab:summary}
\end{table}

\paragraph{Notes on Implementation Details} We briefly note that most practical implementations of policy gradients deviate from these theoretical gradient derivations in two ways:  firstly, the factor $(1-\gamma)$ can be ignored in practice because optimization practices primarily care about the gradient direction rather than the gradient scale.  Modern optimizers also tend to have adaptive step-size mechanisms that in some ways make the scaling factor irrelevant.  Secondly, practical implementations estimate the gradient expectations using empirical on-policy experience that does not exactly represent a sample from $\rho$, since it does not take into account the discount factor present in the definition of $\rho$.  In practice, this leads to some bias. However, it is likely to be a minor bias for most control problems.
In this work, we are interested in analyzing the theoretical properties of the correct gradient definitions. We will therefore consider the discrepancies between the formally correct gradients and common practices as implementation details to be ignored in the theoretical analysis.

\paragraph{Note on Value Function Assumption}
Our following theories are based on the mathematical definitions of value functions rather than direct application of critic models; this is equivalent to assuming that the critic models have been trained sufficiently and accurately until optimal convergence.
\begin{assumption}[Accurate Critics]\label{assumption:accurate-critics}
We assume critic models that are trained to accurately represent the respective values, i.e., 
\begin{align}
    \qmodel(\hs, \as) &= \qpolicies(\hs, \as) \,, \\
    \qmodel_i(h_i, a_i) &= \qpolicies(h_i, a_i) \,, \\
    \qmodel(s, \as) &= \qpolicies(s, \as) \,, \\
    \qmodel(\hs, s, \as) &= \qpolicies(\hs, s, \as) \,.
\end{align}
\end{assumption}
This assumption often exists in the form of infinitesimal step sizes for the actors~\cite{konda2000actor,singh2000nash,zhang2010multi,bowling2001convergence,foerster:aaai18} for convergence arguments of AC, since the critics are on-policy return estimates and the actors need an unbiased and up-to-date critic.
Although this assumption is in line with previous theoretical works, it is nevertheless unrealistic;
we discuss the practical implications of relaxing this assumption in \Cref{sec:real_learning_rate}.

\section{Bias and Variance of History-Based Critics}
\label{sec:bias_analysis}
In this section, we establish the convergence of the two history-based critic methods: Independent Actor-Critic (IAC) and Independent Actor with Centralized History Critic (IACC-H).
We prove that the decentralized policies receive the same expected gradient from the centralized critic and decentralized critics.
We thus prove that the centralized critic provides unbiased and correct on-policy return estimates,
but at the same time, makes the agents suffer from the same action shadowing problem~\cite{matignon2012independent} seen in decentralized learning.
It is reassuring that the centralized critic will not encourage a decentralized policy to pursue a centralized policy that is only achievable in a centralized manner but also calls into question the benefits of a centralized critic.

\subsection{Bias Analysis of Individual History-Based Gradients}
\label{nobias_subsec}

We now show that the gradient updates for the decentralized policies in IAC and IACC-H are the same in expectation.
Our work can be thought to extend an early (but perhaps underappreciated) analysis, the difference is that we consider more types of value functions and we use the AC framework and investigate on-policy return values (potentially learned by a critic), rather than Monte Carlo returns;

We now establish a relationship between the values learned by centralized and decentralized critics.
\begin{restatable}{lemma}{thmidentityqihaqha}
    \label{thm:identity:qiha:qha}
    Value functions $\qpolicies_i(h_i, a_i)$ and $\qpolicies(\hs, \as)$ are related by
    \begin{equation}
        \label{eq:identity:qiha:qha}
        \qpolicies_i(h_i,a_i) = \expect{\hs,\as\mid h_i,a_i}{\qpolicies(\hs, \as)} \,.
    \end{equation}
\end{restatable}
\noindent (Proof in \Cref{sec:value_function_identities}).
\Cref{thm:identity:qiha:qha} implies that after the critics have converged to their respective on-policy values, the policy gradients for IACC-H and IAC are equal in expectation (and hence mutually unbiased).
\begin{theorem}\label{thm:unbiased_gradient}
    The IACC-H sample gradient is an unbiased estimate of the IAC sample gradient, i.e., $\grad_\hs = \Exp\left[ \samplegrad_\hs \right] = \Exp\left[ \samplegrad_i \right] = \nablai J $.
\end{theorem}
\begin{proof}
    From \Cref{thm:identity:qiha:qha}, the decentralized value function becomes a marginal expectation of the centralized value function after convergence.
    Under the expectation over joint histories and joint actions conditioned on $(h_i,a_i)$, these two fixed points are identically equal.
    Thus, when we invoke \Cref{eq:identity:qiha:qha} in the definition of $\grad_\hs$ in \Cref{eq:gradient:qha}, we exactly recover the definition of $\nablai J$ in \Cref{eq:gradient:qiha}:
    \begin{align}\label{thm:unbiased_gradient:proof}
        \grad_\hs &= (1-\gamma) \expect{\hs,\as}{ \qpolicies(\hs,\as) \nablai \log \pi_i(a_i; h_i,\theta_i)} \nonumber \\
        &= (1-\gamma) \expect{h_i,a_i}{ \expect{\hs,\as \mid h_i,a_i}{ \qpolicies(\hs,\as) } \nablai \log \pi_i(a_i; h_i,\theta_i) } \nonumber \\
        &= (1-\gamma) \expect{h_i,a_i}{ \qpolicies(h_i, a_i) \nablai \log \pi_i(a_i; h_i,\theta_i) } \nonumber \\
        &= \nablai J \,.
    \end{align}
\end{proof}

\Cref{thm:unbiased_gradient} establishes the equivalence between the policy gradients given by centralized and decentralized history-based critics.
Intuitively, the proof suggests that the marginalization of other agents' histories $\expect{\hs,\as \mid h_i,a_i}{ \qpolicies(\hs,\as)}$ occurs in different forms with both centralized and decentralized critics.
For decentralized critics, the marginalization is already done implicitly during value learning, as shown in~\Cref{thm:identity:qiha:qha} (\Cref{eq:identity:qiha:qha}).
For centralized critics, this marginalization happens explicitly at the policy gradient step, as shown in~\Cref{thm:unbiased_gradient:proof}.
The implication is significant but may be counter-intuitive.
That is, since the expected gradients given by both centralized and decentralized critics are equal, we must conclude that a centralized critic does not lead to better coordination and cooperation in policy learning.
In other words, in theory, we cannot expect to obtain a better policy with critic centralization.

\subsubsection{Climb Game Example} \label{sec:climb_game_example}

\begin{table}[t]
  \centering
  \renewcommand{\arraystretch}{1.3} 
  \begin{tabular}{l l|r r r}
    & & \multicolumn{3}{c}{agent 1} \\
    & & \(u_1\) & \(u_2\) & \(u_3\) \\
    \hline

    \multirow{3}{*}{agent 2} &

      \(u_1\) & 11  & -30 & 0 \\
    & \(u_2\) & -30 & 7   & 6 \\
    & \(u_3\) & 0   & 0   & 5 \\

  \end{tabular}
  \caption{Return values for Climb Game~\cite{claus1998dynamics}.}
  \label{tab:climb_game_return}
\end{table}

We use the Climb Game~\cite{claus1998dynamics}, a classic matrix game, to intuitively highlight that IAC and IACC-H give the same policy gradient in expectation.
We use $a$ to represent an agent's action variable, and $u$ to represent the concrete values that it can take, e.g., $a_1$ is the first agent's action, while $u_1$ is the first action in the action set.
The Climb Game is a state-less matrix game (see \Cref{tab:climb_game_return}) in which agents must cooperate to achieve the optimal reward of \(11\) by taking the joint action \(\langle a_1=u_1, a_2=u_1 \rangle\), while overcoming the risk of being punished by worse rewards (\(-30\) or \(0\)) when the agents fail to coordinate.
It is notoriously difficult for independent learners to converge to the optimal actions due to the low expected return an agent will receive when the other agent's policy is not already choosing the optimal action.

This cooperation issue arises when a potentially good action $u$ has low on-policy values and high returns are conditional on other agents' policies. In that case, both agents would have to take the lower-valued action (multiple times) to learn to choose it.
This creates a dilemma where agents are less prone to take the optimal action $u$, locking themselves into other local optima.
In the Climb Game, the value of $u_1$ is shadowed because it does not produce a satisfactory return unless the other agent also takes $u_1$ frequently enough.
This commonly occurring multi-agent local optimum is called a \emph{shadowed equilibrium}~\cite{fulda2007predicting,panait2008theoretical}, a known difficulty in independent learning which usually requires an additional coordination mechanism. 

Consider solving the Climb Game with IAC, assuming the agents start with uniformly random policies.
The independent value function $\qpolicies_i$ used in IAC results in action values,
\begin{align}
\qpolicies_1(u_1) &= \frac{11-30}{3} = -\frac{19}{3} \,, &
\qpolicies_1(u_2) &= \frac{-30+7}{3} = -\frac{23}{3} \,, &
\qpolicies_1(u_3) &= \frac{6+5}{3} = \frac{11}{3} \,, \nonumber \\
\qpolicies_2(u_1) &= \frac{11-30}{3} = -\frac{19}{3} \,, &
\qpolicies_2(u_2) &= \frac{-30+7+6}{3} = -\frac{17}{3} \,, &
\qpolicies_2(u_3) &= \frac{5}{3} \nonumber \,,
\end{align}
\noindent making $u_3$ the most attractive action to both agents.
Note that we only compare the action values $\qpolicies$ to determine which one is favored, because in all cases the gradient term $\nablai \log\pi_i(a_i)$ is the same (\Cref{eq:gradient:qiha}). In this case, both agents will update towards favoring $u_3$, leading them to never favor $u_1$ and never converge to the optimal value $Q_i^*(u_1)=11$.

Consider instead solving the Climb Game with IACC-H, again assuming the agents start with uniformly random policies. 
In this case, the centralized value function $\qpolicies(\as)$ is fully capable of representing the individual joint values, mirroring the full reward matrix from \Cref{tab:climb_game_return}. 
However, despite this, the policy gradients are still computed with respect to individual agents, and averaged over other agents, therefore suffering from the same shadowing issue, which we show next for agent 1,
\begin{align}
\grad_\hs &= (1-\gamma) \Exp_{a_1, a_2} \left[ \qpolicies(a_1, a_2) \nablai\log\policy_1(a_1) \right] \nonumber \\
&= (1-\gamma) \frac{1}{9} \left( \qpolicies(u_1, u_1) + \qpolicies(u_1, u_2) + \qpolicies(u_1, u_3) \right) \nablai\log\policy_1(u_1) \nonumber \\
&\phantom{{}={}}+ (1-\gamma) \frac{1}{9} \left( \qpolicies(u_2, u_1) + \qpolicies(u_2, u_2) + \qpolicies(u_2, u_3) \right) \nablai\log\policy_1(u_2) \nonumber \\
&\phantom{{}={}}+ (1-\gamma) \frac{1}{9} \left( \qpolicies(u_3, u_1) + \qpolicies(u_3, u_2) + \qpolicies(u_3, u_3) \right) \nablai\log\policy_1(u_3) \nonumber \\
%
%
&= (1-\gamma) \left( -\frac{19}{9} \nablai\log\policy_1(u_1) -\frac{23}{9} \nablai\log\policy_1(u_2) + \frac{11}{9} \nablai\log\policy_1(u_3) \right) \,,
%
%
\end{align}
\noindent which overall still ends up preferring and promoting the sub-optimal action $u_3$ over the others.
In fact, this should be of no surprise considering that the IAC and AICC-H gradients are the same in expectation, according to \Cref{thm:unbiased_gradient}, and are therefore subject to the same theoretical cooperation issues in expectation.

\begin{figure}
    \centering
    \includegraphics[width=0.4\textwidth]{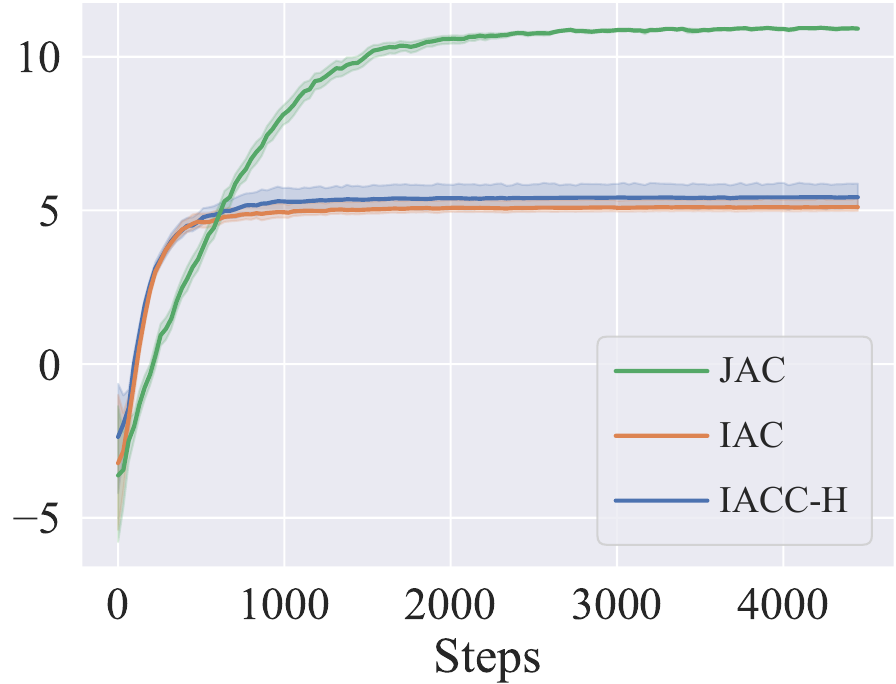}
    \caption{Climb Game empirical returns showing both decentralized and centralized critic methods succumb to the \emph{shadowed equilibrium} problem (showing mean and standard deviation over 50 runs per method).}
    \label{fig:climb_game1}
\end{figure}

Empirical evaluation on the Climb Game (shown in \Cref{fig:climb_game1})
confirms our analysis, showing both methods converge to the suboptimal
solution $\langle u_3, u_3 \rangle$.
At the same time, also unsurprisingly, a centralized controller always gives the optimal solution $\langle u_1,u_1\rangle$.
In general, we observe that the centralized critic has the information of the optimal solution, information that is only obtainable in a centralized fashion and is valuable for agents to break out of their cooperative local optima.
However, this information is unable to be effectively utilized by the individual actors to form a cooperative policy.
Therefore, contrary to the common intuition, in its current form, the centralized critic is unable to foster cooperative behavior more easily than the decentralized critics.

\subsection{Variance Analysis of Individual History-Based Gradients}
\label{sec:variance_analysis}
In this section, we first show that with true on-policy value functions, the centralized critic formulation can increase policy gradient variance.
More precisely, we prove that the policy gradient variance using a centralized critic is at least as large as the policy gradient variance with decentralized critics. 
We again assume that the critics have converged under fixed policies, thus ignoring the variance due to value function learning;
we discuss the relaxation of this assumption in~\Cref{sec:real_learning_rate}.
We begin by comparing the policy gradient variance between centralized and decentralized critics.
\begin{theorem} \label{thm:variance:qiha:qha}
    The IACC-H sample gradient has variance greater or equal than that of the IAC sample gradient, i.e., $\Var\left[ \samplegrad_\hs \right] \ge \Var\left[ \samplegrad_i \right]$.
\end{theorem}

\begin{proof}
    Let $\mu = \Exp\left[ \samplegrad_\hs \right] = \Exp\left[ \samplegrad_i \right]$ denote the equal expectation of the IAC-HS and IAC gradients (\Cref{thm:unbiased_gradient}), and $S = \nablai\log\policy_i(a_i; h_i) \nablai\log\policy_i(a_i; h_i)\T$ denote the outer product of the policy's score-function vector with itself.
    Next, we compare the covariance matrices of the two sample gradients,
    \begin{align}
        &\phantom{{}={}} \Cov \left[ \samplegrad_\hs \right] - \Cov \left[ \samplegrad_i \right] \nonumber \\
        &= \Exp \left[ \samplegrad_\hs \samplegrad_\hs\T \right] - \mu\mu\T - \left( \Exp\left[ \samplegrad_i \samplegrad_i\T \right] - \mu\mu\T \right)  \nonumber \\
        &= \Exp \left[ \samplegrad_\hs \samplegrad_\hs\T \right] - \Exp\left[ \samplegrad_i \samplegrad_i\T \right] \nonumber \\
        &= (1-\gamma) \Exp_{\hs,\as} \left[ \qpolicies(\hs, \as)^2 S \right] - (1-\gamma) \Exp_{h_i, a_i} \left[ \qpolicies_i(h_i, a_i)^2 S \right] \nonumber \\
        &= (1-\gamma) \Exp_{h_i, a_i} \left[ \Exp_{\hs, \as\mid h_i, a_i} \left[ \qpolicies(\hs, \as)^2 \right] S \right] - (1-\gamma) \Exp_{h_i, a_i} \left[ \Exp_{\hs, \as\mid h_i, a_i} \left[ \qpolicies(\hs, \as) \right]^2 S \right] \nonumber \\
        &= (1-\gamma) \Exp_{h_i,a_i} \left[ \left( \Exp_{\hs, \as \mid h_i, a_i} \left[ \qpolicies(\hs, \as)^2 \right] - \Exp_{\hs, \as \mid h_i, a_i} \left[ \qpolicies(\hs, \as) \right]^2 \right) S \right]  \nonumber \\
        &= (1-\gamma) \Exp_{h_i,a_i} \left[ \Var_{\hs, \as \mid h_i, a_i}\left[ \qpolicies(\hs, \as) \right] S \right] \,. \label{eq:covdiff:qhsa:qiha}
    \end{align}
    Because $S$ is positive semi-definite and the variance of any quantity is non-negative, it follows that the matrix in \Cref{eq:covdiff:qhsa:qiha} has non-negative diagonal components, which in turn means that $\diag \left( \Cov\left[ \samplegrad_\hs \right] \right) \succeq \diag \left( \Cov\left[ \samplegrad_i \right] \right)$ element-wise.  Further, $\Var\left[ \samplegrad_\hs \right] = \trace \left( \Cov\left[ \samplegrad_\hs \right] \right) \ge \trace \left( \Cov\left[ \samplegrad_i \right] \right) = \Var\left[ \samplegrad_i \right]$.
    So, not only is the total variance of $\samplegrad_\hs$ greater than that of $\samplegrad_i$, but the individual variances in any of their dimensions also have the same relationship.
\end{proof}

From the perspective of an agent $i$ trained using IACC-H, the value associated with taking action $a_i$ in history $h_i$ is a random variable taking values $\qpolicies(\hs, \as)$ depending on other agents' histories $h_\noti$ and actions $a_\noti$.  The variance of this random variable is dictated by how such values change depending on the other agents' histories and actions, as determined by $\Pr(\hs, \as \mid h_i, a_i)$.
In the following subsections, we analyze this total variance increase by examining two sources:
The ``Multi-Action Variance'' (MAV) induced by the other agents' policies,
and the ``Multi-Observation Variance'' (MOV) induced by uncertainty over the other agents' histories.
In essence, MAV is influenced by the uncertainty over $\as_\noti$, while MOV by the uncertainty over $\hs_\noti$ in ~\Cref{thm:identity:qiha:qha}.

\subsubsection{Multi-Action Variance (MAV)}
\label{sec:mav}

\begin{table}[t]
   \centering
   \renewcommand{\arraystretch}{1.3} 
   \begin{tabular}{c c|c c}
    & & \multicolumn{2}{c}{agent 1} \\
    & & \(\textit{pickles}\) & \(\textit{cereal}\)  \\
    \hline
    \multirow{2}{*}{agent 2} &
      \(\textit{vodka}\) & 1 & 0  \\
    & \(\textit{milk}\) & 0 & 3    \\
   \end{tabular}
   \caption{Return values for the Morning Game. }
   \label{tab:breakfast_game}
\end{table}

The Multi-Action Variance is a component of the total value variance dictated by how other agents choose their own actions, according to their own (stochastic) policies, which randomly influences the values experienced by other agents, contributing only to their value variance $\qpolicies(\hs, \as)$, but to the variance of the IACC-H sample gradient $\samplegrad_\hs$ variance as well.  In contrast, this variance does not exist for the IAC sample gradient $\samplegrad_i$, because its associated value function $\qpolicies_i(h_i, a_i)$ intrinsically integrates out other agents' actions, per \Cref{thm:identity:qiha:qha}.

\paragraph{Morning Game Example}
The Morning Game shown in \Cref{tab:breakfast_game} is a matrix game inspired by previous work~\cite{peshkin2000learning} and consists of two agents collaborating to make breakfast by choosing liquid and solid ingredients, of which the best combination is \(\langle \textit{cereal}, \textit{milk} \rangle\).

A common misleading intuition is that the centralized critic results in lower learning variance.
This could be often justified by two separate reasoning: first, the value learning is more stable for a centralized critic; second, the centralized critic have more information and therefore is able to provide a more accurate and specific value estimate.
We agree with the former reasoning, due to the lack of environmental stochasticity in this example, the centralized critic can robustly learn accurate values after a few samples, while decentralized critics need more training to correctly average learned values over the unobserved teammate actions.
However, when we move on from critic-learning to policy-learning, assume learned or matured (or simply assuming the value function is used) to learn policies, we find the second argument is in fact misleading.
In support of this misleading intuitive argument of precise estimates translates to less variance, one might reason that for a centralized critic $\qmodel(\as)$, the joint action \textit{cereal} with \textit{milk} always returns an environmental reward of 3, while \textit{cereal} with \textit{vodka} always returns an environmental reward of 0.  On the other hand, for the decentralized critic $\qmodel_1(a_1)$, action $a_1=\textit{cereal}$ returns an environmental reward of 3 or 0 stochastically, depending on the second agent's action, making it a harder learning task with higher variance.

Contrary to the above intuition, a centralized critic actually results in \textit{higher} variance for the learning agent.
What the intuition is missing, is the fact that at the moment when the critic models are used to inform the policy gradients, they may result in additional variance intrinsic to the random variables that are given as input.
Suppose agents employ uniform random policies.  Then, the IAC value is deterministically $\qpolicies_1(\textit{cereal}) = \pi_2(\textit{milk}) \qpolicies(\textit{cereal}, \textit{milk}) + \pi_2(\textit{vodka})\qpolicies(\textit{cereal}, \textit{vodka}) = 1.5$ for $\policy_1(\textit{cereal})$ updates, and $\qpolicies_1(\textit{pickles}) = \pi_2(\textit{milk}) \qpolicies(\textit{pickles}, \textit{milk}) + \pi_2(\textit{vodka})\qpolicies(\textit{pickles}, \textit{vodka}) = 0.5$ for $\policy_1(\textit{pickles})$ updates.  In each case, the value associated with the agent's action is deterministic.  On the other hand, the IACC-H value is either $\qpolicies(\textit{cereal},\textit{milk})=3$ or $\qpolicies(\textit{cereal},\textit{vodka})=0$ for $\pi_1(\textit{cereal})$ updates, and $\qpolicies(\textit{pickles},\textit{milk})=0$ or $\qpolicies(\textit{pickles},\textit{vodka})=1$ for $\pi_1(\textit{pickles})$ updates, depending on the actual value of the second agent's random action.
In essence, decentralized values already represent the expectation over other agents' actions, therefore removing the MAV.
Therefore, under both methods, $\pi_1$ updates towards $\textit{cereal}$, but the decentralized-critic updates have less variance and result in a more deterministic improvement in favor of  $\textit{cereal}$.

\begin{figure}
    \centering
    \begin{subfigure}{.49\textwidth}
        \includegraphics[width=\textwidth]{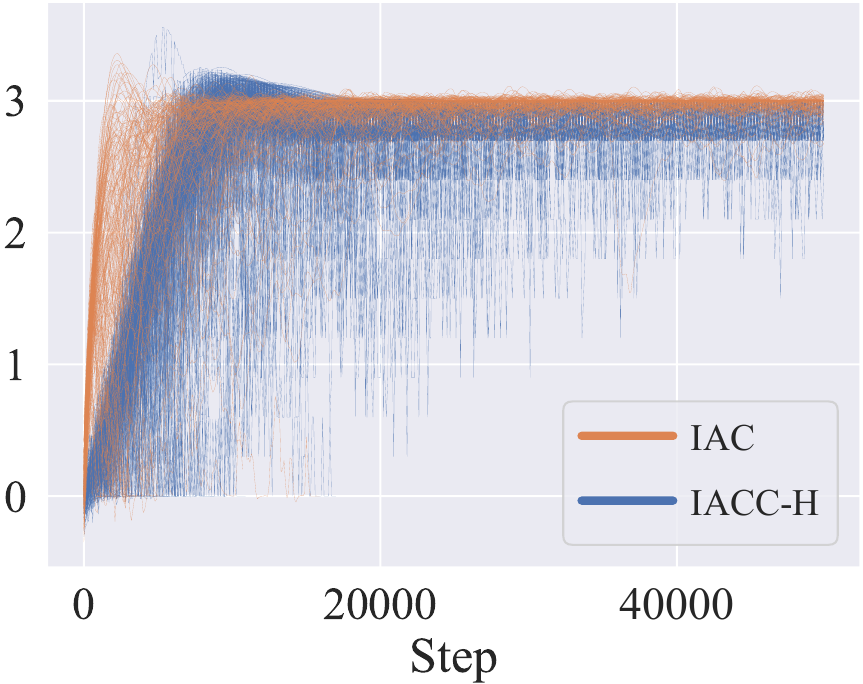}
        \caption{Individual runs, showing 40 runs per method.}\label{fig:morning_game_Qndividual}
    \end{subfigure}
    \begin{subfigure}{.49\textwidth}
        \includegraphics[width=\textwidth]{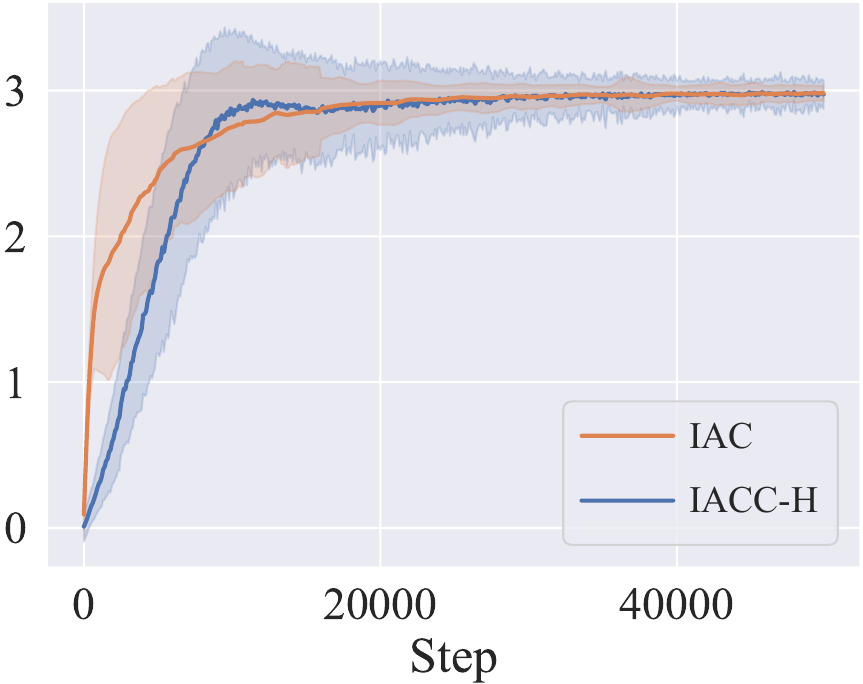}
        \caption{Mean and standard deviation.}\label{fig:morning_game_q}
    \end{subfigure}
    \caption{Q value for updating $\pi(\textit{cereal})$ over time. Showing a larger variance in the values for IACC-H than for IAC, which does not reduce in the long term.}
\end{figure}

Empirically, \Cref{fig:morning_game_Qndividual,fig:morning_game_q} show how the Q-values evolve for the optimal action $a_2$ in both methods for the Morning Game.
First, observe that both types of critics converge to the correct value, around $3$, which confirms our convergence analysis for both methods and that the resulting gradients are unbiased.
Second, the two methods' Q-value variance appears to differ significantly.
Notice that both types of value variances exhibit approximation errors across different seeds, which are also viewed as error bands in the figure.
This type of variance caused by approximation error is the only source of variance for decentralized critics.
On the other hand, centralized critics have the additional variance that comes from joint actions involving the action-in-question and potentially different actions from other agents, e.g., producing a high value when a teammate chooses $\textit{milk}$ and a low value when a teammate chooses $\textit{vodka}$.
Having to average over teammate actions is, therefore, the essence of MAV.

\subsubsection{Multi-Observation Variance (MOV)}
\label{sec:mov}
For history-based critics in the partially observable case, another source of variance in local value $ \Var_{\hs, \as \mid h_i, a_i}\left[ \qpolicies(\hs, \as) \right]$ comes from factored observations.
More concretely, for an agent having observed a particular local trajectory $h_i$, other agents'
trajectories $\hs_\noti$ may vary, and the decentralized critic must average over this observation variance and provide a single expected value for each local trajectory $\qpolicies_i(h_i,a_i)$.
The centralized critic, on the other hand, is able to distinguish each combination of trajectories $h_i$ and $\hs_\noti$, but when used for a decentralized policy at $h_i$, teammate history $\hs_\noti$ is effectively a random variable sampled from $\Pr(\hs_\noti \mid h_i)$,
and we expect the mean estimated return during the update process to be $\Exp_{\hs\mid h_i} \left[ \qpolicies(\hs, \as) \right]$.

\paragraph{Guess Game Example}
Consider a one-step two-agent task with binary uniform random observations and binary actions.
The agents are rewarded with $10$ if both their actions match the other agent's observation, $-10$ if neither does, and $0$ otherwise.
In practice, each agent can only guess randomly which action will lead to the positive reward.  More broadly, the decentralized value function takes a value of $\qpolicies_i(h_i,a_i)=0$ with zero variance for any combination of agent policies.
On the other hand, a centralized value function with global scope can distinguish between the various returns, but will return different values depending on the stochastic values of $\hs$ and $\as$, and hence estimates $\qpolicies(\hs, \as)=10$ with probability $25\%$, $\qpolicies(\hs,\as)=-10$ with probability $25\%$, and $\qpolicies(\hs,\as)=0$ with probability $50\%$, resulting in a total variance of $\frac{1}{2} 10^2 = 50$.
In this example, a centralized critic produces returns estimates with higher variance when agents have varying observations.

\begin{figure}
  \centering
  \includegraphics[width=0.5\textwidth]{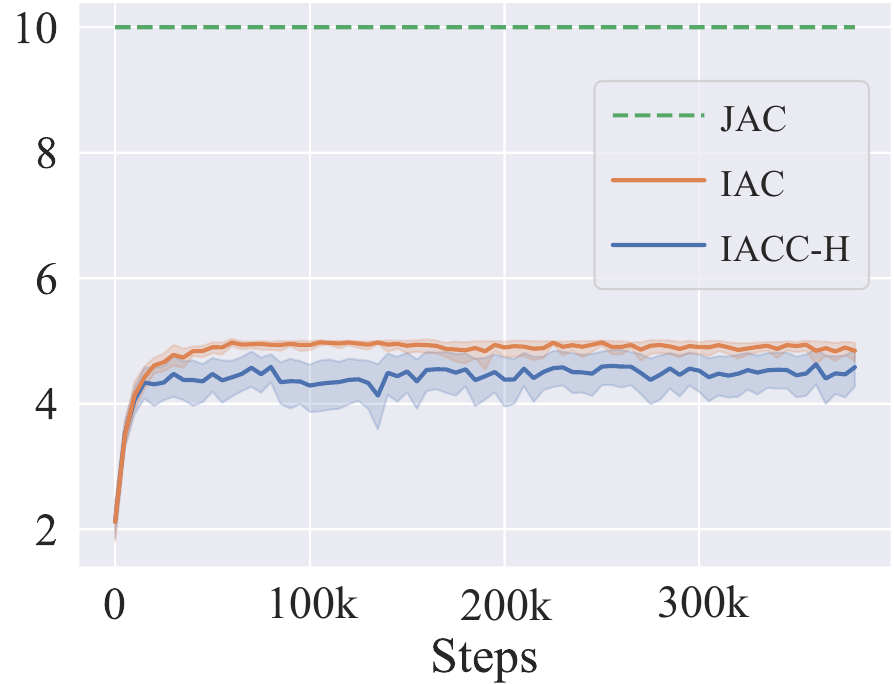}
  \caption{Performance (mean test return) comparison in Guess Game, plotting mean and standard deviation aggregating 40 runs per method; showing centralized critic cannot bias the actors towards the global optimum in the simplest situation.}
  \label{fig:mov_guess_game}
\end{figure}

\section{Bias and Variance of State-Based Critics}\label{sec:state-based-critic}

Having established the convergence of IAC and IACC-H, we will use them as a basis to analyze the convergence of the two state-based critic methods: IACC-S and IACC-HS.
We provide a theoretical bias analysis complemented by simple intuitive examples.
In this section, we extend the above analysis to  commonly used state-based critics, which are often assumed to be strictly beneficial.
However, we emphasize that using a state-based critic to train history-based policies is theoretically incorrect, as we show that it may result in bias in the policy gradient.
Even with CTDE, where we may access the states during training, we are still aiming to solve a Dec-POMDP problem where the desired output is history-based policies (that can execute independently using local information); that is, state information is not available during execution.
Therefore, history-based policies would need to infer, directly or indirectly, the underlying state distribution based on the observed histories.
For example, the state distribution may be high in entropy when little information can be derived from observations; hence in some tasks, agents ought to collect information to gain insight into the state distribution before making important decisions.
Intuitively, providing the underlying state to the history-based policies during offline training can hinder the policies' ability to reason about uncertainty because when the state is known, efforts such as information collection are unnecessary.
This information mismatch between critic (state-based) and actor (history-based) is, thus, intuitively where the bias we discuss in this section originates.
In this section, we show that the state-based critic may introduce bias into the policy gradient compared to the provably convergent history-based critics.
That is, the gradient given by the history-based critics $\nablai J$ (and $\grad_\hs$) leads to convergence to a local optimum, which is guaranteed to be a Nash equilibrium~\cite{peshkin2000learning};
however, as we will show next, the gradient given by the state-based critic is biased, so the policy is not guaranteed to also converge to the same local solution.
Since we have shown (centralized and decentralized) history-based critics give the same expected policy gradient, we will use centralized history-based critics for the comparison below.

\subsection{Bias Analysis of State-Based Gradients}\label{sec:bias}
The gradient from a state-based critic is given by:
\begin{align}
    \grad_s &= (1-\gamma) \Exp_{\hs,s\sim\rho(\hs,s),\as\sim\policies(\hs)}\left[ \qpolicies(s, \as) \nablai\log\policy_i(a_i; h_i) \right] \nonumber \\
    &= (1-\gamma) \Exp_{\hs\sim\rho(\hs),\as\sim\policies(\hs)}\left[ \Exp_{s\sim\rho(s\mid \hs)}\left[ \qpolicies(s, \as) \right] \nablai\log\policy_i(a_i; h_i) \right] .
\end{align}
Using the history-based gradient~(\Cref{eq:gradient:qha}), we see the state-value-based-gradient $\grad_s$ is also unbiased iff
\begin{align}
    \qpolicies(\hs,\as) = \Exp_{s\sim\rho(s\mid \hs)}\left[ \qpolicies(s,\as) \right] \,.
    %
    %
    %
    \label{eq:bias}
\end{align}
\noindent
Therefore, the bias of $\grad_s$ can be analyzed indirectly through $\qpolicies(s,\as)$. 
We adopt the methodology employed by prior work for the single-agent control case~\cite{baisero2022unbiased}, and will treat $\qpolicies(s,\as)$ as an estimator of $\qpolicies(\hs,\as)$.  Consequently, if $\qpolicies(s,\as)$ is unbiased, then $\grad_s$ is also unbiased.

We emphasize that expectation over state values cannot be assumed to have the information necessary to produce the correct history values.
Essentially, we highlight that history values is an expectation of values over different histories-state pairs, $\qpolicies(\hs, \as) = \Exp_{s\sim\rho(s\mid \hs)}\left[ \qpolicies(\hs, s,\as) \right]$, which is different to~\Cref{eq:bias}.
Of course, when $\qpolicies(\hs, s,\as) = \qpolicies(s, \as)$ for all $\hs$ and $s$, we can justify~\Cref{eq:bias}, yet this equality cannot assume to hold in partially observable settings.
We illustrate this point in more detail by comparing history and state value tables and their marginalizations, as well as more intuition and example, in the following paragraphs.

Consider the tabular form of the history-state function under a specific action $\as$, $\qpolicies_\as \in \realset^{|\sset| \times |\bm{\hset}|}$, such that Q table is noted as $\qpolicies_{\as,ij} = \qpolicies(\hs_i, s_j, \as)$ where $i$ and $j$ are used to index the joint history $\hs$ and a state $s$ (we will use this convention for the remainder of this section); if both the state and observation spaces are finite, $\qpolicies_\as$ will be a finite matrix.
Also consider the tabular form of the normalized discounted visitations $\ppolicies\in[0, 1]^{|\sset| \times |\bm{\hset}|}$, such that $\ppolicies_{ij} = \rho(\hs, s)$.
Note that recovering $\rho(\hs\mid s)$ and $\rho(s\mid \hs)$ from $\ppolicies$ requires renormalizing the values in a specific row or column,
\begin{align}
    \rho(\hs\mid s) &= \frac{ \rho(\hs, s) }{ \sum_{s'} \rho(s', \hs) } = \frac{ \ppolicies_{ij} }{ \sum_{j'} \ppolicies_{ij'} } \,, \\
    \rho(s\mid \hs) &= \frac{ \rho(\hs, s) }{ \sum_{\hs'} \rho(\hs, s') } = \frac{ \ppolicies_{ij} }{ \sum_{i'} \ppolicies_{i'j} } \,.
\end{align}
The matrices $\qpolicies_\as$ and $\ppolicies$ contain the necessary information to determine all marginal history $\qpolicies(\hs, \as)$ and state values $\qpolicies(s,\as)$ as normalized dot products:
\begin{align}
    \qpolicies(\hs,\as) &= \sum_s \rho(s\mid \hs) \qpolicies(\hs, s, \as) = \sum_{j'} \frac{ \ppolicies_{ij'} }{ \sum_{j'} \ppolicies_{ij'}  } \qpolicies_{\as,ij'} \,, \label{eq:tabular:vh} \\
    \qpolicies(s,\as) &= \sum_{\hs} \rho(\hs\mid s) \qpolicies(\hs, s, \as) = \sum_{i'} \frac{ \ppolicies_{i'j} }{ \sum_{i''} \ppolicies_{i''j} }
    \qpolicies_{\as,i'j} \,. \label{eq:tabular:vs}
\end{align}
\noindent On the other hand, the correct expected state value as listed in \Cref{eq:bias} is
\begin{align}
    \Exp_{s\sim\rho(s\mid h)}\left[ \qpolicies(s,\as) \right] &= \sum_s \rho(s\mid \hs) \qpolicies(s,\as) \nonumber \\
    &= \sum_s \rho(s\mid \hs) \sum_{\hs'} \rho(\hs'\mid s) \qpolicies(\hs', s, \as) \nonumber \\
    &= \sum_{j'} \frac{ \ppolicies_{ij'} }{ \sum_{j''} \ppolicies_{ij''} } \sum_{i'} \frac{ \ppolicies_{i'j'} }{ \sum_{i''} \ppolicies_{i''j'} } \qpolicies_{\as, i'j'} \,. \label{eq:tabular:Evs}
\end{align}

Note that $\qpolicies(\hs,\as)$ in \Cref{eq:tabular:vh} only involves the elements on a specific row of both $\qpolicies_\as$ and $\ppolicies$, while the expected state value in \Cref{eq:tabular:Evs} involves all elements of both $\qpolicies_\as$ and $\ppolicies$. 
While this provides an intuitive reason for the fact that the two values are not necessarily the same, the inequality proof is still incomplete;
in fact, the values in $\qpolicies_\as$ and $\ppolicies$ are not arbitrary, but are related by problem dynamics, policies, and history-state Bellman equations, which may still result in \Cref{eq:tabular:vh,eq:tabular:Evs} being numerically equivalent.
However, we prove that this is not the case.
\begin{theorem}\label{thm:bias:qha:qsa}
     $\qpolicies(s,\as)$ may be a biased estimate of $\qpolicies(\hs,\as)$, i.e., the equality
    \begin{equation}
    \qpolicies(\hs, \as) = \Exp_{s\mid \hs} \left[ \qpolicies(s, \as) \right]
    \end{equation}
    \noindent cannot be assumed to hold.
\end{theorem}

\begin{proof}
We prove \Cref{thm:bias:qha:qsa} with an argument analogous to that made in \cite{baisero2022unbiased}, which is applicable to our novel always-well-defined state value function $\qpolicies(s, \as)$.
For a generic environment, consider an arbitrary joint action $\as$, and two different joint histories $\hs_A\neq\hs_B$ associated with the same discounted conditional state visitations distribution $\rho(s\mid \hs_A) = \rho(s\mid \hs_B)$ (a reasonably common occurrence in partially observable environments). Because the joint histories are different, there are many history-based policies $\policies$ which result in different future behaviors starting from $\hs_A$ and $\hs_B$, which results in different history values,
\begin{equation}
    \qpolicies(\hs_A, \as) \neq \qpolicies(\hs_B, \as) \,. \label{eq:bias-proof:1}
\end{equation}

On the other hand, the conditional state distributions $\rho(s\mid \hs_A) = \rho(s\mid \hs_B)$ are equal, which implies that the expected state values are also equal,
\begin{equation}
    \Exp_{s\mid\hs_A}\left[ \qpolicies(s, \as) \right] = \Exp_{s\mid\hs_B} \left[ \qpolicies(s, \as) \right] \,. \label{eq:bias-proof:2}
\end{equation}

Assuming that the equality $\qpolicies(\hs, \as) = \Exp_{s\mid \hs}\left[ \qpolicies(s, \as) \right]$ holds leads to a contradiction with \Cref{eq:bias-proof:1,eq:bias-proof:2},
\begin{equation}
    \Exp_{s\mid\hs_A}\left[ \qpolicies(s, \as) \right] = \qpolicies(\hs_A, \as) \neq \qpolicies(\hs_B, \as) = \Exp_{s\mid\hs_B}\left[ \qpolicies(s, \as) \right] \,.
\end{equation}
Therefore, $\qpolicies(\hs, \as) = \Exp_{s\mid \hs}\left[ \qpolicies(s, \as) \right]$ cannot assume to hold in general.
\end{proof}

Intuitively, recall that the state values alias over all the histories that visit the state.
It is often the case that state spaces are considerably more compact than the history space.
As a result, the aliasing compresses values over history spaces into values over state spaces in a way that loses important information regarding \emph{uncertainty}. 
For a particular state, agents may or may not have gathered information through their observations, and this information usually determines their optimal next step and expected returns.
Unfortunately, those distinctions cannot be captured by the state value, hence the expected state value is biased for history-based policies.
Compared to history values, which are already established to provide the correct gradient, state values become questionable as they are not equivalent to history values in expectation, as suggested by the above~\Cref{thm:bias:qha:qsa}.
Indeed, as we will show in the next section, the biased expected values for state-based critics will lead to biased policy gradients.
We now use an example to illustrate this issue intuitively.

\subsubsection{Dec-Tiger Example}
This example also serves as alternative proof for \Cref{thm:bias:qha:qsa}.
Dec-Tiger~\cite{nair2003taming} is a classic Dec-POMDP domain where two agents face a left door and a right door, behind one of which lies a tiger.
The agents can individually \textit{listen}, \textit{open-left} or \textit{open-right}.
The \textit{listen} action is used to detect the tiger location and results in a reward of $-2$, and observations \textit{hear-left} or \textit{hear-right} which indicate the correct tiger location with probability $85\%$.
Opening doors terminate the episode immediately and result in rewards of $-50$ if both agents open the tiger door, $20$ if both agents open the non-tiger door, $-100$ if the agents open different doors, $-101$ if only one agent opens a door, and it is the tiger's door, and $9$ if only one agent opens a door, and it is not the tiger's door.

Consider the pair of agent policies that will listen in the first two timesteps, and open the most promising door in any other timestep based on their individual observations (opening randomly if neither door is more promising than the other), i.e.,
\begin{align}
\policy_i(h_i) \sim \begin{cases}
\delta_\textit{listen} & \text{if } |h_i| < 2 \,, \\
\delta_\textit{open-left} & \text{if } h_i \text{ contains more \textit{hear-right} than \textit{hear-left}} \,, \\
\delta_\textit{open-right} & \text{if } h_i \text{ contains mroe \textit{hear-left} than \textit{hear-right}} \,, \\
\mathcal{U}(\textit{open-left}, \textit{open-right}) & \text{otherwise}
\end{cases}
\end{align}

In \Cref{sec:dec-tiger}, we compute many values associated with these agent policies, including discounted counts, probabilities, state values, history-state values, and history values.  Among those, we have computed the state values associated with the \textit{listen} actions $\qpolicies(s, (\textit{listen}, \textit{listen})) = -18.175$, and the history value associated with the empty history $\qpolicies(\hs=\epsilon, (\textit{listen}, \textit{listen})) = -16.175$.
Note that state values associated with the joint listening action are invariant to the particular state, meaning that history value can never be recovered from the state value. That is, any weighted average over possible states results in the same value,
\begin{align}
\Exp_{s\mid \hs=\epsilon}\left[ \qpolicies(s, (\textit{listen}, \textit{listen})) \right] &= \frac{1}{2} \qpolicies(\textit{tiger-left}, (\textit{listen}, \textit{listen})) \nonumber \\
&\phantom{{}={}} + \frac{1}{2} \qpolicies(\textit{tiger-right}, (\textit{listen}, \textit{listen}))  \nonumber \\
&= -18.175 \,,
\end{align}
\noindent which is clearly different from $\qpolicies(\hs=\epsilon, (\textit{listen}, \textit{listen}) = -16.175$.  In this case, the two values are different but still fairly similar.  However, this need not be the case in general, and large differences may also exist, as we show next.  

Consider the joint history value associated with both agents hearing the tiger on the left twice, denoted as $\hs_{\texttt{L}} = ((\textit{hear-left}, \textit{hear-left}), (\textit{hear-left}, \textit{hear-left}))$, and joint listening action $(\textit{listen}, \textit{listen})$.  Even though the policies on their own would not listen on that history, this is still a well-defined counterfactual value.  Note that the policies will inevitably choose \textit{open-right} next, regardless of which new observation is received.  Therefore, $\qpolicies(\hs_{\texttt{L}}, (\textit{listen}, \textit{listen}))$ is the expected reward obtained when opening the right door,
\begin{align}
\qpolicies(\hs_{\texttt{L}}, (\textit{listen}, \textit{listen})) &= \Exp_{s\mid \hs_\texttt{L}} \left[ R(s, (\textit{open-right}, \textit{open-right})) \right] \nonumber \\
&= 0.999 \cdot 20 + 0.001 \cdot (-50) \nonumber \\
&= 19.93 \,.
\end{align}
On the other hand, as mentioned previously, any weighted average of $\qpolicies(s, (\textit{listen}, \textit{listen}))$ over state probabilities results in the same value,
\begin{align}
\Exp_{s\mid \hs_{\texttt{L}}} \left[ \qpolicies(s, (\textit{listen}, \textit{listen})) \right] &= 0.001 \qpolicies(\textit{tiger-left}, (\textit{listen}, \textit{listen})) \nonumber \\
&\phantom{{}={}} + 0.999 \qpolicies(\textit{tiger-right}, (\textit{listen}, \textit{listen})) \nonumber \\
&= -18.175 \,,
\end{align}
\noindent a much more substantial difference in expected values, and in turn, as we will show next, in expected gradient. Practically, state values can lead to incorrect estimations of the (history-based) policy values and then poor policy learning.

\subsubsection{Biased Gradient}

We note that state values $\qpolicies(s, \as)$ being biased estimates of history values $\qpolicies(\hs, \as)$ (per \Cref{thm:bias:qha:qsa}) does not formally imply that state gradients $\samplegrad_s$ are biased estimates of history gradients $\samplegrad_\hs$;  because an average of biased estimates could technically result in an unbiased estimate, it is still technically possible for the state gradients to be unbiased after all.  Here, we provide a definitive proof that this is not the case.

\begin{restatable}{theorem}{thmgradientbias}\label{thm:gradient-bias}
    The IACC-S and IACC-H gradients may differ, $\grad_s \neq \grad_\hs$.
\end{restatable}
\begin{proof}
See \Cref{sec:biased_state_based_critic}.
\end{proof}
\noindent
To summarize the proof, it uses the fact that state values cannot always be used to obtain history values as in \Cref{thm:bias:qha:qsa}, and for at least some environments and policy parameterizations, this bias in value translates to bias in the policy gradient.

The implication of \Cref{thm:gradient-bias} is that state-based critics are inherently flawed in training history-based policies, for which history values are established to be correct.
This may lead to arbitrarily worse performance for state-based critics;
as shown in the Dec-Tiger example above, the state-based critic provides no information as to how the policy should be updated---both states have the same state value, causing all actions to be updated equally.
This is very different from history values where observed information affects the values and aids policy changes accordingly.
However, states can be compact representations or potentially ease learning.
After discussing the variance in the state-based cased, we analyze the case of using both state and history as an alternative that is both unbiased and potentially easy to learn.

\subsection{Variance Analysis of State-Based Gradients}\label{sec:state-based-variance}

In this subsection, we discuss the effect of state values $\qpolicies(s, \as)$ on the variance of the policy gradient estimate $\samplegrad_s$.
Like our bias analysis, we discuss the oracle values $\qpolicies$ instead of its estimated counterpart $\hat Q$.
The policy gradient theorem for Dec-POMDPs explicitly requires the history value $\qpolicies(\hs, \as)$ to be the value used to weigh the policy's score function~\cite{lyu2021contrasting,bono2018cooperative}.
In that capacity, $\qpolicies(\hs, \as)$ is a specific scalar associated with the history and has no variance. 
On the other hand, using state value $\qpolicies(s, \as)$ as an estimator of $\qpolicies(\hs, \as)$ introduces variance, as $s$ is sampled from the history's associated belief $b(\hs)$.
However, it does not necessarily imply that the corresponding gradient $\samplegrad_s$ also has a higher variance than $\samplegrad_\hs$ in general.
Instead, we show that if $\qpolicies(s, \as)$ is unbiased for a given policy and a Dec-POMDP, then $\samplegrad_s$ has a variance greater than or equal to that of $\samplegrad_\hs$.

\begin{theorem}\label{thm:variance:qha:qsa}
    In the special cases where state values $\qpolicies(s, \as)$ are unbiased, the IACC-S sample gradient has variance greater or equal than that of the IACC-H sample gradient, i.e.,
    \begin{equation}
    \qpolicies(\hs, \as) = \Exp_{s\sim\rho(s\mid\hs)}\left[ \qpolicies(s, \as) \right]
     \implies \Var\left[ \samplegrad_s \right] \ge \Var\left[ \samplegrad_\hs \right] \,.
    \end{equation}
\end{theorem}

\begin{proof}

We prove the theorem by assuming that the value equality $\qpolicies(\hs, \as) = \Exp_{s\mid \hs}\left[ \qpolicies(s, \as) \right]$ holds and showing that the variance inequality also holds.
Let $\mu = \Exp\left[ \samplegrad_s \right] = \Exp\left[ \samplegrad_\hs \right]$ denote the assumed equal expectation of the IACC-S and the IACC-H gradients, and $S = \nablai\log\policy_i(a_i; h_i) \nablai\log\policy_i(a_i; h_i)\T$ denote the outer product of the policy's score-function vector with itself.
Next, we compare the covariance matrices of the two sample gradients,
\begin{align}
    &\phantom{{}={}} \Cov\left[ \samplegrad_s \right] - \Cov\left[ \samplegrad_{\hs} \right] \nonumber \\
    &= \Exp \left[ \samplegrad_s \samplegrad_s\T \right] - \mu\mu\T - \left( \Exp \left[ \samplegrad_\hs \samplegrad_\hs\T \right] - \mu\mu\T \right) \nonumber \\
    &= \Exp \left[ \samplegrad_s \samplegrad_s\T \right] - \Exp \left[ \samplegrad_\hs \samplegrad_\hs\T \right] \nonumber \\
    &= (1-\gamma) \Exp_{\hs,s,\as} \left[ \qpolicies(s, \as)^2 S \right] - (1-\gamma) \Exp_{\hs,\as} \left[ \qpolicies(\hs, \as)^2 S \right] \nonumber \\
    &= (1-\gamma) \Exp_{\hs,\as} \left[ \Exp_{s\mid \hs} \left[ \qpolicies(s, \as)^2 \right] S \right] - (1-\gamma) \Exp_{\hs,\as} \left[ \Exp_{s\mid \hs} \left[ \qpolicies(s, \as) \right]^2 S \right] \nonumber \\
    &= (1-\gamma) \Exp_{\hs,\as} \left[ \left( \Exp_{s\mid \hs} \left[ \qpolicies(s, \as)^2 \right] - \Exp_{s\mid \hs} \left[ \qpolicies(s, \as) \right]^2 \right) S \right] \nonumber \\
    &= (1-\gamma) \Exp_{\hs,\as} \left[ \Var_{s\mid \hs} \left[ \qpolicies(s, \as) \right] S \right] \,. \label{eq:covdiff:qsa:qha}
\end{align}
Because $S$ is positive semi-definite and the variance of any quantity is non-negative, it follows that the matrix in \Cref{eq:covdiff:qsa:qha} has non-negative diagonal components, which in turn means that $\diag \left( \Cov\left[ \samplegrad_s \right] \right) \succeq \diag \left( \Cov\left[ \samplegrad_\hs \right] \right)$ element-wise.  Further, $\Var\left[ \samplegrad_s \right] = \trace \left( \Cov\left[ \samplegrad_s \right] \right) \ge \trace \left( \Cov\left[ \samplegrad_\hs \right] \right) = \Var\left[ \samplegrad_\hs \right]$.
So, when also unbiased, not only is the total variance of $\samplegrad_s$ greater than that of $\samplegrad_\hs$, but the individual variances in any of their dimensions also have the same relationship.

\end{proof}

Although \Cref{thm:variance:qha:qsa} alone contains a result that is conditional to an equality which does not necessarily hold for a generic Dec-POMDP, combining \Cref{thm:bias:qha:qsa,thm:variance:qha:qsa} results in a broader statement about the overall quality of state-based policy gradient estimates.
\begin{corollary}
    IACC-S never has strictly better bias/variance properties than IACC-H, i.e., either its bias is higher (or equal), or its variance is higher (or equal), or neither is lower (or equal).
\end{corollary}

\begin{proof}
    Follows directly from \Cref{thm:gradient-bias,thm:variance:qha:qsa}.
\end{proof}

\paragraph{Beverage Example}  Consider a (single-agent) \emph{beverage} domain, in which the agent is a barista who serves \textit{coffee} or \textit{tea} to a client.  The client, who prefers \textit{coffee} or \textit{tea}, represents the randomly sampled initial state.  The agent does not observe the client's preference (we denote this as $h=\nohistory$) and receives a reward of $1$ if it chooses to serve the correct beverage and $-1$ if it chooses the wrong beverage.  In either case, the episode ends.  Suppose the agent chooses to serve \textit{tea}.  Then,
\begin{align}
    \qpolicy(s=\textit{coffee}, a=\textit{tea}) &= -1 \,, \\
    \qpolicy(s=\textit{tea}, a=\textit{tea}) &= 1 \,, \\
    \qpolicy(h=\nohistory, a=\textit{tea}) &= 0 \,.
\end{align}
While $\qpolicy(h=\nohistory, a=\textit{tea})$ is a constant with zero variance, the random variable $\qpolicy(s, a=\textit{tea})$ conditioned on $h=\nohistory$ has strictly positive variance,
\begin{align}
    \Var_{s\mid h=\nohistory}\left[ \qpolicy(s, a=\textit{tea}) \right] &= \Exp_{s\mid h=\nohistory}\left[ \qpolicy(s, a=\textit{tea}) ^2 \right] - \Exp_{s\mid h=\nohistory}\left[ \qpolicy(s, a=\textit{tea}) \right]^2 \nonumber \\
    &= \Exp_{s\mid h=\nohistory}\left[ 1 \right] - 0^2 \nonumber \\
    &= 1 \,.
\end{align}
Therefore, the policy gradient estimates have a higher variance with the state-based critic.

\section{Bias and Variance of History-State-Based Critics} \label{sec:history_state_based}

In order to avoid the potential bias issue of state-based critics yet also use the centralized state information, we  analyze the History-State-based Critic seen in the Unbiased Asymmetric Actor-Critic method for the single agent case~\cite{baisero2022unbiased}, which we term IACC-HS. 
Many CTDE works use some form of history-state value function, such as QMIX~\cite{QMIX} and its following works, in which the states are used for constructing the mixing hypernetwork; as well as seen in MAPPO~\cite{MAPPO}, where the value estimation as a whole takes both history and state as input.

\subsection{Bias Analysis of History-State-Based Gradients}

As in the single-agent case, history-state values are unbiased estimators of history values, i.e., they are in expectation equivalent.

\begin{restatable}{lemma}{thmidentityqhaqhsa}
    \label{thm:identity:qha:qhsa}
    $\qpolicies(\hs, s, \as)$ is an unbiased estimate of $\qpolicies(\hs, \as)$, i.e., the equality
    \begin{equation}
        \label{eq:identity:qha:qhsa}
        \qpolicies(\hs, \as) = \Exp_{s\mid \hs} \left[ \qpolicies(\hs, s, \as) \right] \,
    \end{equation}
    \noindent holds.
\end{restatable}
\begin{proof}
See \Cref{sec:value_function_identities}.
\end{proof}

We conclude that the joint history-state gradient $\grad_{\hs, s}$ as a whole is unbiased, i.e., equivalent in expectation to the individual history gradient $\grad_\hs$.

\begin{theorem}
\label{thm:gradient:ha:hsa}
The IACC-HS and IACC-H gradients are equal, $\grad_{\hs, s} = \grad_\hs$.
\end{theorem}

\begin{proof}
\begin{align}
    \grad_{\hs, s} &= (1-\gamma) \Exp_{\hs,s,\as} \left[ \qpolicies(\hs, s, \as) \nabla\log\policy_i(a_i; h_i) \right] \nonumber \\
    &= (1-\gamma) \Exp_{\hs,\as} \left[ \Exp_{s\mid \hs} \left[ \qpolicies(\hs, s, \as) \right] \nabla\log\policy_i(a_i; h_i) \right] \nonumber \\
    &= (1-\gamma) \Exp_{\hs, \as} \left[ \qpolicies(\hs, \as) \nabla\log\policy_i(a_i; h_i) \right] \nonumber \\
    &= \grad_\hs \,.
\end{align}
\end{proof}

The equivalence in expected gradients follows directly from the equivalence in expected value functions (\Cref{thm:identity:qha:qhsa}).
Therefore, unlike state-based critics, a history-state-based critic provides a policy gradient with similar convergence guarantees as a history-based critic.

\subsection{Variance Analysis of History-State-Based Gradients}
\label{sec:variance_of_hsc}
History-state-based critics give policy gradients with equal or more variance than history-based critics (IACC-H) because, for the same joint observation history, the system could be in a range of different states which may exhibit different history-state values. 
Since history-state-based critics are not biased, the variance analysis is analogous to \Cref{thm:variance:qiha:qha}.

\begin{theorem}\label{thm:var_ghs_gh}
    The IACC-HS sample gradient has variance greater or equal to that of the IACC-H sample gradient, i.e., $\Var\left[ \samplegrad_{\hs,s} \right] \ge \Var\left[ \samplegrad_\hs \right]$.
\end{theorem}

\begin{proof}

Let $\mu = \Exp\left[ \samplegrad_{\hs,s} \right] = \Exp\left[ \samplegrad_\hs \right]$ denote the equal expectation of the IACC-HS and the IACC-H gradients (\Cref{thm:gradient:ha:hsa}), and $S = \nablai\log\policy_i(a_i; h_i) \nablai\log\policy_i(a_i; h_i)\T$ denote the outer product of the policy's score-function vector with itself.
Next, we compare the covariance matrices of the two sample gradients,
\begin{align}
    &\phantom{{}={}} \Cov\left[ \samplegrad_{\hs,s} \right] - \Cov\left[ \samplegrad_{\hs} \right] \nonumber \\
    &= \Exp \left[ \samplegrad_{\hs,s} \samplegrad_{\hs,s}\T \right] - \mu\mu\T - \left( \Exp \left[ \samplegrad_\hs \samplegrad_\hs\T \right] - \mu\mu\T \right) \nonumber \\
    &= \Exp \left[ \samplegrad_{\hs,s} \samplegrad_{\hs,s}\T \right] - \Exp \left[ \samplegrad_\hs \samplegrad_\hs\T \right] \nonumber \\
    &= (1-\gamma) \Exp_{\hs,s,\as} \left[ \qpolicies(\hs, s, \as)^2 S \right] - (1-\gamma) \Exp_{\hs,\as} \left[ \qpolicies(\hs, \as)^2 S \right] \nonumber \\
    &= (1-\gamma) \Exp_{\hs,\as} \left[ \Exp_{s\mid \hs} \left[ \qpolicies(\hs, s, \as)^2 \right] S \right] - (1-\gamma) \Exp_{\hs,\as} \left[ \Exp_{s\mid \hs} \left[ \qpolicies(\hs, s, \as) \right]^2 S \right] \nonumber \\
    &= (1-\gamma) \Exp_{\hs,\as} \left[ \left( \Exp_{s\mid \hs} \left[ \qpolicies(\hs, s, \as)^2 \right] - \Exp_{s\mid \hs} \left[ \qpolicies(\hs, s, \as) \right]^2 \right) S \right] \nonumber \\
    &= (1-\gamma) \Exp_{\hs,\as} \left[ \Var_{s\mid \hs} \left[ \qpolicies(\hs, s, \as) \right] S \right] \,. \label{eq:covdiff:qhsa:qha}
\end{align}
Because $S$ is positive semi-definite and the variance of any quantity is non-negative, it follows that the matrix in \Cref{eq:covdiff:qhsa:qha} has non-negative diagonal components, which in turn means that $\diag \left( \Cov\left[ \samplegrad_{\hs,s} \right] \right) \succeq \diag \left( \Cov\left[ \samplegrad_\hs \right] \right)$ element-wise.  Further, $\Var\left[ \samplegrad_{\hs,s} \right] = \trace \left( \Cov\left[ \samplegrad_{\hs,s} \right] \right) \ge \trace \left( \Cov\left[ \samplegrad_\hs \right] \right) = \Var\left[ \samplegrad_\hs \right]$.
So, not only is the total variance of $\samplegrad_{\hs,s}$ greater than that of $\samplegrad_\hs$, but the individual variances in any of their dimensions also have the same relationship.
\end{proof}
 
\Cref{thm:var_ghs_gh} reports higher policy gradient variance when using a history-state-based critic compared to a history-based critic.
This is because, for a particular history, the underlying states may vary; hence for updating that particular history, different corresponding history-state pairs are used, for which the history-state value function can learn different values.
Combined with \Cref{thm:variance:qiha:qha}, we get the following corollary relating the variances of the three history-based methods.
\begin{corollary}\label{thm:chained_variance}
    The IACC-HS, IACC-H, and IAC sample gradients have non-monotonically decreasing variance, i.e., $\Var\left[ \samplegrad_{\hs, s} \right] \ge \Var\left[ \samplegrad_\hs \right] \ge \Var\left[ \samplegrad_i \right]$.
\end{corollary}

The history-state-based critic is associated with the largest policy gradient variance, whereas the decentralized history-based critic is associated with the lowest policy gradient variance.
Loosely speaking, the more aligned a critic is to the policy, the less variance there is to its policy gradient.
This theoretical result may be counter-intuitive since a history-state-based critic value function describes a more specific situation, which usually varies less.
However, because the policies are decentralized-history-based, using more specific value functions only increases the variation of values when considering a particular local history.
However, as we shall see in the next section, our theory only tells half of the story.
The three theoretically unbiased critics exhibit different levels of approximation error, where history-state-based critics are often preferable, despite the most policy gradient variance in theory.

\section{Empirical Findings and Discussions}\label{sec:exp_and_dis}

In this section, we present experimental results comparing different types of critics.
We test on a variety of popular research domains including, but not limited to, classical matrix games, the StarCraft Multi-Agent Challenge (SMAC)~\cite{samvelyan19smac}, the Multi-agent Particle Environments (MPE)~\cite{mordatch2017emergence}, and the MARL Environments Compilation~\cite{shuo2019maenvs}.
We provide open-source implementations\footnote{ \texttt{https://github.com/lyu-xg/on-centralized-critics-in-marl}} used for our experiments.
We first discuss, for history-based critics, how higher policy gradient variance in centralized critics affects performance.
We identify several problems caused by the higher policy gradient variance, including exploration and scalability issues.
Second, we show empirical evidence that the unbiased centralized history-based critic is subject to the same independent learning pathologies seen with decentralized critics. 
Third, we highlight the bias and some practical issues with state-based critics.
We also discuss insufficiencies of popular benchmarks, which we identify as a sub-class of Dec-POMDPs.

Finally, we report overall preferable performance for history-state-based critics.
From \Cref{thm:chained_variance}, we have seen that the history-state-based critic has the most variance. However, in the following experiments, we will see that the approach performs exceptionally well due to its unbiased nature (compared to state-based critics) and low approximation error compared to the history-based critic.  
We consider the approximation error as an approximation bias, which will become an essential practical trade-off we discuss in this section.
We explain why despite having the highest theoretical policy gradient variance, the history-state-based critic may often be favorable due to being unbiased and typically having low approximation bias in practice.
In other words, a state-history-based critic is not only theoretically sound but also potentially easier to learn than other critics.

\subsection{Policy Gradient Variance May Affect Performance}
\label{sec:variance_problem}

\begin{figure}
    \centering
    \begin{subfigure}{.32\textwidth}
        \includegraphics[width=\linewidth]{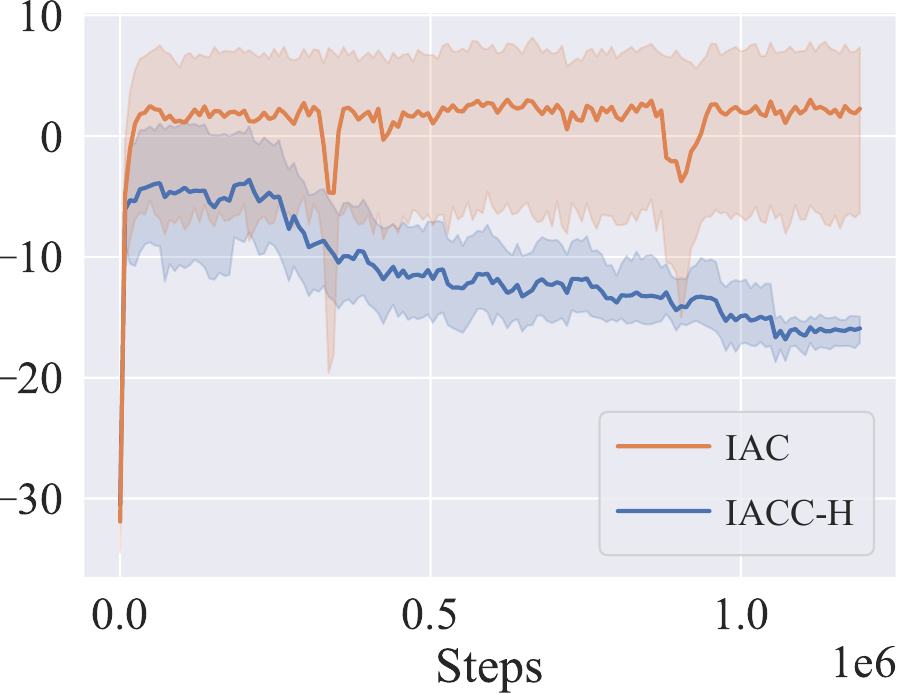}
        \caption{Dec-Tiger.}
        \label{fig:dectiger_results}
    \end{subfigure}
    \hfill
    \begin{subfigure}{.32\textwidth}
        \includegraphics[width=\linewidth]{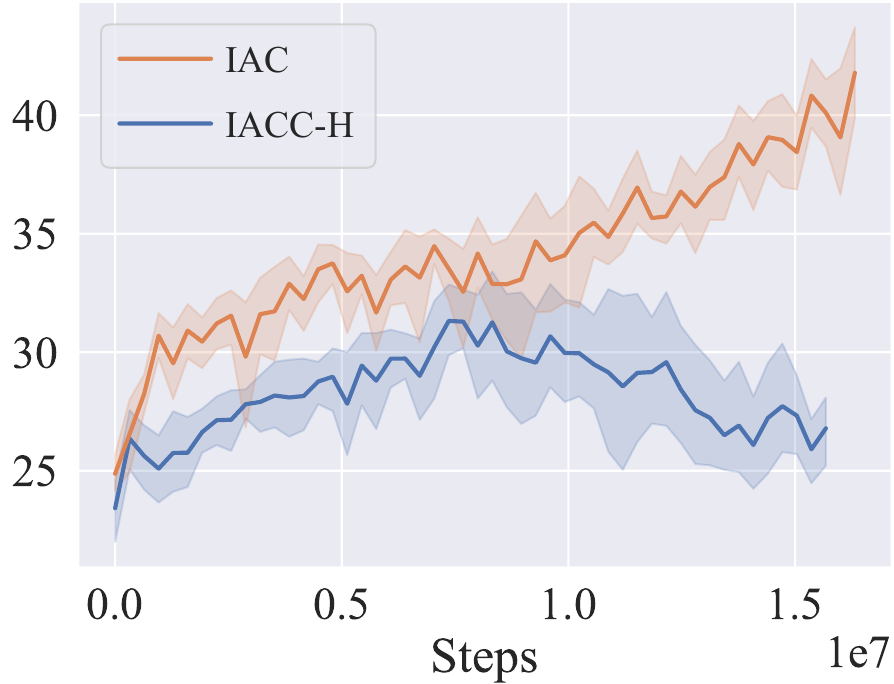}
        \caption{Cleaner}
        \label{fig:cleaner}
    \end{subfigure}
    \hfill
    \begin{subfigure}{.32\textwidth}
        \includegraphics[width=\linewidth]{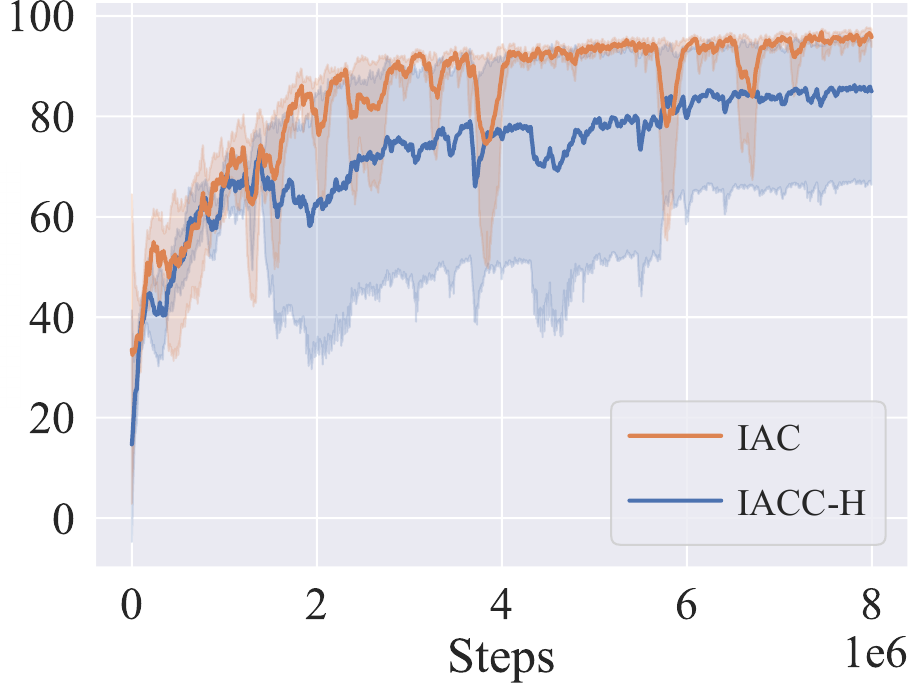}
        \caption{Find Treasure.}
    \end{subfigure}
    \caption{Performances of IAC and IACC-H in different domains (showing mean and standard deviation over 20 runs per method).}
    \label{fig:unstable_policies}
\end{figure}

We first discuss the empirical performance issues of history-based centralized critics (IACC-H) caused by higher policy gradient variance by comparing their performance to decentralized history-based critics (IAC).
Although their performance is often identical in most environments, the higher variance in policy learning signal could lead to unstable policies, which may hinder policy learning for one particular agent, or also affect its teammates as shifts in policy cause non-stationarity of the environment from a local perspective.
In this section, we discuss, with real benchmark examples, some of those impacts on performance and how the policy could become unstable due to additional policy gradient variance.

\begin{figure}
    \centering
    \begin{subfigure}{.32\textwidth}
        \includegraphics[width=\textwidth]{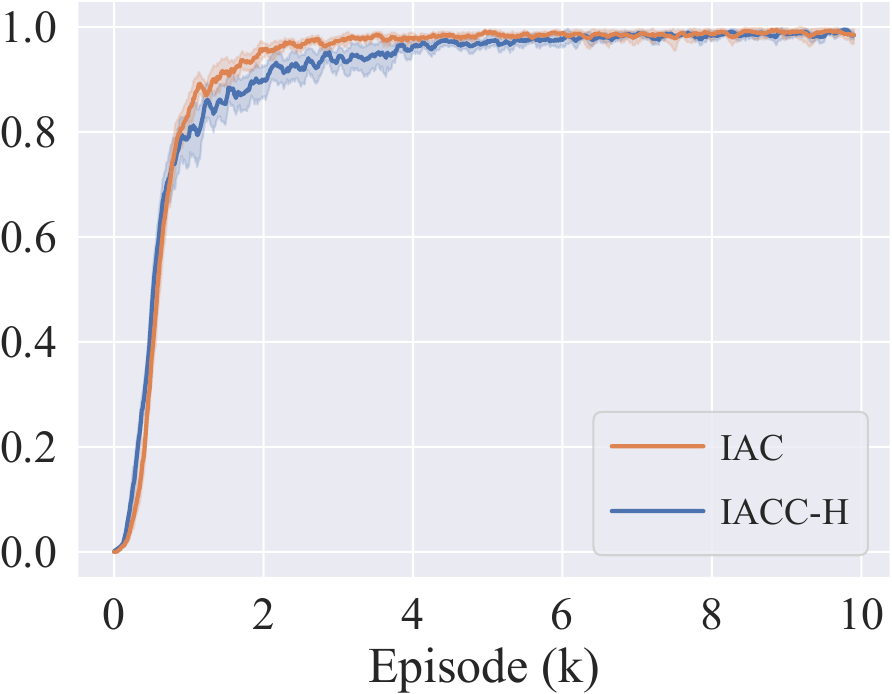}
        \caption{SMAC 3m.}
    \end{subfigure}
    \begin{subfigure}{.32\textwidth}
        \includegraphics[width=\textwidth]{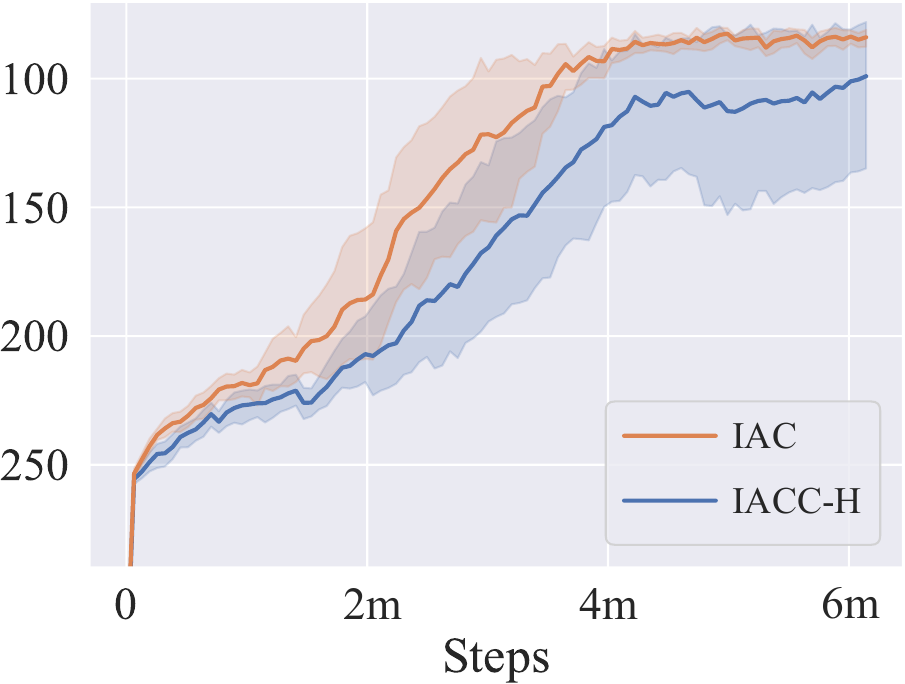}
        \caption{Cross.}
    \end{subfigure}
    \begin{subfigure}{.32\textwidth}
        \includegraphics[width=\textwidth]{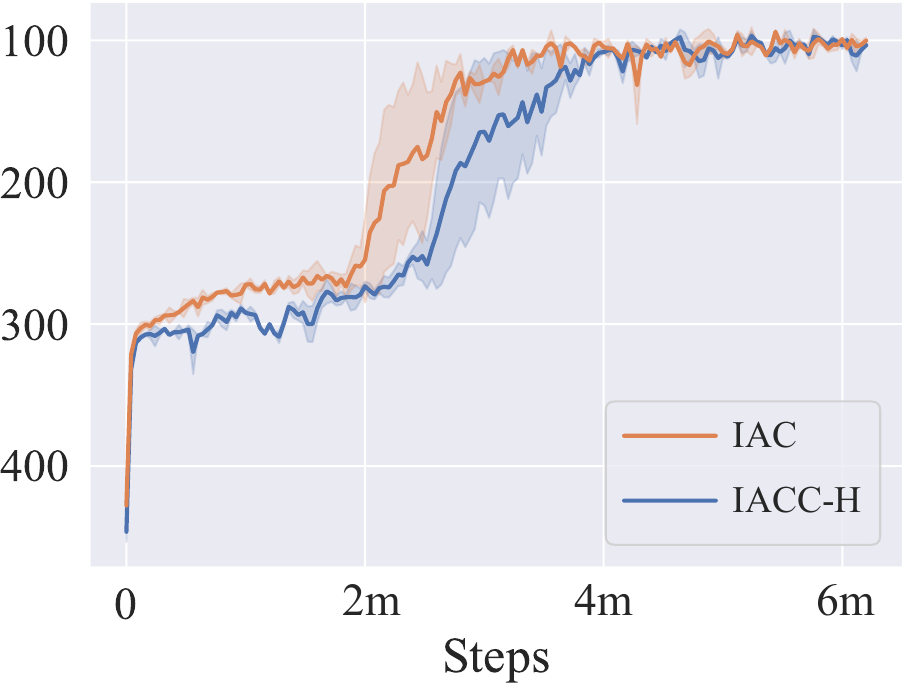}
        \caption{Antipodal.}
    \end{subfigure}
    \caption{Performance comparison in SMAC 3m domain and cooperative navigation domains. In these domains, agents navigate to designated target locations for reward, and are penalized for collisions (showing mean and standard deviation over 20 runs per method).}
    \label{fig:larger_var}
\end{figure}

We report that IAC (independent actor-critic) and IACC-H (an independent actor with centralized history-based critic) usually perform identically
on most tasks~\cite{lyu2021contrasting}, including Go Together~\cite{shuo2019maenvs}, Merge~\cite{CM3}, Predator and Prey~\cite{lowe2017multi}, Capture Target~\cite{omidshafiei2017deep,xiao_corl_2019}, Small Box Pushing and SMAC~\cite{samvelyan19smac} tasks.
IACC-H (with a centralized history-based critic) is less stable in only a few domains shown in \Cref{fig:unstable_policies,fig:larger_var}.
Since the performance of the two history-based critics is similar in most results, we expect that it is because both are unbiased asymptotically (\Cref{thm:unbiased_gradient}).
We observe that, although decentralized critics might be more biased when considering finite training, it does not affect real-world performance in a significant fashion in these domains.
Recent investigations of independent learners such as IPPO~\cite{MAPPO} also suggest similar results where decentralized critics show strong performance.

In cooperative navigation domains Antipodal, Cross~\cite{mordatch2017emergence,CM3}, and Find Treasure~\cite{shuo2019maenvs},
we observe a slightly more pronounced performance difference among runs.
These cooperative navigation domains, shown in \Cref{fig:larger_var},
have no suboptimal equilibria that trap the agents, and on most of the timesteps, the optimal action aligns with the locally greedy action.
Those tasks only require agents to coordinate their actions for a few timesteps to avoid collisions.
Those tasks appear easy to solve, but the observation space is continuous, thus causing large MOV (observation-based variance) in the gradient updates for IACC-H.
Observe that some IACC-H runs struggle to reach the optimal solution robustly while IAC robustly converges, conforming to our scalability discussion regarding large MOV.
A centralized critic induces higher variance for policy updates, where the shifting policies can drag on the value estimates, which, in turn, hinders improving the policies themselves.

We observe similar policy degradation, more specifically the lazy-agent problem, in the Cleaner domain~\cite{shuo2019maenvs}, a grid-world maze in which agents are rewarded for stepping onto novel locations in the maze.
The performance is shown in \Cref{fig:cleaner}.
The optimal policy is to have the agents split up and cover as much ground as possible.
The maze has two non-colliding paths so that as soon as the agents split up, they can follow a locally greedy policy to get an optimal return.
However, with a centralized critic (IACC-H), both agents start to take the longer path with more locations to ``clean.''
When policy-shifting agents are not completely locally greedy, the issue is that they cannot ``clean'' enough ground in their paths.
Subsequently, they discover that having both agents go for the longer path (the lower path) yields a better return, converging to a suboptimal solution.
Again, we see that in IACC-H with a centralized critic, due to the high variance, which we will discuss in \Cref{complexity_and_scalability}, the safer option is favored, resulting in both agents completely ignoring the other path (performance shown in \Cref{fig:cleaner}). 
Overall, high variance in the policy gradient (in the case of the centralized critic) makes the policy more volatile and can indeed result in poor coordination performance in environments that require coordinated series of actions to discover the optimal solution.
The same phenomenon is seen in IPPO~\cite{MAPPO} where centralized critics hinder decentralized policy learning.

\subsubsection{Hindering Exploration}

We test on the classic yet challenging domain Dec-Tiger~\cite{nair2003taming}.
Recall that in Dec-Tiger, to end an episode, each agent has a high-reward action (opening a door with treasure inside) and a high-punishment action (opening a door with a tiger inside).
The treasure and tiger are randomly initialized in each episode,
hence, a third action (\textit{listen}) gathers noisy information regarding which of the two doors is the rewarding one.
The multi-agent extension of Tiger requires two agents to open the correct door simultaneously to gain maximum return.
Conversely, for simultaneous undesirable actions, the agents take less punishment.
Note that any fast-changing decentralized policies are less likely to coordinate the simultaneous actions with high probability, thus lowering return estimates for the critic and hindering joint policy improvement.
As expected, we see in \Cref{fig:dectiger_results} that IACC-H (with a centralized critic and higher policy gradient variance) does not perform as well as IAC.
In the end, the IACC-H agent learns to avoid high punishment (agents simultaneously open different doors, $-100$) by not gathering any information (\textit{listen}) and opening an agreed-upon door on the first timestep.
That is, IACC-H gives up completely on high returns (where both agents \textit{listen} for some timesteps and open the correct door at the same time, $+20$) because the unstable policies make coordinating a high return of $+20$ extremely unlikely.
IAC, on the other hand, has comparatively reduced variance in policy learning, so the environment (including other agents) from the local perspective is more stable, and the agent can thus find a good response policy for adequate performance and cooperation.

\subsubsection{Scalability}
\label{complexity_and_scalability}

Another critical consideration for critic design is the scale of the task.
A centralized history-based critic's feature representation needs to scale linearly (in the best case) or exponentially (in the worse case) with the number of agents.
In contrast, decentralized history-based critics' features can remain constant, and in homogeneous-agent systems, decentralized history-based critics can even share parameters to leverage the problem symmetry.
Also, some environments may only require an agent to reason a small amount about the other agents.
For example, in environments where agents' decisions rarely depend on other agents' trajectories, the gain of learning joint value functions is likely minimal. Therefore, we expect decentralized critics to perform better while having better sample efficiency in those domains.

\begin{figure}[t]
    \centering
    \begin{subfigure}{.40\textwidth}
        \includegraphics[width=\textwidth]{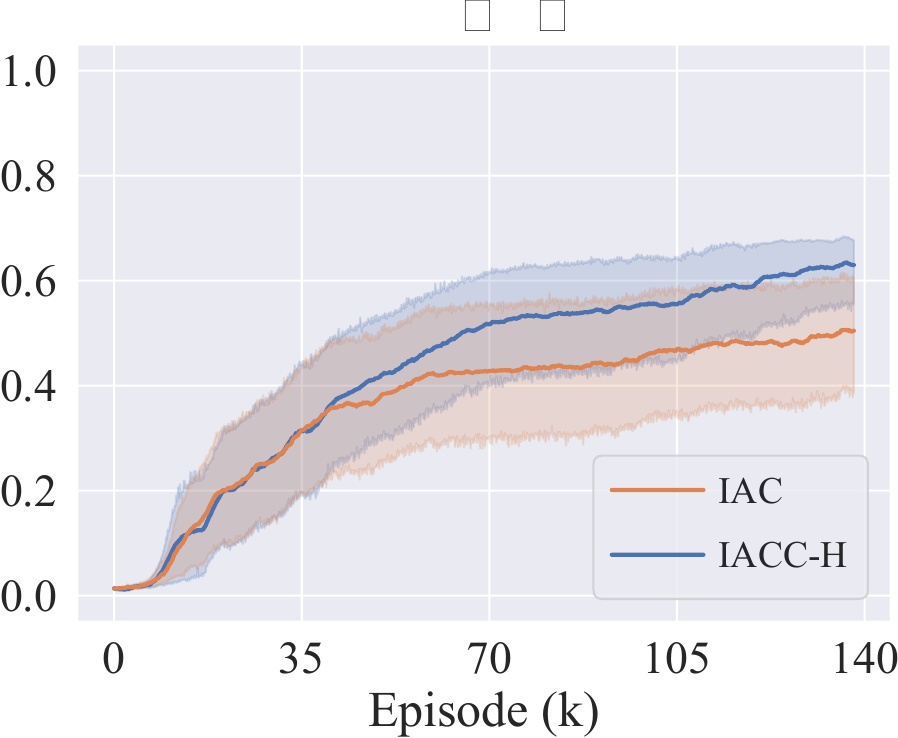}
        \caption{Size 6x6.}
        \label{}
    \end{subfigure}
    \begin{subfigure}{.40\textwidth}
        \includegraphics[width=\textwidth]{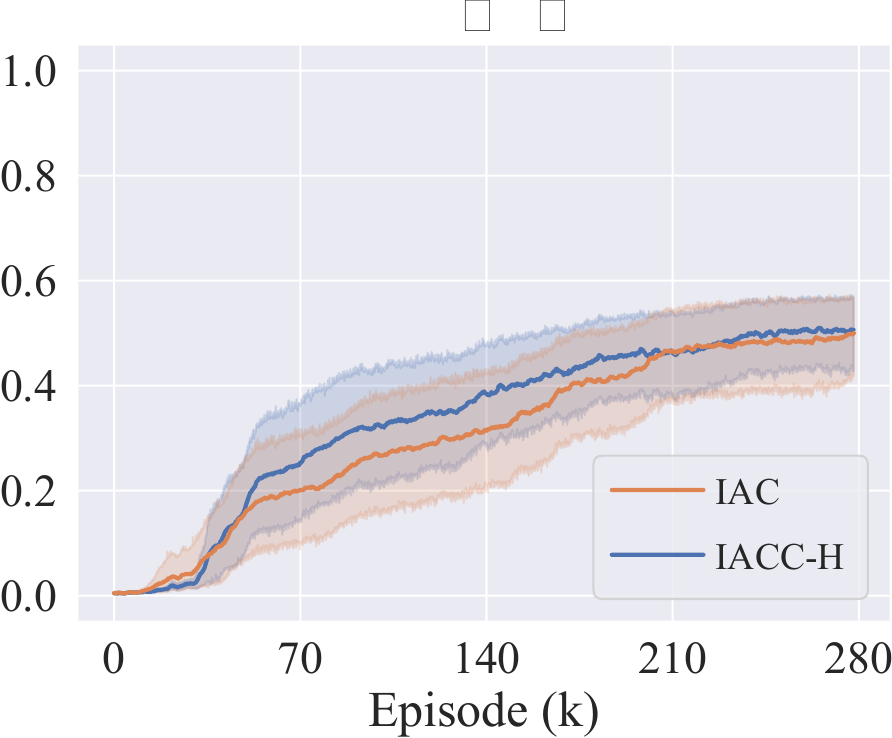}
        \caption{Size 8x8.}
        \label{}
    \end{subfigure}
    \begin{subfigure}{.40\textwidth}
        \includegraphics[width=\textwidth]{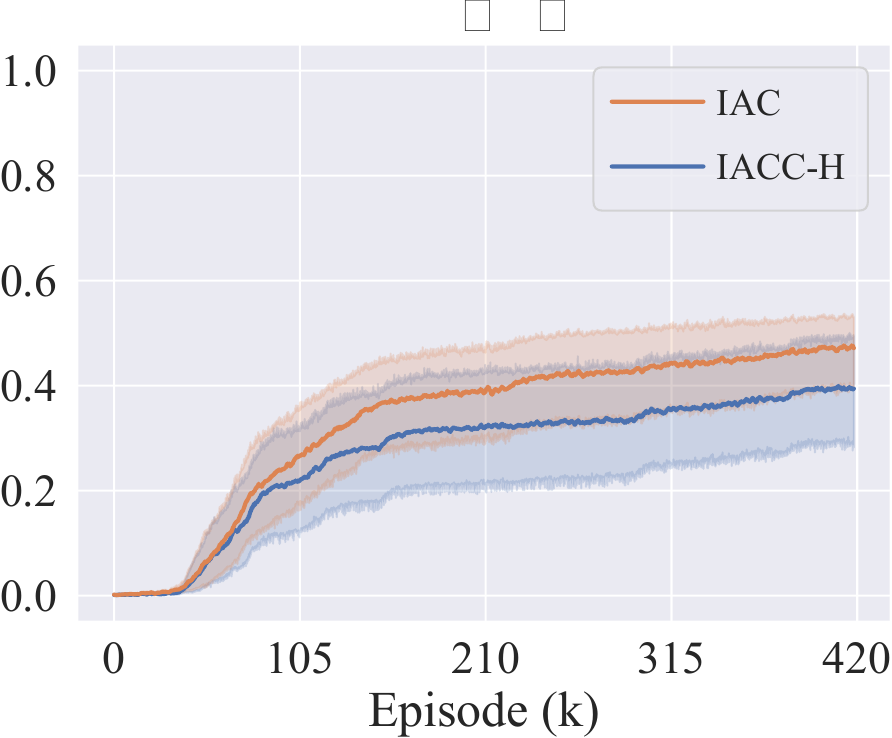}
        \caption{Size 10x10.}
        \label{}
    \end{subfigure}
    \begin{subfigure}{.40\textwidth}
        \includegraphics[width=\textwidth]{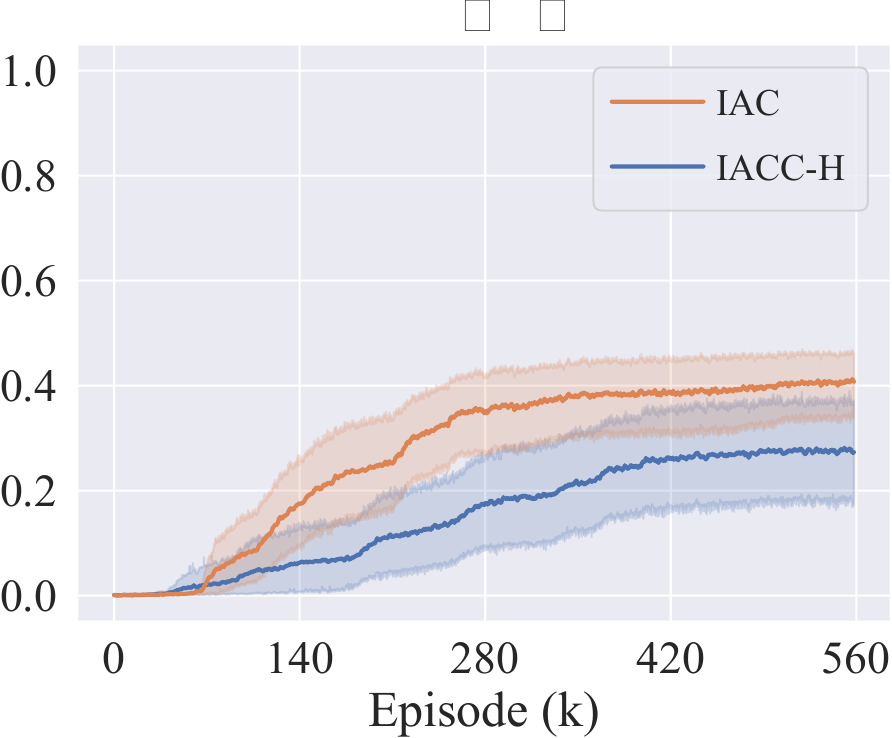}
        \caption{Size 12x12.}
        \label{}
    \end{subfigure}
    \caption{Evaluation on Capture Target domain of IAC and IACC-H (showing mean and standard deviation over 20 runs per method).}
    \label{fig:capture_target_results}
\end{figure}

To empirically test our hypothesis, we find environments where we can artificially increase the observation space to highlight the scalability issue.
In Capture Target~\cite{lyu2020likelihood}
where agents are rewarded by simultaneously catching a moving target in a grid world (\Cref{fig:capture_target_results}),
by increasing the grid size from $6\times6$ to $12\times12$, we see a notable comparative drop in overall performance for the centralized critic approach, IACC-H (\Cref{fig:capture_target_results}).
Since an increase in observation space leads to an increase in Multi-Observation Variance (MOV), it indicates that here the policies of IACC-H do not handle MOV as well as the decentralized critics in IAC.
The result may generalize that, for large environments with neural networks, decentralized critics scale better in the face of MOV because they do not involve MOV in policy learning.

The impact of variance will also change as the number of agents increases.
In particular, when learning stochastic policies with a centralized critic in IACC-H,
the potential variance in the policy gradient also scales with the number of agents combined with exponentially growing action and observation spaces.
On the other hand, IAC's decentralized critics potentially have less stable learning targets in critic bootstrapping as the number of agents increases, but the policy updates still have low variance.
Therefore, scalability remains an issue for history-based critics, and the actual performance is likely to depend on the domain, function approximation setup, and other factors.
Therefore, IAC should be a better starting point due to more stable policy updates and potentially shared parameters.
For state-based critics, although without an explosive feature space, the variance issue similar to that of IACC-H remains (and the issue of bias may cause additional problems as discussed below).

  \subsection{Cooperation Pathologies}
  We now investigate how cooperation problems arise with different critic choices.
  We point out that centralized history-based critic is subjected to the same cooperation issues as their decentralized counterparts.

  
  Move Box~\cite{shuo2019maenvs} is another commonly used domain where grid-world agents are rewarded by pushing a heavy box (requiring both agents) onto any of two target locations.
  The farther destination gives a $+100$ reward while the nearer destination gives $+10$.
  Naturally, the optimal policy is for both agents to go for $+100$, but if \textit{either} of the agents is unwilling to do so, this optimal option is ``shadowed'', and \textit{both} agents will have to go for $+10$.
  We see in \Cref{fig:exps1}(d) that both methods (IAC and IACC-H) fall for the \textit{shadowed equilibrium} (mentioned in \Cref{sec:climb_game_example}), favoring the safe but less rewarding option.
  
  Analogous to the Climb Game, even if the centralized critic is able to learn the optimal values due to its unbiased on-policy nature,
  the on-policy return of the optimal option is inadequate due to an uncooperative teammate policy;
  thus, the optimal actions are rarely sampled when updating the policies.
  The same applies to both agents, so the system reaches a suboptimal equilibrium, even though IACC-H is trained in a centralized manner.
  As a result, same as in Climb Game, Cleaner, and Dec-Tiger, we would not expect centralized critics to help in learning cooperative policies; this connects with \Cref{thm:unbiased_gradient} where IAC and IACC-H have the same expected policy gradient.

  \subsection{Analysis of State-Based Critics}
  We now discuss how the bias introduced by state-based critics does not, as one may hope, bias the policies toward a higher-quality solution but often pushes the policies toward suboptimality with limited examples of improved performance. 
  
  We show that it cannot facilitate cooperative behavior any better than IAC in our experiments.
  For example, we consider a multi-agent recycling task with results shown in~\Cref{fig:exps1}. 
  In the multi-agent recycling task, we expect the agents to work together while maintaining battery levels.
  The task contains recyclable small and large targets, and the large ones require two agents to act simultaneously.
  The agents only observe their own battery levels.
  For policies that recycle large targets to be competitive against small-target-policies (reactive), agents must estimate their teammate's battery level based on their observation history (non-reactive).
  
  \Cref{fig:exps1} shows performance for Recycling agents with various critics, in which none of the methods learns to recycle large targets.
  Note that a reactive policy cannot achieve optimal performance, which suggests that the methods all fail to cooperate (more discussion in \Cref{sec:subclass}).
  Learning to cooperate in this case is especially difficult because the agents suffer the problem of shadowed equilibrium~\cite{matignon2012independent} despite the usage of a centralized critic (as we mentioned in the previous subsection and in \Cref{sec:climb_game_example}).
  Therefore, we observe that state-based critics cannot bias policies in a significant and specific way such that it learns to cooperate.
  As a result, using state-based critic methods are still subjected to shadowed equilibrium and other cooperation issues.
  
  Interestingly, in the Move Box domain (\Cref{fig:exps1}), the state-based critic, being potentially biased, managed to escape the action shadowing Nash equilibrium and managed to obtain higher rewards by moving the box to the farther destination (Pareto optimal).
  It suggests that in practice, bias and variance may, in special cases, lead to improved performance, such as helping in breaking out of suboptimal local solutions. 
  
\begin{figure}[t]
    \centering
    \begin{subfigure}{.32\textwidth}
        \includegraphics[width=\textwidth]{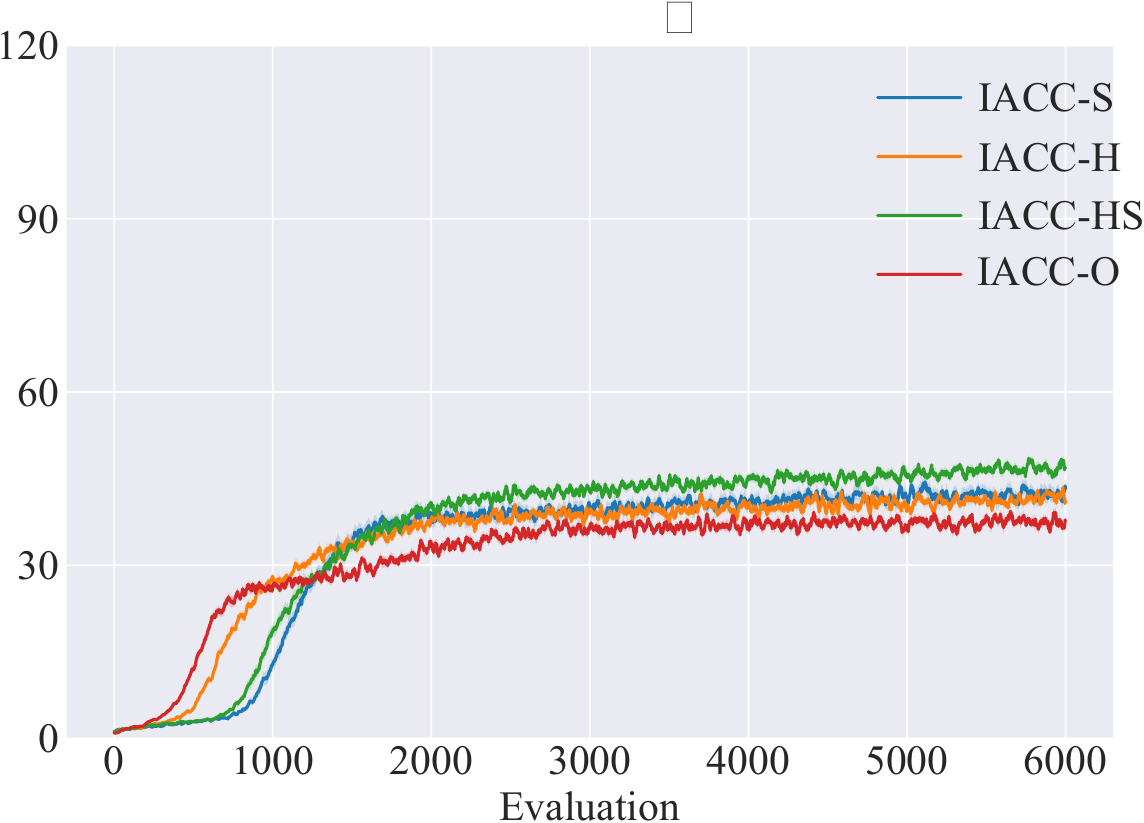}
        \caption{Observation radius 4.}
    \end{subfigure}
    \begin{subfigure}{.32\textwidth}
        \includegraphics[width=\textwidth]{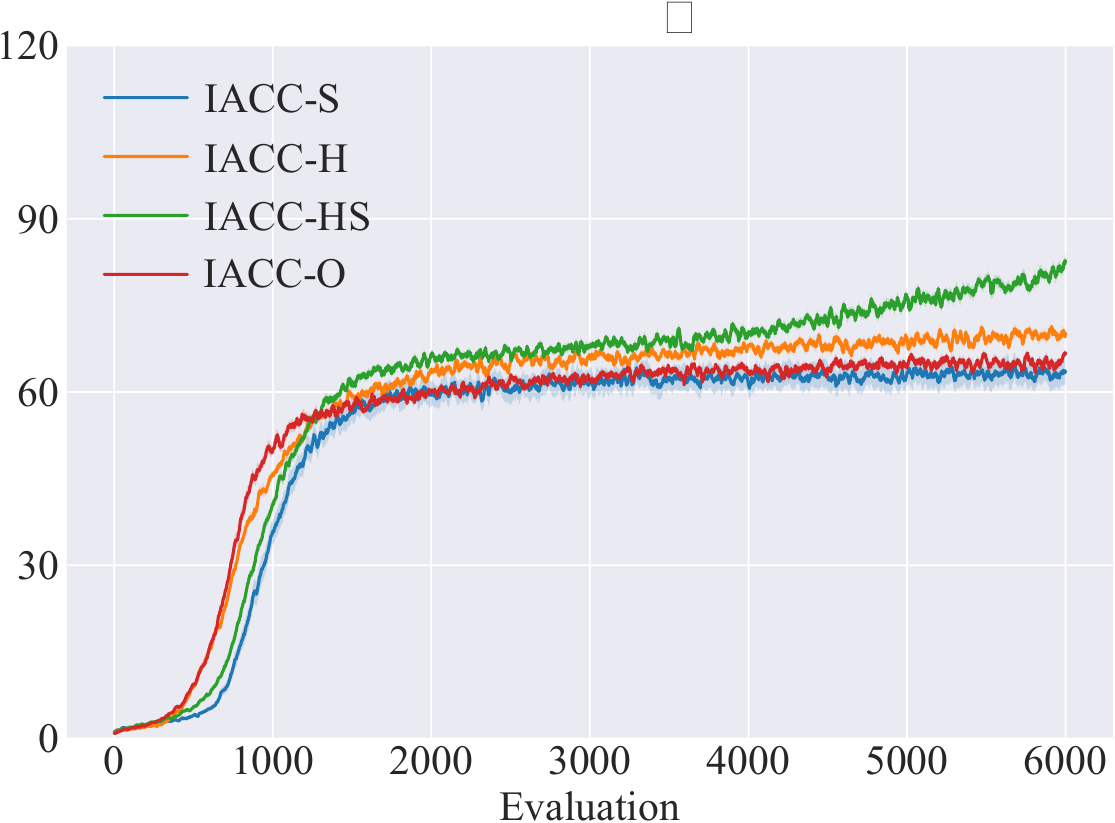}
        \caption{Observation radius 8.}
    \end{subfigure}
    \begin{subfigure}{.32\textwidth}
        \includegraphics[width=\textwidth]{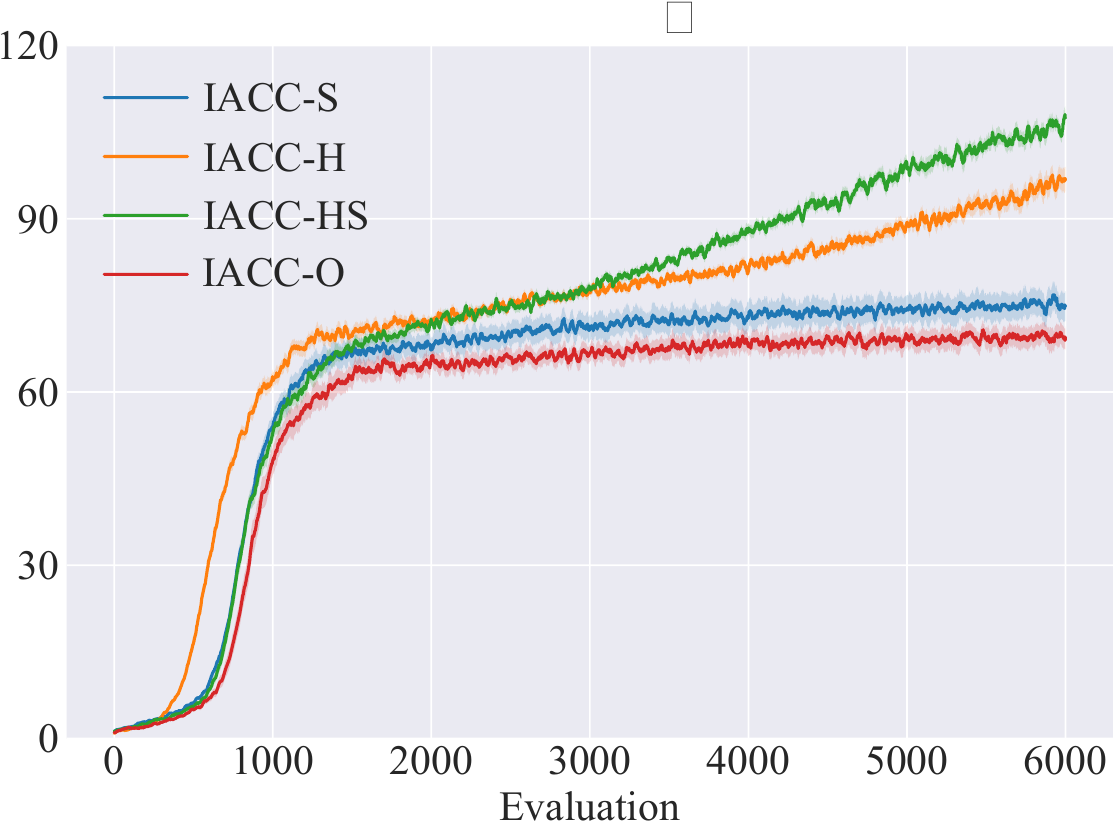}
        \caption{Observation radius 16.}
    \end{subfigure}
    \caption{Performance comparison in Predator and Prey, showing higher performance of history-state-based critic; that is, highlighting deficiencies of using only state or history-based critic. The results also suggest that the state-based critic can be asymptotically biased thus hindering the convergence of decentralized policies (showing mean and standard deviation over 20 runs per method).}
      \label{fig:predator_and_prey}
  \end{figure}

\subsubsection{Special Cases of Dec-POMDPs}
\label{sec:subclass}

\begin{figure}[t]
    \centering
    \begin{subfigure}{.32\textwidth}
        \includegraphics[width=\textwidth]{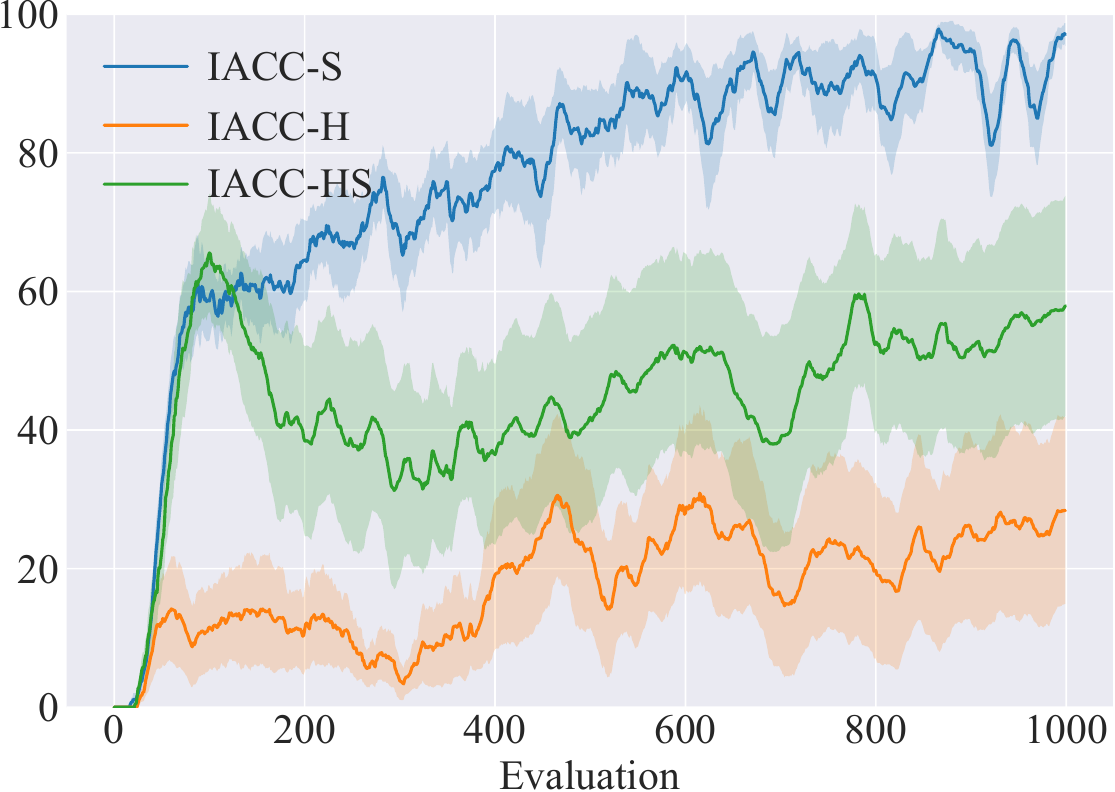}
        \caption{SMAC 2s\_vs\_1sc.}
    \end{subfigure}
    \begin{subfigure}{.32\textwidth}
        \includegraphics[width=\textwidth]{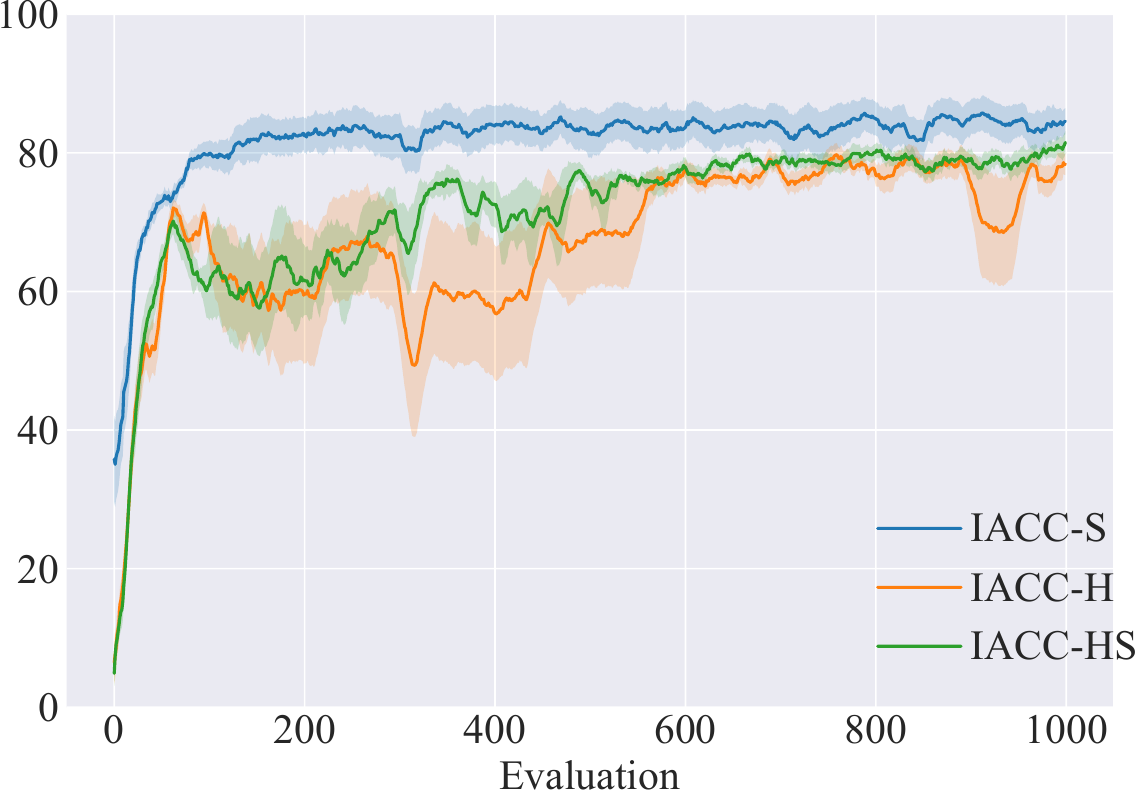} 
        \caption{SMAC 3m.}
    \end{subfigure}
    \begin{subfigure}{.32\textwidth}
        \includegraphics[width=\textwidth]{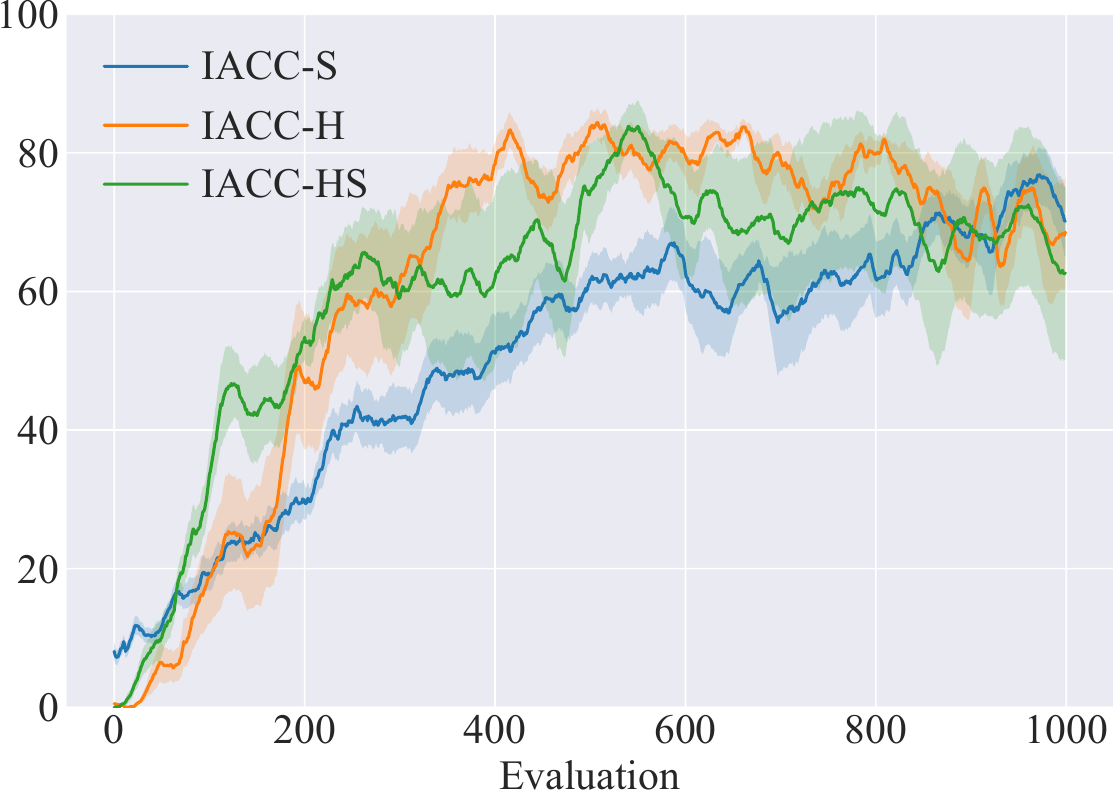}
        \caption{SMAC 2s3z.}
    \end{subfigure}
    \begin{subfigure}{.32\textwidth}
        \includegraphics[width=\textwidth]{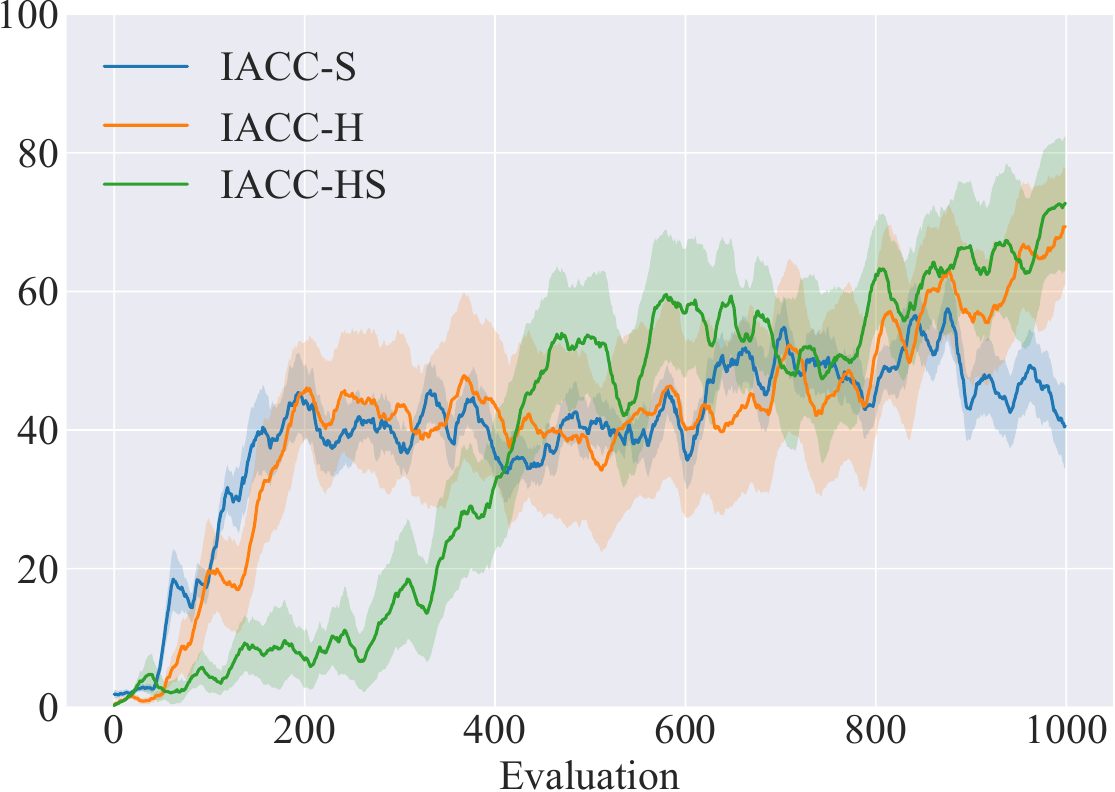}
        \caption{SMAC 1c3s5z.}
    \end{subfigure}
    \begin{subfigure}{.32\textwidth}
        \includegraphics[width=\textwidth]{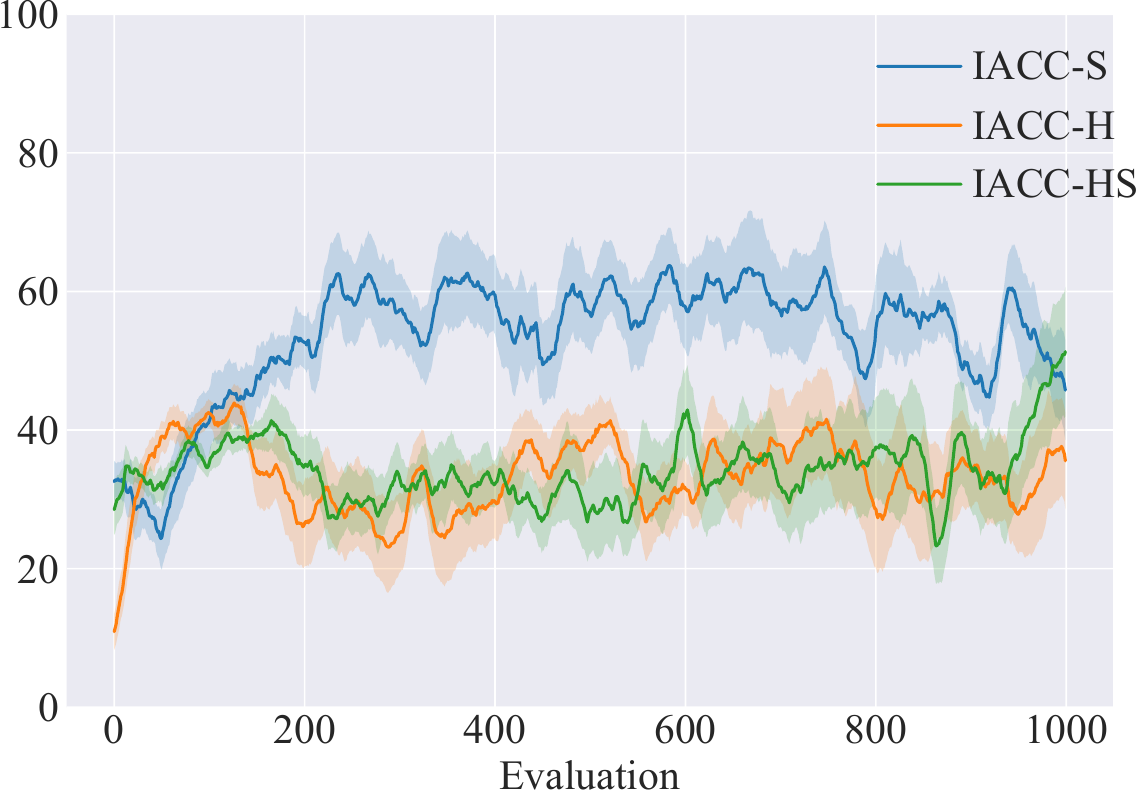}
        \caption{SMAC bane\_vs\_bane.}
    \end{subfigure}
    \caption{Performance comparison of COMA with different critics in multiple SMAC scenarios (showing mean and standard deviation over 20 runs per variant).}
    \label{fig:smac}
\end{figure}

\begin{figure}[t]
    \centering
    \begin{subfigure}{.49\textwidth}
        \includegraphics[width=\textwidth]{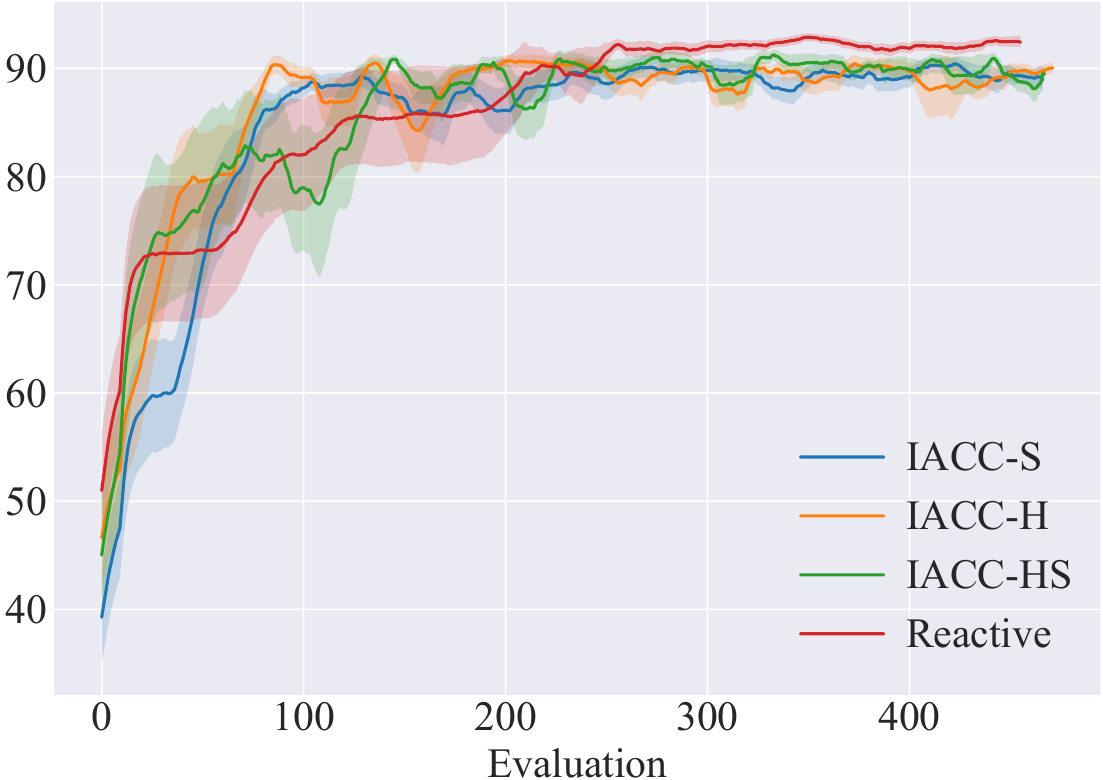}
        \caption{Meeting-in-a-Grid~\cite{amato2009incremental}.}
    \end{subfigure}
    \begin{subfigure}{.49\textwidth}
        \includegraphics[width=\textwidth]{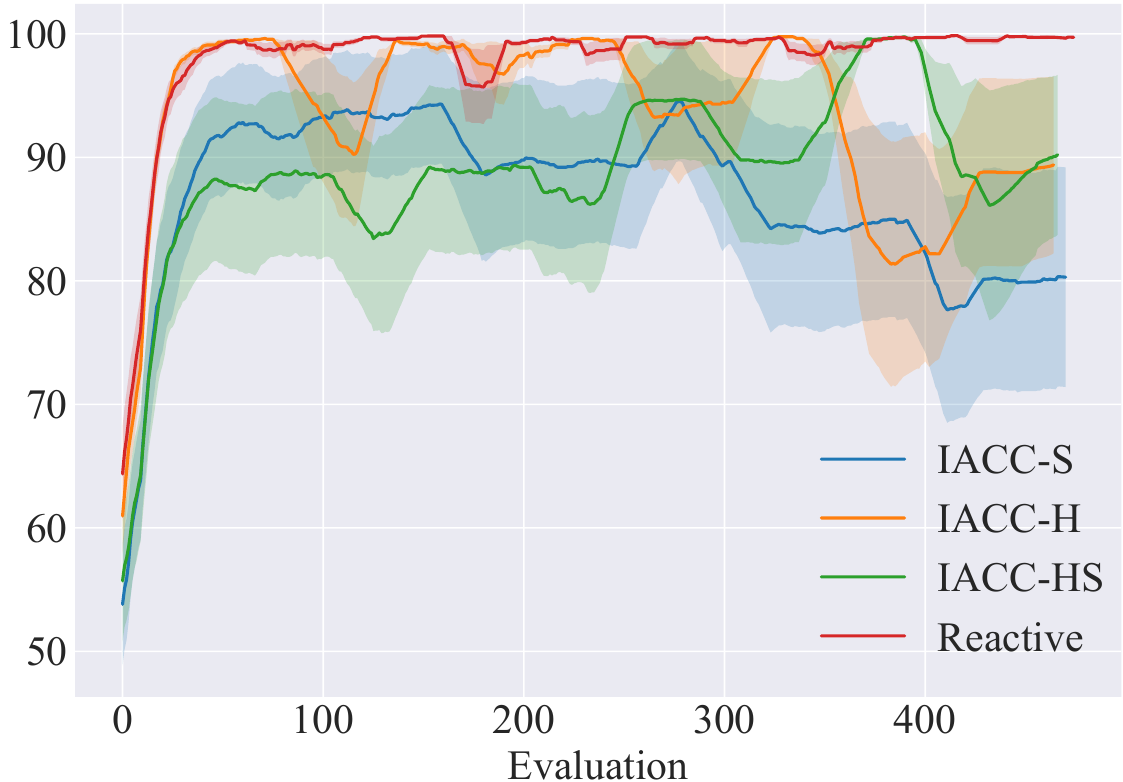}
        \caption{Find Treasure~\cite{shuo2019maenvs}.}
    \end{subfigure}
    \begin{subfigure}{.49\textwidth}
        \includegraphics[width=\textwidth]{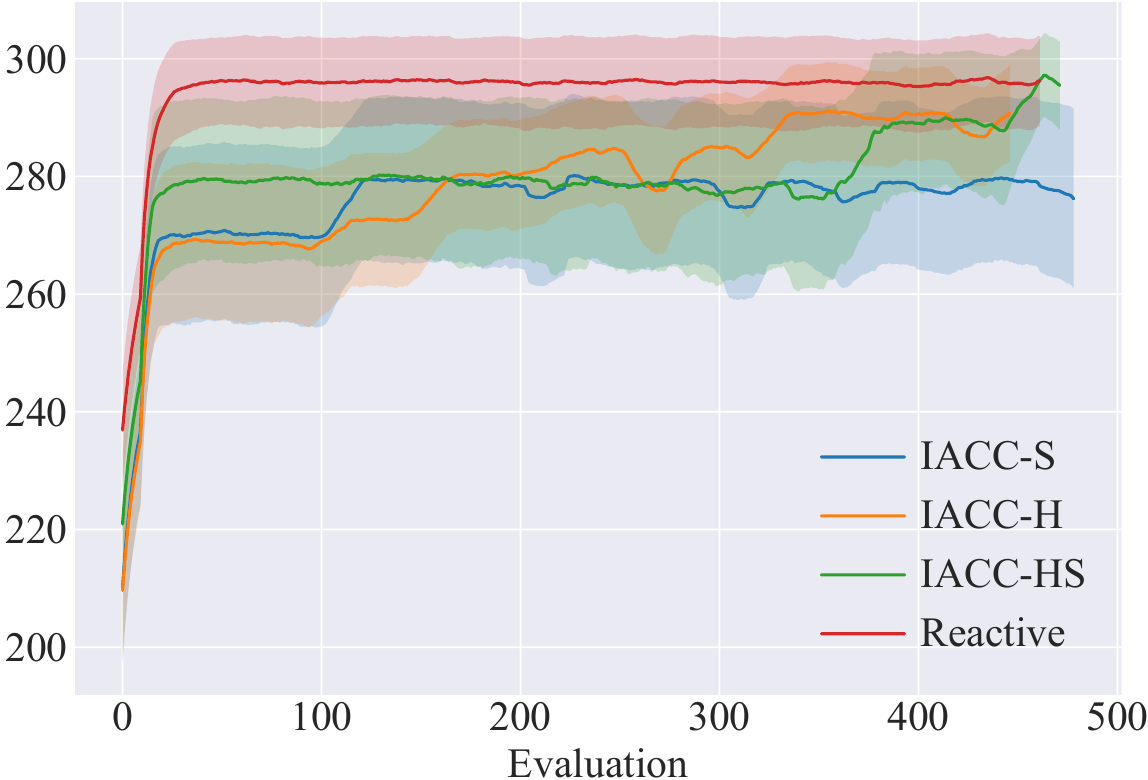}
        \caption{Recycling~\cite{amato2007optimizing}.}
    \end{subfigure}
    \begin{subfigure}{.49\textwidth}
        \includegraphics[width=\textwidth]{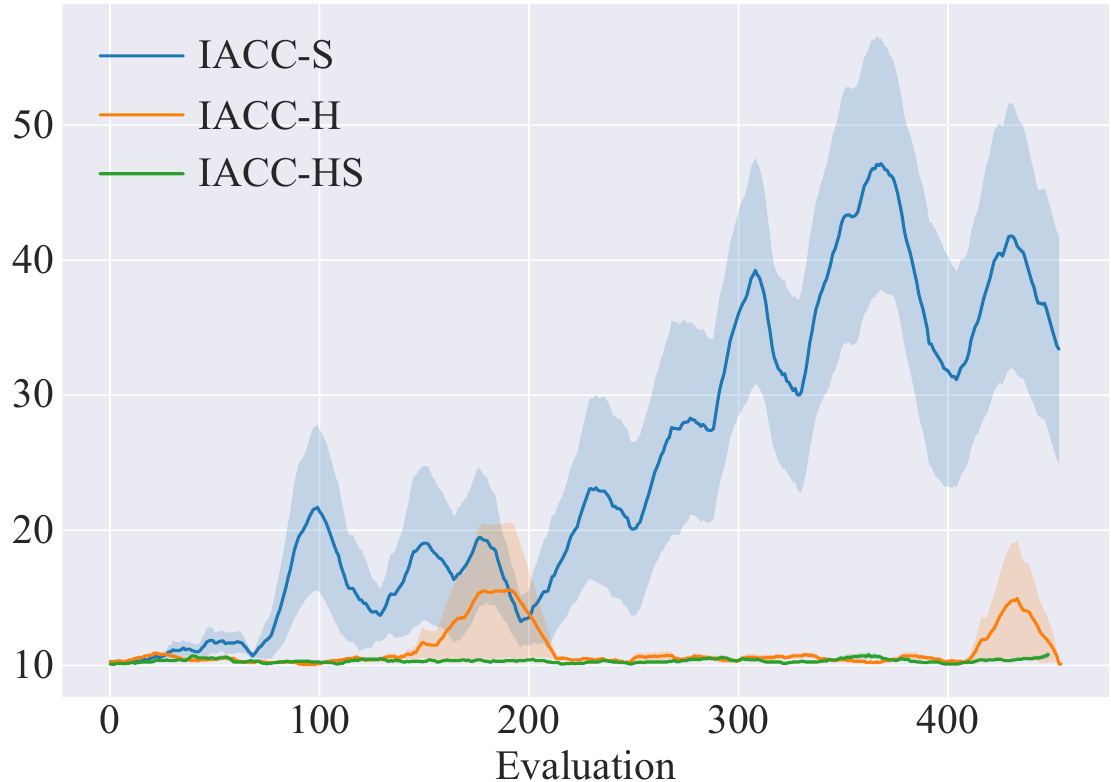}
        \caption{Move Box~\cite{shuo2019maenvs}.}
    \end{subfigure}
    \caption{Performance comparison of IACC-S, IACC-H and IACC-HS in cooperative Dec-POMDPs (showing mean and standard deviation over 20 runs per method).}
    \label{fig:exps1}
\end{figure}

We now investigate the usage of state-based critics, which can be biased in theory (\Cref{sec:state-based-critic}) in a sub-class of Dec-POMDPs. In particular, 
we identify a special sub-class of Dec-POMDPs, which is surprisingly common in practice, in which the state-based critic is not critically (or at all) biased. 
Some environments give partial yet ``sufficient'' local information, in other words, a decent policy does not depend on the entire history and is not required to retain information gathered from the past.
In that case, the observation value and history values are mostly (or exactly) value-equivalent. That is, more information from the full observation history does not add to the value expectation over using just the last observation; in some tasks, reactive policies (policies that only condition on the last observation) can achieve optimal performance (see reactive policies' performance in \Cref{fig:exps1}).
For example, in the Meeting-in-a-Grid domains~\cite{bernstein2005bounded,amato2009incremental}, agents observe their own location but not the teammate while trying to meet in a grid-world.
Optimally, agents would navigate to a predetermined spot (e.g., the center) and wait.
Another example is Find Treasure~\cite{shuo2019maenvs} in which one agent has to step onto a trigger location to open a door while the other agent goes through the door to find treasure.
Again, even though the task is clearly partially observable, and the agents only observe local information, the optimal policy does not require remembering history information because policies do not benefit (in expected returns) from the additional observation history.
In these environments, one may safely assume that a given state will produce similar return distributions with different histories because the history information does not meaningfully affect the policy and the return distributions
since the policy conditions on the last observation, which is produced by the state.
However, we note that in harder and more partially-observable tasks, we expect that the optimal policies are not reactive.
 Environments that require more than a reactive policy require the agent to collect information and retain information over time.
If such information gathering is needed, then we can conclude that the state-based critics are, in theory, biased, even assuming no approximation error.

\paragraph{A Subclass of Observation Models}
Depending on the type of observation model in the task,
we also identify acceptable or preferred tasks to use state-based critics, which we call an Agent-Centered Observation Model (ACOM).
ACOM can usually indicate where reactive policies may be optimal or near-optimal, thus making state-based critics unbiased or nearly unbiased.
Therefore, we also see tasks where it is acceptable or even preferred to use state-based critics.
A typical example is the (version 1) StarCraft Multi-Agent Challenge (SMAC) benchmark~\cite{samvelyan19smac}. As seen in \Cref{fig:smac}, most scenarios show that state-based critics exhibit the best overall performance.
In SMAC, each agent controls a combat unit to engage enemy units, and the state includes the status of friendly agents and opponent units. In contrast, the observations for a given agent are the status of itself and the units whose distances are within a specific range.
It uses a deterministic observation model, and in some SMAC tasks, we find that information about far-away units rarely affects an agent's decisions (in a value-equivalent sense).
In other words, the occluded information contributes little to the expected values.
Thus, the history values associated with a state are usually value-equivalent; as a result, using a state-based critic does not introduce significant bias in this case.
In specific scenarios of SMAC, shown in \Cref{fig:smac}, we report that state-based critics perform better than history-based ones.
Since both the state and the observation have similar and concise representations, while the history will contain redundant information,
it is reasonable to expect that state-based critics may give a more reliable estimate for policy training. In theory, the agents will learn to ignore the redundant history information, which may be time-consuming or difficult to do in practice.
\Cref{fig:smac} uses the original COMA~\cite{foerster:aaai18} paired with a different type of critics; the high-performing state-based critic shows why subsequent centralized critics literature tends to inherit state-based critics from COMA.
Compared to COMA, vanilla multi-agent actor-critic methods also show similar trends~\cite{lyu2021contrasting}.

%
We note that using reactive policies is insufficient to make state-based critics unbiased in the general case because at any time step, the observation produced by the state may still cause the policy to behave differently (\Cref{eq:bias-proof:1}).
However, if the environment is almost fully observable, where reactive policies can achieve high performance, then the bias from state-based critic would likely be minimal.
It is precisely why MADDPG is not biased in their particle environments~\cite{lowe2017multi}.
This situation is also not uncommon, which we see in numerous benchmarks, including the classic task Recycling and environments with radius-based observation models such as SMAC.
We hence note that the benchmarks used in recent state-of-the-art works usually have the aforementioned deterministic observation model with limited partial observability, which is, in fact, a special case of imperfect information.
 
\subsection{The Empirical Bias-Variance Trade-Off}
\label{sec:real_learning_rate}
Recall that our theoretical analysis is based on the assumption of real value functions, or converged and accurate critic models, and the theory does not strictly hold for learned critics, as those models may inherently carry arbitrary bias, which we call approximation bias.
However, we still expect the theorems to provide valuable insights that broadly apply to learned critics as well.
In those cases, the approximation bias associated with the real-life learned critics should be considered an additional form of bias on top of the biases (if any) already discussed for the pure value functions in our theories.
In practice, as we have seen, value functions are difficult to learn.
It is generally more difficult to learn history value functions, especially decentralized history value functions, compared to state or history-state value functions.
Therefore, we come to an important observation that, the methods that are more preferable for policy learning in theory have more approximation bias.
For example, IAC enjoys the best theoretical policy gradient properties but its decentralized history-based values are the most difficult to learn.
This is why some of our theories may seem counter-intuitive, and it is a necessary trade-off to consider when designing and selecting an algorithm for different tasks.

In MARL, the on-policy value is non-stationary since the return distributions depend heavily on the current set of policies.
When policies update and change, the learned critic becomes at least partially (if not entirely) obsolete, and biased toward the historical policy-induced return distribution.
Bootstrapping from outdated values creates additional bias on top of the modeling bias inherent to the model itself. Compounding biases may lead to the divergence of the decentralized policies depending on the learning rate and how drastically the policy changes.

These non-stationarity issues affect all critic approaches since they are all on-policy estimates.
However, centralized value function learning is generally better equipped, and state values are generally easier to learn due to their commonly simple representations (compared to histories).
Therefore, bootstrapping can be more stable in the case of a centralized critic, especially a centralized state-based critic. 
As a result, in a cooperative environment with a moderate number of agents, we expect a centralized critic would learn more stably and be less biased (in the actual policy gradient), perhaps counteracting the effect of having a larger variance in the policy gradient, or the case state-based critics, some inherent bias.

Combining our discussions of the variance problems from \Cref{sec:variance_problem}, we conclude that whether to use critic centralization can be essentially considered a bias-variance trade-off decision.
More specifically, it is a trade-off between the variance of policy updates and the bias of the critic model:
a centralized history critic should have a lower bias because it will have more stable values that can be updated straightforwardly when policies change, but higher variance because the policy updates need to be averaged over (potentially many) other agents;
an easy-to-learn state-based critic may have even less practical bias, especially in certain ACOM environments;
a history-state-based critic can further reduce the policy gradient bias in practice but at the cost of higher policy gradient variance (\Cref{thm:chained_variance}).
In short, the more policy gradient bias we reduce in practice, the more we pay in variance.
Regardless, we note that the centralized critic likely faces more severe scalability issues in not only critic learning but also in policy gradient variance. 
As a result, we cannot expect one method will always dominate the other in terms of performance.

We also identify scalability issues with both centralized and decentralized critics.
We observe that the exponential action and history spaces that come with a scaling set of agents significantly increase the centralized critics' policy gradient variance, not to mention that larger models are needed for the factorized spaces.
The latter problem can be mitigated by learning a state-based critic if, as identified earlier, we can avoid its bias for the given environment.
On the other hand, for decentralized critics, a larger set of agents adds difficulty in value learning due to larger learning target variance for the decentralized critics themselves, a trade-off requiring consideration for each task.

\subsubsection{The Safe Option: History-State-Based Critics}
\label{sec:state_history_based_critics}
We test IACC-HS by concatenating the state and history without changing other aspects of implementation.
Unlike IACC-S, IACC-HS is unbiased (\Cref{thm:gradient:ha:hsa}) even with large amounts of partial observability. 
However, IACC-HS is an extreme on the bias-variance trade-off scale in the sense that it should have the least policy gradient bias in practice but the most variance (\Cref{thm:chained_variance}) in theory.
We report that for most environments, we obtain similar performance compared to simple history or state-based critics, but for tasks that are more complicated or partially observable, IACC-HS usually outperforms others.
Therefore, it is the critic option most widely adaptable to different environments and ought to be the default choice for a task whose nature is unclear.
For example, in the Predator and Prey domain in \Cref{fig:predator_and_prey}, we see that the history-state-based critics have a clear advantage over critics that use state or history information alone, especially in the later stages of training.
It suggests that there exist situations where the history-state-based critic can benefit from the advantages of both the state as well as the history as discussed in \Cref{sec:subclass}.
In general, both Box Pushing and Cleaner are grid-world tasks with observations local to the cells around the agent.
However, the agent needs to estimate their teammates' locations or paths to act optimally; this makes a purely state-based critic biased as it cannot encourage information gathering.
On the other hand, by incorporating histories, the history-state-based critic is not biased while still having the state to help with value learning.

\section{Conclusions}

This paper summarizes, standardizes, and analyzes an exhaustive set of centralized critics in multi-agent reinforcement learning.
We provide a theoretical basis with bias and variance analysis for each critic variant.
Specifically, we prove convergence and policy gradient equivalence between centralized and decentralized history-based critics.
We show that history-based centralized critics have higher policy gradient variance in theory than their decentralized counterparts.
The implication of this theoretical result is crucial in understanding that the centralized critics not only cannot meaningfully mitigate cooperation issues but also may harm performance due to its higher policy gradient variance.

For the commonly used state-based critic, we prove that it may produce biased policy gradients compared to the history-based critics. As a result, state-based critics are not theoretically sound and may lead to poor performance. 
We also give an analysis of specific environment features, which may remove or reduce said biases and explains impressive performance on certain benchmark domains. 

We also provide comprehensive practical insights and advice on top of the theoretical results.
We argue that the most prominent trade-off is between policy gradient bias and variance; specifically, the higher the policy gradient variance, the less policy gradient bias occurs practically during training, making none of the methods a clear winner on all tasks.
However, in general, we recommend the robust IACC-HS (history-state-based critic) as the first choice for a generic or unfamiliar task.
The history-state-based critic has the highest theoretical policy gradient variance but often yields favorable performance due to its ease of value learning and unbiased policy gradient. 


\section{Acknowledgements}
This research is supported by the U.S.~Office of Naval Research award N00014-19-1-2131, Army Research Office award W911NF20-1-0265 and NSF awards 1816382 and 2044993.

\appendix

\section{Problems and Solutions of $\Pr(\as\mid s)$ and $\qpolicies(s, \as)$}
\label{sec:pr_as_s}

In \Cref{sec:value_functions:s}, we showed a tentative definition for the state value function $\qpolicies(s, \as)$ as the solution to a Bellman equality which we repeat here for clarity,
\begin{equation}
\qpolicies(s, \as) = R(s, \as) + \gamma\Exp_{s'\mid s, a}\left[ \sum_{\as'} \Pr(\as'\mid s') \qpolicies(s', \as') \right] \,,
\end{equation}
\noindent and then argued that this definition is improper due to issues with the definition of the $\Pr(\as\mid s)$ term, which conceptually represents the statistical dependency of the actions taken from the system state, but is actually subtly ill-defined.  In this section, we first provide a deeper explanation of the problems with $\Pr(\as\mid s)$ and why it is not generally well-defined. The issue is often overlooked by practical methods and implementations that attempt to learn centralized state value functions in the form of critic models, but was recently identified and exposed for the single-agent control case~\cite{baisero2022unbiased} and for the finite-horizon multi-agent control case~\cite{lyu2022deeper};  here, we extend this prior work to the infinite-horizon multi-agent control case. We then consider a few options to ``patch'' the issue, providing an alternative definition for $\qpolicies(s, \as)$ which is mathematically sound.

\subsection{Problems}

At a high level, the issue with the naive definition of state values $\qpolicies(s, \as)$ is that it relies upon a \emph{time-invariant} notion of conditional probabilities $\Pr(\as\mid s)$, which turns out to be generally ill-defined because probabilities $\Pr(\As_t=\as\mid S_t=s)$ are actually \emph{time-variant}.
More formally, the issue is that the term $\Pr(\as\mid s)$ is ambiguous and does not adequately identify the random variables $\As$ and $S$ onto which the values $\as$ and $s$ are assigned.  The graphical model associated with the Dec-POMDP defines a joint distribution over \emph{timed} random variables ($S_0, S_1, S_2, \ldots$, $\As_0$, $\As_1$, $\As_2, \ldots$, and others), but not over \emph{time-less} random variables ($S, \As$), so probabilities are formally defined only for the timed variables.  This issue is explained in more detail for the single-agent control case in \cite{baisero2022unbiased}.

\subsection{Solutions}
\label{appendix_solutions}

Having determined this fundamental flaw in the tentative definition of $\qpolicies(s, \as)$, we must either resign to the idea that state values are an intrinsically flawed concept, or we can try to determine a different viable definition for state values.  In this section, we explore a few ways to fix the issue with the definition of $\qpolicies(s, \as)$ by replacing $\Pr(\as\mid s)$ with a separate expression $p(\as\mid s)$.  We identify $3$ candidates (although more can also be formulated), of which only one is viable.

\subsubsection{Limiting Distribution}

The limiting distribution $p_L$ is defined as the limit for $t\to\infty$ of the timed distribution,
\begin{align}
p_L(s) &\doteq \lim_{t\to\infty} \Pr(S_t=s) \,, \\
p_L(s, \as) &\doteq \lim_{t\to\infty} \Pr(S_t=s, \As_t=\as) \,, \\
p_L(\as\mid s) &\doteq \frac{ p_L(s, \as) }{ p_L(s) } \,.
\end{align}
However, the theory of Markov chains tells us that limiting distributions are not guaranteed to converge.  Since we are looking for a notion of value function which is well-defined for all possible Dec-POMDPs and policies, the limiting distribution does not satisfy our needs.

\subsubsection{Total Visitations Distribution}

Another option is the total visitation distribution $p_{TV}$, defined as the limit for $N\to\infty$ of the average distribution over the first $N$ timesteps,
\begin{align}
p_{TV}(s) &\doteq \lim_{N\to\infty} \frac{1}{N} \sum_{t=0}^{N-1} \Pr(S_t=s) \,, \\
p_{TV}(s, \as) &\doteq \lim_{N\to\infty} \frac{1}{N} \sum_{t=0}^{N-1} \Pr(S_t=s, \As_t=a) \label{eq:p_sa:total-visitations} \,, \\
p_{TV}(\as\mid s) &\doteq \frac{ p_{TV}(s, \as) }{ p_{TV}(s) } \,. \label{eq:p_as_s:total-visitations}
\end{align}

Unfortunately, this limit is also not guaranteed to exist for all Dec-POMDPs and policies, and therefore inadequate to satisfy our needs.  We show this with a simple counter-example.

\paragraph{Counter-Example of Total Visitations Distribution}

Consider a Dec-POMDP with a singleton state space $\sset = \{ \zeta \}$, binary actions $\aset_i = \{ 0, 1 \}$, and a singleton observation space $\oset_i = \{ \omega \}$.  We note that the agent policies are intrinsically time-variant due to their dependence on their respective history and history lengths;  therefore, even though each agent can only ever receive the same observation over and over, the behaviors of the agents can vary arbitrarily with time.  Further, policies in Dec-POMDP are exclusively dependent on time $\policy_i(a; t)$, and deterministic policies can be represented directly as sequences of actions taken at each timestep, e.g., $\policy_i \equiv (0, 1, 0, 1, \ldots)$ denotes a policy that takes first action $0$, second action $1$, third action $0$, fourth action $1$, and so on.

Consider the deterministic agent policies $\policy_i \equiv (0, 1, 0, 1, 0, 0, 1, 1, 0, 0, 0, 0, 1, 1, 1, 1, \ldots)$, which alternates the emission of actions in sequences of ever-increasing length (i.e., one 0, one 1, one 0, one 1, two $0$s, two $1$s, four $0$s, four $1$s, eight $0$s, eight $1$s, sixteen $0$s, sixteen $1$s, etc).  In this case, the average in \Cref{eq:p_sa:total-visitations} as a function of time oscillates forever, never converging, e.g., $\frac{1}{N}\sum_{t=0}^{N-1} \Pr(S_t=\zeta, \As_t=\bm{1}) = \frac{1}{2}$ when $N=2, 4, 8, 16, 32, \ldots$, and $\frac{1}{N}\sum_{t=0}^{N-1} \Pr(S_t=\zeta, \As_t=\bm{1}) = \frac{1}{3}$ when $N=3, 6, 12, 24, 48, \ldots$.  Interestingly, since the counter-example uses a singleton state space, the issue with $p_{TV}$ is not even related to state uncertainty, but rather to the general notion of computing non-discounted action counts for time-indefinite horizon sequences of actions produced by time-variant policies.

\subsubsection{Discounted Visitations Distribution}

Finally, consider the discounted visitation distribution $p_{DV}$, defined as the average discounted distribution over all timesteps,
\begin{align}
p_{DV}(s) &\doteq (1-\gamma) \sum_{t=0}^\infty \gamma^t \Pr(S_t=s) \,, \\
p_{DV}(s, \as) &\doteq (1-\gamma) \sum_{t=0}^\infty \gamma^t \Pr(S_t=s, \As_t=\as) \,, \\
p_{DV}(\as\mid s) &\doteq \frac{ p_{DV}(s, \as) }{ p_{DV}(s) } \,.
\end{align}

In this case, the geometric factor $\gamma^t$ guarantees convergence in all cases, even in the above counter-example. This is a notion closely related to actual visitation counts, albeit slightly ``warped'' by the weighting, which makes later visitations count less than earlier visitations.  The effect of this ``warping'' is obviously larger for larger discounting (smaller $\gamma$), and smaller for smaller discounting (large $\gamma$). Despite this warping, we have finally achieve a somewhat reasonable notion of state-conditioned action distribution, with which to define the state value function as the unique solution to the following \emph{state} Bellman equality,
\begin{equation}
\qpolicies(s, \as) = R(s, \as) + \gamma\Exp_{s'\mid s, \as}\left[ \sum_{\as'} p_{DV}(\as'\mid s') \qpolicies(s', \as') \right] \,.
\end{equation}

\paragraph{Final Notes about Notation and Consistency}

We conclude by noting that $p_{DV}$ is the same as the $\rho$ defined in \Cref{sec:rho}, which is the preferred notation in all other sections of this paper.  Finally, we note that this ``patch'' is in some sense still consistent with the definitions of the other value functions, because in the cases where the conditional action distribution also takes some history into account, then $\rho(\as\mid\cdot)$ is equivalent to $\policies(\as\mid\cdot)$, e.g., $\rho(\as\mid \hs) = \policies(\as; \hs)$ and $\rho(\as\mid \hs,s) = \policies(\as; \hs)$.

\section{Proofs}

\subsection{Uniqueness of Bellman Equation Solutions}
\label{sec:convergence_of_history_based_critics}

Here, we prove that the Bellman equations in \Cref{eq:bellman:q_hs_as,eq:bellman:qi_h_a,eq:bellman:q_s_as,eq:bellman:q_hs_s_as} each have a unique solution.
To simplify the following analysis, we represent Q-functions and the reward function as vectors, which allows us to prove the uniqueness of the solution for a wide class of Bellman-like equations simultaneously.

Let $Q \in \R^n$ be an arbitrary Q-function and let $R \in \R^m$ be a reward function.
\Cref{eq:bellman:q_hs_as,eq:bellman:qi_h_a,eq:bellman:q_s_as,eq:bellman:q_hs_s_as} can be written in the general form
\begin{equation}
    Q = P_0 R + \gamma P_1 Q
    \,,
\end{equation}
where $P_0 \in \R^{n \times m}$ and $P_1 \in \R^{n \times n}$ are stochastic matrices.
(To see why, notice that each element of $P_0 R$ and $P_1 Q$ is an expectation over the values of $R$ or $Q$, since the rows of $P_0$ and $P_1$ represent probability distributions.)
Consequently, we have $0 \leq P_k \leq 1$ elementwise and $P_k e = e$, $\forall~k \in \{0,1\}$, where $e$ is the vector of ones with appropriate dimensions.
\begin{proposition}
    \label{prop:general_convergence}
    Define the mapping $B \colon Q \mapsto P_0 R + \gamma P_1 Q$.
    When $\gamma < 1$, the equation
    \begin{equation}
        \label{eq:bellman_operator}
        Q = B Q
    \end{equation}
    has a unique solution $Q^* \doteq \sum_{k=0}^\infty (\gamma P_1)^k P_0 R$.
\end{proposition}
\begin{proof}
    Let $\norm{Q}$ denote the maximum norm of $Q$.
    Since the infinity norm is submultiplicative (and coincides with the maximum norm for vectors), then for every $Q,Q' \in \R^n$,
    \begin{align}
        \norm{B Q - B Q'}
        &= \norm{P_0 R + \gamma P_1 Q - P_0 R - \gamma P_1 Q'} \nonumber \\
        &= \norm{\gamma P_1 (Q - Q')} \nonumber \\
        &\leq \gamma \norm{P_1}_\infty \norm{Q - Q'} \,.
    \end{align}
    The facts that $P_1 \geq 0$ and $P_1 e = e$ imply that $\norm{P_1}_\infty = 1$.
    Therefore,
    \begin{equation}
        \label{eq:contraction_map}
        \norm{B Q_1 - B Q_2} \leq \gamma \norm{Q_1 - Q_2}
        ,
    \end{equation}
    and $B$ is a contraction mapping with respect to the maximum norm.
    We conclude that $B$ admits a unique fixed point $Q^*$;
    if $B$ had two distinct fixed points $Q^*_1 \neq Q^*_2$, then ${\norm{B Q^*_1 - B Q^*_2} = \norm{Q^*_1 - Q^*_2}}$,
    which would contradict \Cref{eq:contraction_map} for $\gamma < 1$.
    
    To complete the proof, we must identify $Q^*$.
    Expanding \Cref{eq:bellman_operator} gives
    \begin{equation*}
        Q = P_0 R + \gamma P_1 Q
        \,,
    \end{equation*}
    and rearranging the terms yields the solution
    \begin{equation*}
        Q = (I - \gamma P_1)^{-1} P_0 R
        \,.
    \end{equation*}
    From the infinite geometric series, we have $(I - \gamma P_1)^{-1} = \sum_{k=0}^\infty (\gamma P_1)^k$, and therefore $Q^* = \sum_{k=0}^\infty (\gamma P_1)^k P_0 R$.
\end{proof}

\subsection{Value Function Identities}
\label{sec:value_function_identities}

\thmidentityqihaqha*

\begin{proof}
    $\qpolicies_i(h_i,a_i)$ is defined as the unique solution to \Cref{eq:bellman:qi_h_a}.
    Next, we show that $\expect{\hs,\as\mid h_i,a_i}{\qpolicies(\hs, \as)}$ (as a function of $h_i$ and $a_i$) is also a solution to \Cref{eq:bellman:qi_h_a};
    since the solution is unique by \Cref{prop:general_convergence}, the two functions must be identically equal and \Cref{eq:identity:qiha:qha} must hold.

    We first relate the two history-based reward functions:
    \begin{align}
        R_i(h_i,a_i)
        &= \expect{s,\as_\noti \mid h_i}{R(s, \as)} \nonumber \\
        &= \expect{s,\as \mid h_i,a_i}{R(s, \as)} \nonumber \\
        &= \expect{\hs,s,\as \mid h_i,a_i}{R(s, \as)} \nonumber \\
        &= \expect{\hs,\as \mid h_i,a_i}{
            \expect{s \mid \hs}{R(s, \as)}
        } \nonumber \\
        &= \expect{\hs,\as \mid h_i,a_i}{R(\hs, \as)} \,.
    \end{align}
    \noindent
    Next, we show the following identity:
   \begin{align}
        &\phantom{{}={}} \expect{o_i\mid h_i,a_i}{ \sum_{a'_i} \policy_i(a'_i; h_i a_i o_i) \expect{\hs\as\os,\as'\mid h_i a_i o_i,a_i'}{\qpolicies(\hs\as\os,\as')}
        } \nonumber \\
        &= \expect{o_i\mid h_i,a_i}{ \Exp_{a'_i\mid h_i a_i o_i} \left[ \expect{\hs\as\os,\as'\mid h_i a_i o_i,a_i'}{\qpolicies(\hs\as\os,\as')} \right]
        } \nonumber \\
        &= \expect{\hs,\as,\os,\as'\mid h_i,a_i}{ \qpolicies(\hs\as\os,\as')
        } \nonumber \\
        &= \expect{\hs,\as\mid h_i,a_i}{ \expect{\os\mid\hs,\as}{ \expect{\as'\mid\hs\as\os}{ \qpolicies(\hs\as\os,\as') } }
        } \nonumber \\
        &= \expect{\hs,\as\mid h_i,a_i}{ \expect{\os\mid\hs,\as}{ \sum_{\as'}{ \policies(\as'; \hs\as\os) \qpolicies(\hs\as\os,\as') } }
        }
        \,.
    \end{align}
    \noindent
    Therefore, when we substitute
    $\expect{\hs\as\os,\as'\mid h_i a_i o_i,a'_i}{\qpolicies(\hs\as\os,\as')}$
    (analogous to
    $\expect{\hs,\as\mid h_i,a_i}{\qpolicies(\hs,\as)}$)
    into \Cref{eq:bellman:qi_h_a}, we obtain
   \begin{align}
        &\phantom{{}={}} R_i(h_i, a_i) + \gamma \expect{o_i\mid h_i,a_i}{ \sum_{a'_i} \policy_i(a'_i; h_i a_i o_i) \expect{\hs\as\os,\as'\mid h_i a_i o_i,a_i'}{\qpolicies(\hs\as\os,\as')}
        } \nonumber \\
        &= \expect{\hs,\as\mid h_i,a_i}{
            R(\hs, \as) + \gamma \expect{\os\mid \hs,\as}{ \sum_{\as'} \policies(\as'; \hs\as\os) \qpolicies(\hs\as\os,\as') }
        } \nonumber \\
        &= \expect{\hs,\as\mid h_i,a_i}{\qpolicies(\hs,\as)}
        \,,
    \end{align}
    where the final equality follows from \Cref{eq:bellman:q_hs_as}.
    Because the original quantity
    $\expect{\hs,\as\mid h_i,a_i}{\qpolicies(\hs,\as)}$
    is unchanged after the substitution into \Cref{eq:bellman:qi_h_a}, we conclude that it must be the unique fixed point guaranteed by \Cref{prop:general_convergence}, and hence $\qpolicies_i(h_i,a_i) = \expect{\hs,\as\mid h_i,a_i}{\qpolicies(\hs,\as)}$.
\end{proof}

\thmidentityqhaqhsa*

\begin{proof}
$\qpolicies(\hs,\as)$ is defined as the unique solution to \Cref{eq:bellman:q_hs_as}.  Next, we show that $\Exp_{s\mid \hs, \as}\left[ \qpolicies(\hs, s, \as) \right]$ (as a function of $\hs$ and $\as$) is also a solution to \Cref{eq:bellman:q_hs_as};  since the solution is unique by \Cref{prop:general_convergence}, the two functions must be identically equal and \Cref{eq:identity:qha:qhsa} must hold.
We first recall the definition of the joint history reward function, $R(\hs, \as) \doteq \Exp_{s\mid \hs}\left[ R(s, \as) \right]$.
Next, we show the following identity:
\begin{align}
&\phantom{{}={}} \Exp_{\os\mid \hs,\as} \left[ \sum_{\as'} \policies(\as'; \hs\as\os) \Exp_{s'\mid \hs\as\os} \left[ \qpolicies(\hs\as\os, s', \as') \right] \right] \nonumber \\
&= \Exp_{\os\mid \hs,\as} \left[ \Exp_{\as'\mid \hs\as\os} \left[ \Exp_{s'\mid \hs\as\os} \left[ \qpolicies(\hs\as\os, s', \as') \right] \right] \right] \nonumber \\
&= \Exp_{\os,s',\as' \mid \hs,\as} \left[ \qpolicies(\hs\as\os, s', \as') \right] \nonumber \\
&= \Exp_{s,\os,s',\as' \mid \hs,\as} \left[ \qpolicies(\hs\as\os, s', \as') \right] \nonumber \\
&= \Exp_{s\mid \hs} \left[ \Exp_{s', \os\mid s, \as} \left[ \sum_{\as'} \policies(\as'; \hs\as\os) \Exp_{s'\mid \hs\as\os} \left[ \qpolicies(\hs\as\os, s', \as') \right] \right] \right] \,. 
\end{align}

Therefore, when we substitute $\Exp_{s'\mid \hs\as\os}\left[ \qpolicies(\hs\as\os, s', \as') \right]$ (analogous to $\Exp_{s\mid\hs}\left[ \qpolicies(\hs, s, \as) \right]$) into \Cref{eq:bellman:q_hs_as}, we obtain
\begin{align}
&\phantom{{}={}} R(\hs, \as) + \gamma\Exp_{\os\mid\hs\as} \left[ \sum_{\as'} \policies(\as'; \hs\as\os) \Exp_{s'\mid \hs\as\os} \left[ \qpolicies(\hs\as\os, s', \as') \right] \right] \nonumber \\
&= \Exp_{s\mid \hs} \left[ R(s, \as) + \gamma\Exp_{s', \os\mid s, \as} \left[ \sum_{\as'} \policies(\as'; \hs\as\os) \qpolicies(\hs\as\os, s', \as') \right] \right] \nonumber \\
&= \Exp_{s\mid \hs} \left[ \qpolicies(\hs, s, \as) \right] \,,
\end{align}
\noindent where the final equality follows from \Cref{eq:bellman:q_hs_s_as}.  Because the original quantity $\Exp_{s\mid \hs}\left[ \qpolicies(\hs, s, \as) \right]$ is unchanged after the substitution into \Cref{eq:bellman:q_hs_as}, we conclude that it must be the unique fixed point guaranteed by \Cref{prop:general_convergence}, and hence $\qpolicies(\hs, \as) = \Exp_{s\mid \hs}\left[ \qpolicies(\hs, s, \as) \right]$.

\end{proof}

\subsection{Biased State-Based Critic}
\label{sec:biased_state_based_critic}

\begin{lemma}\label{thm:identity:qiha:qsa}
$\qpolicies(s, \as)$ may be a biased estimate of $\qpolicies_i(h_i, a_i)$, i.e., the equality
\begin{equation}
\qpolicies_i(h_i, a_i) = \Exp_{s, \as\mid h_i, a_i} \left[ \qpolicies(s, \as) \right]
\end{equation}
\noindent does not necessarily hold.
\end{lemma}

\begin{proof}
We note that \Cref{thm:bias:qha:qsa} does not imply this theorem, however, we prove this theorem using an analogous argument by contradiction, which is applicable to our novel always-well-defined state value function $\qpolicies(s, \as)$.  For a generic environment, consider an arbitrary action $a_i$, and two different individual histories $h_{(i,A)}\neq h_{(i,B)}$ associated with the same state and joint action distributions $\rho(s, \as\mid h_{(i,A)}, a_i) = \rho(s, \as\mid h_{(i,B)}, a_i)$ (a plausible occurrence in partially observable environments).  Because the individual histories are different, there are many history-based policies $\policy_i$ which result in different future behaviors starting from $h_{(i,A)}$ and $h_{(i,B)}$, which result in different history values,
\begin{equation}
\qpolicies_i(h_{(i,A)}, a_i) \neq \qpolicies_i(h_{(i,B)}, a_i) \,. \label{eq:bias-proof:qiha:qsa:1}
\end{equation}

On the other hand, the conditional state and joint action distributions $\rho(s, \as\mid h_{(i,A)}, \as) = \rho(s, \as\mid h_{(i,B)}, \as)$ are equal, which implies that the expected state values are also equal,
\begin{equation}
\Exp_{s, \as\mid h_{(i,A)}, a_i} \left[ \qpolicies(s, \as) \right] = \Exp_{s, \as\mid h_{(i,B)}, a_i} \left[ \qpolicies(s, \as) \right] \,. \label{eq:bias-proof:qiha:qsa:2}
\end{equation}

Assuming that the equality $\qpolicies_i(h_i, a_i) = \Exp_{s, \as\mid h_i, a_i} \left[ \qpolicies(s, \as) \right]$ holds leads to a contradiction with \Cref{eq:bias-proof:qiha:qsa:1,eq:bias-proof:qiha:qsa:2},
\begin{equation}
    \Exp_{s, \as\mid h_{(i,A)}, a_i}\left[ \qpolicies(s, \as) \right] = \qpolicies_i(h_{(i,A)}, a_i) \neq \qpolicies_i(h_{(i,B)}, a_i) = \Exp_{s, \as\mid h_{(i,B)}, a_i}\left[ \qpolicies(s, \as) \right] \,.
\end{equation}
Therefore, $\qpolicies_i(h_i, a_i) = \Exp_{s, \as\mid h_i, a_i}\left[ \qpolicies(s, \as) \right]$ does not hold in general.
\end{proof}

\thmgradientbias*

\begin{proof}
    To cover different angles and intuitions, we provide two proofs;  one by contradiction, and one by example.  Both proofs are based on a simple policy parameterization for which the statements of this theorem and \Cref{thm:bias:qha:qsa} imply each other.  Since \Cref{thm:bias:qha:qsa} is already proven to hold in a way that is invariant to the policy parameterization, then this theorem must also hold.  The policy parameterization is the tabular one where $\policy_i(a; h) = \theta_{i,h,a}$, which can likewise be written as a linear model $\policy_i(a; h) = \theta_i\T\ones_{h,a}$ where $\ones_{h,a}$ is an indicator vector of $0$s with a $1$ only in the dimension associated with the $h, a$ pair;  this parameterization requires constraints $\theta_{i,h,a} \ge 0$ and $\sum_a \theta_{i,h,a} = 1$, but this does not influence the conclusion.
    \paragraph{Proof by Contradiction}
    For tabular policies, we have
    \begin{align}
    \grad_s &= (1-\gamma) \Exp_{\hs, s, \as}\left[ \qpolicies(s, \as) \nablai\log\policy_i(a_i; h_i) \right] \nonumber \\
    &= (1-\gamma) \Exp_{h_i, a_i} \left[ \Exp_{\hs, s, \as\mid h_i, a_i} \left[ \qpolicies(s, \as) \nablai\log\policy_i(a_i; h_i) \right] \right] \nonumber \\
    &= (1-\gamma) \Exp_{h_i, a_i} \left[ \Exp_{s, \as\mid h_i, a_i} \left[ \qpolicies(s, \as) \right] \frac{ \ones_{h_i, a_i} }{ \policy_i(a_i; h_i) } \right] \nonumber \\
    &= (1-\gamma) \sum_{h_i, a_i} \rho(h_i, a_i) \Exp_{s, \as\mid h_i, a_i} \left[ \qpolicies(s, \as) \right] \frac{ \ones_{h_i, a_i} }{ \policy_i(a_i; h_i) } \,, \\
    \grad_\hs &= (1-\gamma) \Exp_{\hs, \as}\left[ \qpolicies(\hs, \as) \nablai\log\policy_i(a_i; h_i) \right] \nonumber \\
    &= (1-\gamma) \Exp_{h_i, a_i} \left[ \Exp_{\hs, \as\mid h_i, a_i} \left[ \qpolicies(\hs, \as) \nablai\log\policy_i(a_i; h_i) \right] \right] \nonumber \\
    &= (1-\gamma) \Exp_{h_i, a_i} \left[ \qpolicies_i(h_i, a_i) \frac{ \ones_{h_i, a_i} }{ \policy_i(a_i; h_i) } \right] \nonumber \\
    &= (1-\gamma) \sum_{h_i, a_i} \rho(h_i, a_i) \qpolicies_i(h_i, a_i) \frac{ \ones_{h_i, a_i} }{ \policy_i(a_i; h_i) } \,.
    \end{align}

    More specifically, the dimensions of $\grad_s$ and $\grad_\hs$ associated with $h_i, a_i$ take on values
    \begin{align}
    \left[ \grad_s \right]_{h_i,a_i} &= (1-\gamma) \rho(h_i, a_i) \Exp_{s, \as\mid h_i, a_i} \left[ \qpolicies(s, \as) \right] \frac{1}{\policy_i(a_i; h_i)} \,, \\
    \left[ \grad_\hs \right]_{h_i, a_i} &= (1-\gamma) \rho(h_i, a_i) \qpolicies_i(h_i, a_i) \frac{1}{\policy_i(a_i; h_i)} \,,
    \end{align}
    \noindent which are equal if and only if $\qpolicies_i(a_i, h_i) = \Exp_{s, \as\mid h_i, a_i} \left[ \qpolicies(s, \as) \right]$.  \Cref{thm:identity:qiha:qsa} generally disproves such an equality.  Therefore, the gradient equality $\grad_s = \grad_\hs$ does not necessarily hold.
    \paragraph{Proof by Example}
Continuing from the Dec-Tiger example in~\Cref{sec:bias}, we compare the IACC-H and IACC-S gradients $\grad_\hs$ and $\grad_s$ for agent 1 at one specific dimension and show that they are different.  We focus on the values associated with $h_1 = (\textit{listen}, \textit{hear-left}, \textit{listen}, \textit{hear-left})$, which is intrinsically associated with action $a_1 = \textit{open-right}$.  We denote this dimension of the two gradient as $\left[ \grad_\hs \right]_{h_1,a_1}$ and $\left[ \grad_s \right]_{h_1,a_1}$.  
To compute these values, we need to reference all the joint histories that are compatible with the individual history $h_1$;  we denote these using labels that reference the second agent's observations,
\begin{align}
\hs_{LL} &= ((\textit{listen}, \textit{listen}), (\textit{hear-left}, \textit{hear-left}), (\textit{listen}, \textit{listen}), (\textit{hear-left}, \textit{hear-left})) \,, \\
\hs_{LR} &= ((\textit{listen}, \textit{listen}), (\textit{hear-left}, \textit{hear-left}), (\textit{listen}, \textit{listen}), (\textit{hear-left}, \textit{hear-right})) \,, \\
\hs_{RL} &= ((\textit{listen}, \textit{listen}), (\textit{hear-left}, \textit{hear-right}), (\textit{listen}, \textit{listen}), (\textit{hear-left}, \textit{hear-left})) \,, \\
\hs_{RR} &= ((\textit{listen}, \textit{listen}), (\textit{hear-left}, \textit{hear-right}), (\textit{listen}, \textit{listen}), (\textit{hear-left}, \textit{hear-right})) \,.
\end{align}
We also need to reference all the joint actions that are compatible with the individual action $a_1$;  we denote these using labels that reference the second agent's action, $\as_L = (\textit{open-right}, \textit{open-left})$ and $\as_R = (\textit{open-right}, \textit{open-right})$.
We do not consider other joint histories and actions because they do not contribute to $\left[ \grad_\hs \right]_{h_1,a_1}$ or $\left[ \grad_s \right]_{h_1,a_1}$. Given the details of the Dec-Tiger problem, we can compute the conditional probabilities of these joint histories and actions $\Pr(\hs, \as\mid h_1, a_1)$, contained in \Cref{tab:dectiger:pr_joint_ha_given_individual_ha}.  To start, we have
\begin{align}
\rho(h_1, a_1) &= \rho(h_1) \policy_1(a_1; h_1) \nonumber \\
&= \sum_s \rho(h_1\mid s) \rho(s) \policy_1(a_1; h_1) \nonumber \\
&= 0.85^2 \cdot \frac{1}{2} \cdot 1 + 0.15^2 \cdot \frac{1}{2} \cdot 1 \nonumber \\
&= 0.373 \,.
\end{align}

Then, using probabilities from \Cref{tab:dectiger:pr_joint_ha_given_individual_ha,tab:dectiger:pr_sa_given_individual_ha} and values from \Cref{sec:dec-tiger},
\begin{align}
	\left[ \grad_\hs  \right]_{h_1,a_1} &= (1-\gamma) \rho(h_1, a_1) \Exp_{\hs,\as\mid h_1,a_1} \left[ \qpolicies(\hs, \as) \right] \frac{1}{\policy_1(a_1; h_1)} \nonumber \\
    &= (1-\gamma) \cdot 0.373 \cdot (-0.854) \cdot \frac{1}{1} \nonumber \\
    &= -0.319 (1-\gamma) \,. \\
    \left[ \grad_s \right]_{h_1,a_1} &= (1-\gamma) \rho(h_1, a_1) \Exp_{s,\as\mid h_1, a_1} \left[ \qpolicies(s, \as) \right] \frac{1}{\policy_1(a_1; h_1)} \nonumber \\
    &= (1-\gamma) \cdot 0.373 \cdot (4.465) \cdot \frac{1}{1} \nonumber \\
    &= 1.665 (1-\gamma) \,.
\end{align}
Comparing values, we have $\left[ \grad_s \right]_{h_1,a_1} \neq \left[ \grad_\hs \right]_{h_1,a_1}$, and therefore $\grad_s \neq \grad_h$.
\end{proof}

\section{Dec-Tiger Values and Gradients}\label{sec:dec-tiger}

Here, we manually calculate the state values $\qpolicies(s, \as)$ for the Dec-Tiger problem.  We consider the case of two agents who always listen twice, then open the most promising door:  left door, if the tiger has been heard more on the right;  right door, if the tiger has been heard more on the right;  otherwise, randomly.
We proceed by progressively computing all values of $\eta(s)$, $\eta(s, \as)$, $\rho(\as\mid s)$, and $\qpolicies(s, \as)$ for state \texttt{L} and the viable joint actions;  by symmetry, analogous calculations can be made for state \texttt{R}.

To simplify notation, we denote state \textit{tiger-left} as \texttt{L}, state \textit{tiger-right} as \texttt{R}, observation \textit{hear-left} as \texttt{L}, observation \textit{hear-right} as \texttt{R}, action \textit{listen} as \texttt{Li}, action \textit{open-left} as \texttt{L}, and action \textit{open-right} as \texttt{R}.

\subsection{Discounted Counts and Probabilities}

\begin{table}
\centering
\caption{$\Pr(\As_2=\as\mid\Hs_2=\hs) = \policies(\as; \hs)$.  In this table, a joint observation $\hs=((A, B), (C, D))$ means that the two agents have received observations $A$ and $B$ on the first timestep, and $C$ and $D$ on the second timestep.}
\label{tab:dectiger:pr_as_hs}
\begin{tabular}{cllll}
    $\as$ & $(\texttt{L}, \texttt{L})$ & $(\texttt{L}, \texttt{R})$ & $(\texttt{R}, \texttt{L})$ & $(\texttt{R}, \texttt{R})$ \\
    \toprule
    $\hs$ \\
    \toprule
    $((\texttt{L}, \texttt{L}), (\texttt{L}, \texttt{L}))$ & & & & 1.0 \\
    $((\texttt{L}, \texttt{L}), (\texttt{L}, \texttt{R}))$ & & & 0.5 & 0.5 \\
    $((\texttt{L}, \texttt{L}), (\texttt{R}, \texttt{L}))$ & & 0.5 & & 0.5 \\
    $((\texttt{L}, \texttt{L}), (\texttt{R}, \texttt{R}))$ & 0.25 & 0.25 & 0.25 & 0.25 \\
    $((\texttt{L}, \texttt{R}), (\texttt{L}, \texttt{L}))$ & & & 0.5 & 0.5 \\
    $((\texttt{L}, \texttt{R}), (\texttt{L}, \texttt{R}))$ & & & 1.0 & \\
    $((\texttt{L}, \texttt{R}), (\texttt{R}, \texttt{L}))$ & 0.25 & 0.25 & 0.25 & 0.25 \\
    $((\texttt{L}, \texttt{R}), (\texttt{R}, \texttt{R}))$ & 0.5 & & 0.5 & \\
    $((\texttt{R}, \texttt{L}), (\texttt{L}, \texttt{L}))$ & & 0.5 & & 0.5 \\
    $((\texttt{R}, \texttt{L}), (\texttt{L}, \texttt{R}))$ & 0.25 & 0.25 & 0.25 & 0.25 \\
    $((\texttt{R}, \texttt{L}), (\texttt{R}, \texttt{L}))$ & & 1.0 & & \\
    $((\texttt{R}, \texttt{L}), (\texttt{R}, \texttt{R}))$ & 0.5 & 0.5 & & \\
    $((\texttt{R}, \texttt{R}), (\texttt{L}, \texttt{L}))$ & 0.25 & 0.25 & 0.25 & 0.25 \\
    $((\texttt{R}, \texttt{R}), (\texttt{L}, \texttt{R}))$ & 0.5 & & 0.5 & \\
    $((\texttt{R}, \texttt{R}), (\texttt{R}, \texttt{L}))$ & 0.5 & 0.5 & & \\
    $((\texttt{R}, \texttt{R}), (\texttt{R}, \texttt{R}))$ & 1.0 & & & \\
    \bottomrule
\end{tabular}
\end{table}

\begin{table}
\centering
\caption{$\Pr(\Hs_2=\hs\mid S_2=s) = 0.85^n \cdot 0.15^{2-n}$, where $n$ is the number of times the tiger was heard through the correct door. In this table, a joint observation $\hs=((A, B), (C, D))$ means that the two agents have received observations $A$ and $B$ on the first timestep, and $C$ and $D$ on the second timestep.}
\label{tab:dectiger:pr_hs_s}
\begin{tabular}{crr}
    $s$ & $L$ & $R$ \\
    \toprule
    $\hs$ \\
    \toprule
    $((\texttt{L}, \texttt{L}), (\texttt{L}, \texttt{L}))$ & 0.522 & 0.001 \\
    $((\texttt{L}, \texttt{L}), (\texttt{L}, \texttt{R}))$ & 0.092 & 0.003 \\
    $((\texttt{L}, \texttt{L}), (\texttt{R}, \texttt{L}))$ & 0.092 & 0.003 \\
    $((\texttt{L}, \texttt{L}), (\texttt{R}, \texttt{R}))$ & 0.016 & 0.016 \\
    $((\texttt{L}, \texttt{R}), (\texttt{L}, \texttt{L}))$ & 0.092 & 0.003 \\
    $((\texttt{L}, \texttt{R}), (\texttt{L}, \texttt{R}))$ & 0.016 & 0.016 \\
    $((\texttt{L}, \texttt{R}), (\texttt{R}, \texttt{L}))$ & 0.016 & 0.016 \\
    $((\texttt{L}, \texttt{R}), (\texttt{R}, \texttt{R}))$ & 0.003 & 0.092 \\
    $((\texttt{R}, \texttt{L}), (\texttt{L}, \texttt{L}))$ & 0.092 & 0.003 \\
    $((\texttt{R}, \texttt{L}), (\texttt{L}, \texttt{R}))$ & 0.016 & 0.016 \\
    $((\texttt{R}, \texttt{L}), (\texttt{R}, \texttt{L}))$ & 0.016 & 0.016 \\
    $((\texttt{R}, \texttt{L}), (\texttt{R}, \texttt{R}))$ & 0.003 & 0.092 \\
    $((\texttt{R}, \texttt{R}), (\texttt{L}, \texttt{L}))$ & 0.016 & 0.016 \\
    $((\texttt{R}, \texttt{R}), (\texttt{L}, \texttt{R}))$ & 0.003 & 0.092 \\
    $((\texttt{R}, \texttt{R}), (\texttt{R}, \texttt{L}))$ & 0.003 & 0.092 \\
    $((\texttt{R}, \texttt{R}), (\texttt{R}, \texttt{R}))$ & 0.001 & 0.552 \\
    \bottomrule
\end{tabular}
\end{table}

\begin{table}
\centering
\caption{$\Pr(\As_2=\as\mid S_2=s) = \sum_\hs \Pr(\As_2=\as \mid \Hs_2=\hs) \Pr(\Hs_2=\hs\mid S_2=s)$.}
\label{tab:dectiger:pr_as_s}
\begin{tabular}{crr}
    $s$ & $L$ & $R$ \\
    \toprule
    $\as$ \\
    \toprule
    $(\texttt{L}, \texttt{L})$ & 0.0225 & 0.7225 \\
    $(\texttt{L}, \texttt{R})$ & 0.1275 & 0.1275 \\
    $(\texttt{R}, \texttt{L})$ & 0.1275 & 0.1275 \\
    $(\texttt{R}, \texttt{R})$ & 0.7225 & 0.0225 \\
    \bottomrule
\end{tabular}
\end{table}

\begin{table}
\centering
\caption{$\Pr(\hs, \as\mid h_1, a_1)$. In this table, a joint observation $\hs=((A, B), (C, D))$ means that the two agents have received observations $A$ and $B$ on the first timestep, and $C$ and $D$ on the second timestep.}
\label{tab:dectiger:pr_joint_ha_given_individual_ha}
\begin{tabular}{cll}
    $\as$ & $(\texttt{R}, \texttt{L})$ & $(\texttt{R}, \texttt{R})$ \\
    \toprule
    $\hs$ \\
    \toprule
    $((\texttt{L}, \texttt{L}), (\texttt{L}, \texttt{L}))$ & 0.0 & 0.7014 \\
    $((\texttt{L}, \texttt{L}), (\texttt{L}, \texttt{R}))$ & 0.0638 & 0.0638 \\
    $((\texttt{L}, \texttt{R}), (\texttt{L}, \texttt{L}))$ & 0.0638 & 0.0638 \\
    $((\texttt{L}, \texttt{R}), (\texttt{L}, \texttt{R}))$ & 0.0436 & 0.0 \\
    \bottomrule
\end{tabular}
\end{table}

\begin{table}
\centering
\caption{$\Pr(s, \as\mid h_1, a_1)$.}
\label{tab:dectiger:pr_sa_given_individual_ha}
\begin{tabular}{ccc}
    $s$ & $\as$ & $\Pr(s, \as\mid h_1, a_1)$ \\
    \toprule
    $\texttt{L}$ & $(\texttt{R}, \texttt{L})$ & 0.145 \\
    $\texttt{L}$ & $(\texttt{R}, \texttt{R})$ & 0.824 \\
    $\texttt{R}$ & $(\texttt{R}, \texttt{L})$ & 0.026 \\
    $\texttt{R}$ & $(\texttt{R}, \texttt{R})$ & 0.005 \\
    \bottomrule
\end{tabular}
\end{table}

We start by programmatically computing relevant probabilities, in \Cref{tab:dectiger:pr_as_hs,tab:dectiger:pr_hs_s,tab:dectiger:pr_as_s,tab:dectiger:pr_joint_ha_given_individual_ha,tab:dectiger:pr_sa_given_individual_ha}~\footnote{See \url{https://github.com/lyu-xg/on-centralized-critics-in-marl/tree/main/dectiger}}.
Because Dec-Tiger is episodic, and our policies guarantee that the episode terminates after timestep 3, we get the discounted state counts
\begin{align}
\eta(s=\texttt{L}) &= \Pr(S_0=\texttt{L}) + \gamma \Pr(S_1=\texttt{L}) + \gamma^2 \Pr(S_2=\texttt{L}) \nonumber \\
&= \frac{1+\gamma+\gamma^2}{2} \,.
\end{align}
Similarly, we get the discounted state-action counts
\begin{align}
\eta(s=\texttt{L}, \as=(\texttt{Li}, \texttt{Li})) &= \Pr(S_0=\texttt{L}, \As_0=(\texttt{Li}, \texttt{Li})) \nonumber \\
&\phantom{{}={}} + \gamma \Pr(S_1=\texttt{L}, \As_1=(\texttt{Li}, \texttt{Li})) \nonumber \\
&\phantom{{}={}} + \gamma^2 \Pr(S_2=\texttt{L}, \As_2=(\texttt{Li}, \texttt{Li})) \nonumber \\
&= \frac{1+\gamma}{2} \,, \\
\eta(s=\texttt{L}, \as=(\texttt{L}, \texttt{L})) &= \gamma^2 \Pr(S_2=\texttt{L}, \As_2=(\texttt{L}, \texttt{L})) \nonumber \\
&= \gamma^2 \Pr(S_2=\texttt{L}) \Pr(\As_2=(\texttt{L}, \texttt{L}) \mid S_2=\texttt{L}) \nonumber \\
&= \gamma^2 \frac{1}{2} 0.0225 \,, \\
\eta(s=\texttt{L}, \as=(\texttt{L}, \texttt{R})) &= \gamma^2 \Pr(S_2=\texttt{L}, \As_2=(\texttt{L}, \texttt{R})) \nonumber \\
&= \gamma^2 \Pr(S_2=\texttt{L}) \Pr(\As_2=(\texttt{L}, \texttt{R}) \mid S_2=\texttt{L}) \nonumber \\
&= \gamma^2 \frac{1}{2} 0.1275 \,, \\
\eta(s=\texttt{L}, \as=(\texttt{R}, \texttt{L})) &= \gamma^2 \Pr(S_2=\texttt{L}, \As_2=(\texttt{R}, \texttt{L})) \nonumber \\
&= \gamma^2 \Pr(S_2=\texttt{L}) \Pr(\As_2=(\texttt{R}, \texttt{L}) \mid S_2=\texttt{L}) \nonumber \\
&= \gamma^2 \frac{1}{2} 0.1275 \,, \\
\eta(s=\texttt{L}, \as=(\texttt{R}, \texttt{R})) &= \gamma^2 \Pr(S_2=\texttt{L}, \As_2=(\texttt{R}, \texttt{R})) \nonumber \\
&= \gamma^2 \Pr(S_2=\texttt{L}) \Pr(\As_2=(\texttt{R}, \texttt{R})\mid S_2=\texttt{L}) \nonumber \\
&= \gamma^2 \frac{1}{2} 0.7225 \,,
\end{align}
\noindent and the discounted distributions
\begin{align}
\rho(\as = (\texttt{Li}, \texttt{Li}) \mid s = L) &= \frac{ 1 + \gamma }{ 1 + \gamma + \gamma^2 } \\
\rho(\as = (\texttt{L}, \texttt{L}) \mid s = L) &= \frac{ 0.0225 \gamma^2 }{ 1 + \gamma + \gamma^2 } \\
\rho(\as = (\texttt{L}, \texttt{R}) \mid s = L) &= \frac{ 0.1275 \gamma^2 }{ 1 + \gamma + \gamma^2 } \\
\rho(\as = (\texttt{R}, \texttt{L}) \mid s = L) &= \frac{ 0.1275 \gamma^2 }{ 1 + \gamma + \gamma^2 } \\
\rho(\as = (\texttt{R}, \texttt{R}) \mid s = L) &= \frac{ 0.7225 \gamma^2 }{ 1 + \gamma + \gamma^2 }
\end{align}

\subsection{State Values}

The state values of actions \texttt{L} and \texttt{R} are straightforward due to the episode terminating immediately, and are simply the respective rewards,
\begin{align}
\qpolicies(s=\texttt{L}, \as=(\texttt{L}, \texttt{L})) &= -50 \,, \\
\qpolicies(s=\texttt{L}, \as=(\texttt{L}, \texttt{R})) &= -100 \,, \\
\qpolicies(s=\texttt{L}, \as=(\texttt{R}, \texttt{L})) &= -100 \,, \\
\qpolicies(s=\texttt{L}, \as=(\texttt{R}, \texttt{R})) &= 20 \,.
\end{align}
On the other hand, the state value associated with both agents listening is obtained using all of the previously calculated probabilities, and a recursive relation,
\begin{align}
\qpolicies(s=\texttt{L}, \as=(\texttt{Li}, \texttt{Li})) &= -2 + \gamma \sum_{\as'} \rho(\as'\mid s'=\texttt{L}) \qpolicies(s'=\texttt{L}, \as') \nonumber \\
&= -2 
+ \frac{ 0.0225\gamma^3 ( -50 ) }{ 1+\gamma+\gamma^2 }
+ \frac{0.1275\gamma^3 ( -100 ) }{ 1+\gamma+\gamma^2} \nonumber \\
&\phantom{{}={}} + \frac{0.1275\gamma^3 ( -100 ) }{ 1+\gamma+\gamma^2} 
+ \frac{0.7225\gamma^3 ( 20 ) }{ 1+\gamma+\gamma^2} 
\nonumber \\
&\phantom{{}={}} + \frac{ \gamma+\gamma^2 }{ 1+\gamma+\gamma^2 } \qpolicies(s=\texttt{L}, \as=(\texttt{Li}, \texttt{Li})) \nonumber \\
&= -2 -12.175 \frac{ \gamma^3 }{ 1+\gamma+\gamma^2 } \nonumber \\
&\phantom{{}={}} + \frac{ \gamma+\gamma^2 }{ 1+\gamma+\gamma^2 } \qpolicies(s=\texttt{L}, \as=(\texttt{Li}, \texttt{Li})) \,, \\
\intertext{then, fixing $\gamma=1.0$,}
\qpolicies(s=\texttt{L}, \as=(\texttt{Li}, \texttt{Li})) &= -\frac{ 18.175 }{3} + \frac{2}{3} \qpolicies(s=\texttt{L}, \as=(\texttt{Li}, \texttt{Li})) \,, \\
\intertext{which results in}
\qpolicies(s=\texttt{L}, \as=(\texttt{Li}, \texttt{Li})) &= -18.175 \,.
\end{align}

By the symmetries of the Dec-Tiger problem and the given policies, we similarly get
\begin{align}
\qpolicies(s=\texttt{R}, \as=(\texttt{L}, \texttt{L})) &= 20 \,, \\
\qpolicies(s=\texttt{R}, \as=(\texttt{L}, \texttt{R})) &= -100 \,, \\
\qpolicies(s=\texttt{R}, \as=(\texttt{R}, \texttt{L})) &= -100 \,, \\
\qpolicies(s=\texttt{R}, \as=(\texttt{R}, \texttt{R})) &= -50 \,. \\
\qpolicies(s=\texttt{R}, \as=(\texttt{Li}, \texttt{Li})) &= -18.175 \,.
\end{align}

\subsection{History-State Values}

Here, we compute the history-state values $\qpolicies(\hs, s, \as)$ for some relevant combinations of history, state and actions.  
We begin by noting that, because all door-opening combinations lead to the episode terminating, then each respective value is invariant to the given history, and simply equal to the respective reward,
\begin{equation}
\qpolicies(\hs, s, \as) = R(s, \as) \quad\quad \forall\as\in \{ (\texttt{L}, \texttt{L}), (\texttt{L}, \texttt{R}), (\texttt{R}, \texttt{L}), (\texttt{R}, \texttt{R}) \} \,.
\end{equation}

Then, we compute the values associated with the beginning of the episode, i.e., with the empty joint history $\hs=\epsilon$.  Because the second action for both agents will be to listen again, and since the state will not change, we can quickly compute the initial history-state values as,
\begin{align}
&\phantom{{}={}} \qpolicies(\hs=\epsilon, s=\texttt{L}, \as=(\texttt{Li}, \texttt{Li})) \nonumber \\
&= -2 -2\gamma +\gamma^2 \sum_\hs \Pr(\Hs_2=\hs\mid S_2=\texttt{L}) \sum_\as \Pr(\As_2=\as \mid \Hs_2=\hs) \qpolicies(\hs, s=\texttt{L}, \as=\as) \nonumber \\
&= -2 -2\gamma +\gamma^2 \sum_\as \Pr(\As_2=\as\mid S_2=\texttt{L}) R(s=\texttt{L}, \as=\as) \nonumber \\
&= -2 -2\gamma +\gamma^2 \left( 0.0225 \cdot (-50) +0.1275 \cdot (-100) +0.1275 \cdot (-100) + 0.7225 \cdot 20 \right) \nonumber \\
&= -2 -2\gamma -12.175 \gamma^2 \,, \\
\intertext{then, fixing $\gamma=1.0$,}
&= -16.175 \,.
\end{align}

By the symmetries of the Dec-Tiger problem and the given policies, we similarly get
\begin{equation}
\qpolicies(\hs=\epsilon, s=\texttt{R}, \as=(\texttt{Li}, \texttt{Li})) = -16.175 \,.
\end{equation}

\subsection{History Values}

Finally, we use the equivalence from \Cref{thm:identity:qha:qhsa} to compute history values $\qpolicies(\hs, \as)$.  For door-opening actions, we obtain
\begin{align}
\qpolicies(\hs, \as) &= \Exp_{s\mid \hs} \left[ \qpolicies(\hs, s, \as) \right] \,, \nonumber \\
&= \Exp_{s\mid\hs} \left[ R(s, \as) \right] \quad\quad \forall\as\in\{ (\texttt{L}, \texttt{L}), (\texttt{L}, \texttt{R}), (\texttt{R}, \texttt{L}), (\texttt{R}, \texttt{R}) \} \,.
\end{align}
The state probabilities $\Pr(s\mid \hs)$ are computed and shown in \Cref{tab:dectiger:beliefs}, and the respective history values are shown in \Cref{tab:dectiger:qha}.
Finally, we also compute the history value associated with the beginning of the episode, i.e., with the empty joint history $\hs = \epsilon$, for which trivially $\Pr(s\mid \hs=\epsilon) = \frac{1}{2}$,
\begin{align}
\qpolicies(\hs=\epsilon, \as=(\texttt{Li}, \texttt{Li}) &= \frac{1}{2} \sum_{s\in\{ \texttt{L}, \texttt{R} \}} \qpolicies(\hs=\epsilon, s, \as=(\texttt{Li}, \texttt{Li})) \nonumber \\
&= -16.175 \,.
\end{align}

\begin{table}
\centering
\caption{$\Pr(s\mid \hs) \propto \Pr(\hs\mid s) \Pr(s)$, which can be calculated from the values in \Cref{tab:dectiger:pr_hs_s}. In this table, a joint observation $\hs=((A, B), (C, D))$ means that the two agents have received observations $A$ and $B$ on the first timestep, and $C$ and $D$ on the second timestep.}
\label{tab:dectiger:beliefs}
\begin{tabular}{crr}
    $s$ & $L$ & $R$ \\
    \toprule
    $\hs$ \\
    \toprule
    $((\texttt{L}, \texttt{L}), (\texttt{L}, \texttt{L}))$ & 0.999 & 0.001 \\
    $((\texttt{L}, \texttt{L}), (\texttt{L}, \texttt{R}))$ & 0.970 & 0.030 \\
    $((\texttt{L}, \texttt{L}), (\texttt{R}, \texttt{L}))$ & 0.970 & 0.030 \\
    $((\texttt{L}, \texttt{L}), (\texttt{R}, \texttt{R}))$ & 0.500 & 0.500 \\
    $((\texttt{L}, \texttt{R}), (\texttt{L}, \texttt{L}))$ & 0.970 & 0.030 \\
    $((\texttt{L}, \texttt{R}), (\texttt{L}, \texttt{R}))$ & 0.500 & 0.500 \\
    $((\texttt{L}, \texttt{R}), (\texttt{R}, \texttt{L}))$ & 0.500 & 0.500 \\
    $((\texttt{L}, \texttt{R}), (\texttt{R}, \texttt{R}))$ & 0.030 & 0.970 \\
    $((\texttt{R}, \texttt{L}), (\texttt{L}, \texttt{L}))$ & 0.970 & 0.030 \\
    $((\texttt{R}, \texttt{L}), (\texttt{L}, \texttt{R}))$ & 0.500 & 0.500 \\
    $((\texttt{R}, \texttt{L}), (\texttt{R}, \texttt{L}))$ & 0.500 & 0.500 \\
    $((\texttt{R}, \texttt{L}), (\texttt{R}, \texttt{R}))$ & 0.030 & 0.970 \\
    $((\texttt{R}, \texttt{R}), (\texttt{L}, \texttt{L}))$ & 0.500 & 0.500 \\
    $((\texttt{R}, \texttt{R}), (\texttt{L}, \texttt{R}))$ & 0.030 & 0.970 \\
    $((\texttt{R}, \texttt{R}), (\texttt{R}, \texttt{L}))$ & 0.030 & 0.970 \\
    $((\texttt{R}, \texttt{R}), (\texttt{R}, \texttt{R}))$ & 0.001 & 0.999 \\
    \bottomrule
\end{tabular}
\end{table}

\begin{table}
\centering
\caption{$\qpolicies(\hs, \as) = \Exp_{s\mid \hs} \left[ R(s, \as) \right] \quad\quad \forall\as\in\{ (\texttt{L}, \texttt{L}), (\texttt{L}, \texttt{R}), (\texttt{R}, \texttt{L}), (\texttt{R}, \texttt{R}) \}$. In this table, a joint observation $\hs=((A, B), (C, D))$ means that the two agents have received observations $A$ and $B$ on the first timestep, and $C$ and $D$ on the second timestep.}
\label{tab:dectiger:qha}
\begin{tabular}{crrrr}
    $\as$ & $(\texttt{L}, \texttt{L})$ & $(\texttt{L}, \texttt{R})$ & $(\texttt{R}, \texttt{L})$ & $(\texttt{R}, \texttt{R})$ \\
    \toprule
    $\hs$ \\
    \toprule
    $((\texttt{L}, \texttt{L}), (\texttt{L}, \texttt{L}))$ & -49.93 & -100.00 & -100.00 &  19.93 \\
    $((\texttt{L}, \texttt{L}), (\texttt{L}, \texttt{R}))$ & -47.89 & -100.00 & -100.00 &  17.89 \\
    $((\texttt{L}, \texttt{L}), (\texttt{R}, \texttt{L}))$ & -47.89 & -100.00 & -100.00 &  17.89 \\
    $((\texttt{L}, \texttt{L}), (\texttt{R}, \texttt{R}))$ & -15.00 & -100.00 & -100.00 & -15.00 \\
    $((\texttt{L}, \texttt{R}), (\texttt{L}, \texttt{L}))$ & -47.89 & -100.00 & -100.00 &  17.89 \\
    $((\texttt{L}, \texttt{R}), (\texttt{L}, \texttt{R}))$ & -15.00 & -100.00 & -100.00 & -15.00 \\
    $((\texttt{L}, \texttt{R}), (\texttt{R}, \texttt{L}))$ & -15.00 & -100.00 & -100.00 & -15.00 \\
    $((\texttt{L}, \texttt{R}), (\texttt{R}, \texttt{R}))$ &  17.89 & -100.00 & -100.00 & -47.89 \\
    $((\texttt{R}, \texttt{L}), (\texttt{L}, \texttt{L}))$ & -47.89 & -100.00 & -100.00 &  17.89 \\
    $((\texttt{R}, \texttt{L}), (\texttt{L}, \texttt{R}))$ & -15.00 & -100.00 & -100.00 & -15.00 \\
    $((\texttt{R}, \texttt{L}), (\texttt{R}, \texttt{L}))$ & -15.00 & -100.00 & -100.00 & -15.00 \\
    $((\texttt{R}, \texttt{L}), (\texttt{R}, \texttt{R}))$ &  17.89 & -100.00 & -100.00 & -47.89 \\
    $((\texttt{R}, \texttt{R}), (\texttt{L}, \texttt{L}))$ & -15.00 & -100.00 & -100.00 & -15.00 \\
    $((\texttt{R}, \texttt{R}), (\texttt{L}, \texttt{R}))$ &  17.89 & -100.00 & -100.00 & -47.89 \\
    $((\texttt{R}, \texttt{R}), (\texttt{R}, \texttt{L}))$ &  17.89 & -100.00 & -100.00 & -47.89 \\
    $((\texttt{R}, \texttt{R}), (\texttt{R}, \texttt{R}))$ &  19.93 & -100.00 & -100.00 & -49.93 \\
    \bottomrule
\end{tabular}
\end{table}

\clearpage
\section{Pseudocode}
\label{pseudocode}

We note that practical implementations of Actor-Critic methods do not model the Q-values $\qpolicy$ directly using a Q-model $\qmodel$, but rather using a bootstrapped V-model $\vmodel$, e.g., using 1-step returns $\qmodel_t \approx r_t + \gamma \vmodel_{t+1}$ (other generalizations like n-step returns or lambda-returns are also common).  Hence, although the theory of Actor-Critic is based on Q-values and models, the pseudocode shown in this section employs V-values and models.  

\begin{algorithm}
\caption{IAC}
\begin{algorithmic}[1]
\label{pseudocode:iac}
\REQUIRE Individual actor models $\policy_i(a; h)$, parameterized by $\theta_i$
\REQUIRE Individual critic models $\vmodel_i(h)$, parameterized by $\vartheta_i$
\LOOP
    \STATE Sample environment episode $(\os_0, \as_0, r_0, \os_1, \as_1, r_1, \ldots, \os_{T-1}, \as_{T-1}, r_{T-1})$
    \STATE Denote individual observed histories as $h_{i,t} \gets (o_{i,0}, a_{i,0}, \ldots o_{i,t-1}, a_{i,t-1}, o_{i,t}) \,, \forall i, t$
    \FOR{each agent $i$}
        \STATE Compute individual advantage estimates $\delta_{i,t} \gets r_t + \gamma \vmodel_i(h_{i,t+1}) - \vmodel_i(h_{i,t}) \,, \forall t$
        \STATE Compute actor gradient estimate $\sum_{t=0}^{T-1} \gamma^t \delta_{i,t} \nabla\log\policy_i(a_{i,t}; h_{i,t})$
        \STATE Update actor parameters $\theta_i$ using gradient estimate
        \STATE Compute critic gradient estimate $\frac{1}{T} \sum_{t=0}^{T-1} \delta_{i,t} \nabla \vmodel_i(h_{i,t})$
        \STATE Update critic parameters $\vartheta_i$ using gradient estimate
    \ENDFOR
\ENDLOOP
\end{algorithmic}
\end{algorithm}

\begin{algorithm}
\caption{IACC-H}
\begin{algorithmic}[1]
\label{pseudocode:iacch}
\REQUIRE Individual actor models $\policy_i(a; h)$, parameterized by $\theta_i$
\REQUIRE Centralized critic model $\vmodel(\hs)$, parameterized by $\vartheta$
\LOOP
    \STATE Sample environment episode $(\os_0, \as_0, r_0, \os_1, \as_1, r_1, \ldots, \os_{T-1}, \as_{T-1}, r_{T-1})$
    \STATE Denote joint observed histories as $\hs_t \gets (\os_0, \as_0, \ldots \os_{t-1}, \as_{t-1}, \os_{t}) \,, \forall t$
    \STATE Denote individual observed histories as $h_{i,t} \gets (o_{i,0}, a_{i,0}, \ldots o_{i,t-1}, a_{i,t-1}, o_{i,t}) \,, \forall i, t$
    \STATE Compute centralized advantage estimates $\delta_t \gets r_t + \gamma \vmodel(\hs_{t+1}) - \vmodel(\hs_t) \,, \forall t$
    \FOR{each agent $i$}
        \STATE Compute actor gradient estimate $\sum_{t=0}^{T-1} \gamma^t \delta_t \nabla\log\policy_i(a_{i,t}; h_{i,t})$
        \STATE Update actor parameters $\theta_i$ using gradient estimate
    \ENDFOR
    \STATE Compute critic gradient estimate $\frac{1}{T} \sum_{t=0}^{T-1} \delta_t \nabla \vmodel(\hs_t)$
    \STATE Update critic parameters $\vartheta$ using gradient estimate
\ENDLOOP
\end{algorithmic}
\end{algorithm}

\begin{algorithm}
\caption{IACC-S}
\begin{algorithmic}[1]
\label{pseudocode:iaccs}
\REQUIRE Individual actor models $\policy_i(a; h)$, parameterized by $\theta_i$
\REQUIRE Centralized critic model $\vmodel(s)$, parameterized by $\vartheta$
\LOOP
    \STATE Sample environment episode $(s_0, \os_0, \as_0, r_0, s_1, \os_1, \as_1, r_1, \ldots, s_{T-1}, \os_{T-1}, \as_{T-1}, r_{T-1})$
    \STATE Denote individual observed histories as $h_{i,t} \gets (o_{i,0}, a_{i,0}, \ldots o_{i,t-1}, a_{i,t-1}, o_{i,t}) \,, \forall i, t$
    \STATE Compute centralized advantage estimates $\delta_t \gets r_t + \gamma \vmodel(s_{t+1}) - \vmodel(s_t) \,, \forall t$
    \FOR{each agent $i$}
        \STATE Compute actor gradient estimate $\sum_{t=0}^{T-1} \gamma^t \delta_t \nabla\log\policy_i(a_{i,t}; h_{i,t})$
        \STATE Update actor parameters $\theta_i$ using gradient estimate
    \ENDFOR
    \STATE Compute critic gradient estimate $\frac{1}{T} \sum_{t=0}^{T-1} \delta_t \nabla \vmodel(s_t)$
    \STATE Update critic parameters $\vartheta$ using gradient estimate
\ENDLOOP
\end{algorithmic}
\end{algorithm}

\begin{algorithm}
\caption{IACC-HS}
\begin{algorithmic}[1]
\label{pseudocode:iacchs}
\REQUIRE Individual actor models $\policy_i(a; h)$, parameterized by $\theta_i$
\REQUIRE Centralized critic model $\vmodel(\hs, s)$, parameterized by $\vartheta$
\LOOP
    \STATE Sample environment episode $(s_0, \os_0, \as_0, r_0, s_1, \os_1, \as_1, r_1, \ldots, s_{T-1}, \os_{T-1}, \as_{T-1}, r_{T-1})$
    \STATE Denote joint observed histories as $\hs_t \gets (\os_0, \as_0, \ldots \os_{t-1}, \as_{t-1}, \os_{t}) \,, \forall t$
    \STATE Denote individual observed histories as $h_{i,t} \gets (o_{i,0}, a_{i,0}, \ldots o_{i,t-1}, a_{i,t-1}, o_{i,t}) \,, \forall i, t$
    \STATE Compute centralized advantage estimates $\delta_t \gets r_t + \gamma \vmodel(\hs_{t+1}, s_{t+1}) - \vmodel(\hs_t, s_t) \,, \forall t$
    \FOR{each agent $i$}
        \STATE Compute actor gradient estimate $\sum_{t=0}^{T-1} \gamma^t \delta_t \nabla\log\policy_i(a_{i,t}; h_{i,t})$
        \STATE Update actor parameters $\theta_i$ using gradient estimate
    \ENDFOR
    \STATE Compute critic gradient estimate $\frac{1}{T} \sum_{t=0}^{T-1} \delta_t \nabla \vmodel(\hs_t, s_t)$
    \STATE Update critic parameters $\vartheta$ using gradient estimate
\ENDLOOP
\end{algorithmic}
\end{algorithm}

\section{Convergence Properties of IAC and IACC Variants}
\label{sec:convergence}

In this section, we analyze the convergence properties of basic gradient-based optimization algorithms when using the expected gradients $\nablai J$, $\grad_\hs$, and $\grad_{\hs,s}$, and the sample gradients $\samplegrad_i$, $\samplegrad_\hs$, and $\samplegrad_{\hs,s}$.
We first fully derive \Cref{eq:gradient:qiha,eq:gradient:qha,eq:gradient:qhsa} as the gradient of the episodic return objective $J$, which in itself serves as an auxiliary proof that the expected gradients $\nablai J = \grad_{\hs} = \grad_{\hs,s}$ are equal to each other.
Then, we consider vanilla gradient descent based on the expected gradients and show that, under mild conditions, it converges to the same local optimum when using either type of gradient from the same starting conditions.
Finally, we consider stochastic gradient descent based on the sample gradients and show that, under the same mild conditions, it still converges to a local optimum when using any of the sample gradients from the same starting conditions, although the local optimum may differ depending on which sample gradient is used.

\paragraph{Notation.}
We employ $t$ ad $k$ to denote time indices, $\tau = (s_0, \os_0, \as_0, s_1, \os_1, \as_1, \ldots)$ to denote a full episodic trajectory (including all states, observations, and actions), $x_{i,t}$ to denote a variable associated with agent $i$ at timestep $t$, and $x_{(t:)}$ to denote a subsequence of variables starting from timestep $t$ (included) onwards.

\subsection{Derivations of Gradients}
\label{sec:derivation}

In this section, we show the derivations of \Cref{eq:gradient:qiha,eq:gradient:qha,eq:gradient:qhsa}, for which it is necessary to consider the following lemma.

\begin{lemma}\label{thm:lemma}
  For timesteps $t<k$, $\Exp\left[ \nablai\log\policy_i(a_{i,k}; h_{i,k}) R(s_t, \as_t) \right] = 0$.
\end{lemma}

\begin{proof}
  \begin{align}
    & \Exp\left[ \nablai\log\policy_i(a_{i,k}; h_{i,k}) R(s_t, \as_t) \right] \nonumber \\
    &= \Exp_{h_{i,k}, a_{i,k}, s_t, \as_t} \left[ \nablai\log\policy_i(a_{i,k}; h_{i,k}) R(s_t, \as_t) \right] \nonumber \\
    &= \Exp_{h_{i,k}, s_t, \as_t} \left[ \Exp_{a_{i,k}\mid h_{i,k}, s_t, \as_t} \left[ \nablai\log\policy_i(a_{i,k}; h_{i,k}) \right] R(s_t, \as_t) \right] \nonumber \\
    &\phantom{{}={}} \text{(conditional independence valid when $t<k$)} \nonumber \\
    &= \Exp_{h_{i,k}, s_t, \as_t} \left[ \Exp_{a_{i,k}\mid h_{i,k}} \left[ \nablai\log\policy_i(a_{i,k}; h_{i,k}) \right] R(s_t, \as_t) \right] \nonumber \\
    &\phantom{{}={}} \text{(expectation of score function is zero)} \nonumber \\
    &= 0 \,.
  \end{align}
\end{proof}


\begin{theorem}[IAC Policy Gradient]
\label{thm:pg:iac}
  The gradient of $J = \Exp\left[ \sum_t R(s_t, \as_t) \right]$ with respect to agent $i$'s parameters $\theta_i$ is
  \begin{align}
    \nablai J &= \Exp\left[ \sum_t \gamma^t \qpolicies_i(h_{i,t}, a_{i,t})
    \nablai\log\policy(a_{i,t}; h_{i,t}) \right] \nonumber \\
    &= (1-\gamma) \Exp_{\hs,\as\sim \rho(\hs, \as)} \left[ \qpolicies_i(h_i, a_i) \nablai \log\policy_i(a_i; h_i, \theta_i) \right] \,.
  \end{align}
\end{theorem}

\begin{proof}
  \begingroup
  \allowdisplaybreaks
  \begin{align}
    \nablai J
    &= \nablai \Exp_\tau \left[ \sum_t \gamma^t R(s_t, \as_t) \right] \nonumber \\
    %
    &= \Exp_\tau \left[ \nablai\log\Pr(\tau) \sum_t \gamma^t R(s_t, \as_t) \right] \nonumber \\
    &\phantom{{}={}} \text{(note two independent time indices $k$ and $t$)} \nonumber \\
    &= \Exp_\tau \left[ \sum_k \nablai\log\policy_i(a_{i,k}; h_{i,k}) \sum_t \gamma^t R(s_t, \as_t) \right] \nonumber \\
    &\phantom{{}={}} \text{(see \Cref{thm:lemma})} \nonumber \\
    &= \sum_{k,t} \gamma^t \Exp_\tau \left[ \nablai\log\policy_i(a_{i,k}; h_{i,k}) R(s_t, \as_t) \right] \nonumber \\
    &= \cancel{ \sum_{k,t<k} \gamma^t \Exp_\tau \left[ \nablai\log\policy_i(a_{i,k}; h_{i,k}) R(s_t, \as_t) \right] } + \sum_{k,t\ge k} \gamma^t \Exp_\tau \left[ \nablai\log\policy_i(a_{i,k}; h_{i,k}) R(s_t, \as_t) \right] \nonumber \\
    &= \sum_{k,t\ge k} \gamma^k \gamma^{t-k} \Exp_{\tau} \left[ \nablai\log\policy_i(a_{i,k}; h_{i,k}) R(s_t, \as_t) \right] \nonumber \\
    &= \sum_k \gamma^k \Exp_\tau \left[ \nablai\log\policy_i(a_{i,k}; h_{i,k}) \sum_{t\ge k} \gamma^{t-k} R(s_t, \as_t) \right] \nonumber \\
    &= \sum_k \gamma^k \Exp_{h_{i,k}, a_{i,k}, s_{(k:)}, \as_{(k:)}} \left[ \nablai\log\policy_i(a_{i,k}; h_{i,k}) \sum_{t\ge k} \gamma^{t-k} R(s_t, \as_t) \right] \nonumber \\
    &= \sum_k \gamma^k \Exp_{h_{i,k}, a_{i,k}} \left[ \nablai\log\policy_i(a_{i,k}; h_{i,k}) \Exp_{s_{(k:)}, \as_{(k:)}\mid h_{i,k}, a_{i,k}} \left[ \sum_{t\ge k} \gamma^{t-k} R(s_t, \as_t) \right] \right] \nonumber \\
    &= \sum_k \gamma^k \Exp_{h_{i,k}, a_{i,k}} \left[ \nablai\log\policy_i(a_{i,k}; h_{i,k}) \qpolicies_i(h_{i,k}, a_{i,k}) \right] \nonumber \\
    &= \sum_k \gamma^k \Exp_\tau \left[ \nablai\log\policy_i(a_{i,k}; h_{i,k}) \qpolicies_i(h_{i,k}, a_{i,k}) \right] \nonumber \\
    &= \Exp\left[ \sum_k \gamma^k \nablai\log\policy_i(a_{i,k}; h_{i,k}) \qpolicies_i(h_{i,k}, a_{i,k}) \right] \nonumber \\
    &\phantom{{}={}} \text{(change of variable $k \to t$)} \nonumber \\
    &= \Exp\left[ \sum_t \gamma^t \qpolicies_i(h_{i,t}, a_{i,t}) \nablai\log\policy_i(a_{i,t}; h_{i,t}) \right] \nonumber \\
    &= (1-\gamma) \Exp_{\hs,\as\sim\rho(\hs,\as)}\left[ \qpolicies_i(h_i, a_i) \nablai\log\policy_i(a_i; h_i) \right] \,.
  \end{align}
  \endgroup
\end{proof}


\begin{theorem}[IACC-H Policy Gradient]
\label{thm:pg:iacc-h}
  The gradient of $J = \Exp\left[ \sum_t R(s_t, \as_t) \right]$ with respect to agent $i$'s parameters $\theta_i$ is
  \begin{align}
    \grad_\hs &= \Exp\left[ \sum_t \gamma^t \qpolicies(\hs_t, \as_t)
    \nablai\log\policy(a_{i,t}; h_{i,t}) \right] \nonumber \\
    &= (1-\gamma) \Exp_{\hs,\as\sim \rho(\hs, \as)} \left[ \qpolicies(\hs, \as) \nablai \log\policy_i(a_i; h_i, \theta_i) \right] \,.
  \end{align}
\end{theorem}

\begin{proof}
  \begingroup
  \allowdisplaybreaks
  \begin{align}
    \grad_\hs
    &= \nablai \Exp_\tau \left[ \sum_t \gamma^t R(s_t, \as_t) \right] \nonumber \\
    %
    &= \Exp_\tau \left[ \nablai\log\Pr(\tau) \sum_t \gamma^t R(s_t, \as_t) \right] \nonumber \\
    &\phantom{{}={}} \text{(note two independent time indices $k$ and $t$)} \nonumber \\
    &= \Exp_\tau \left[ \sum_k \nablai\log\policy_i(a_{i,k}; h_{i,k}) \sum_t \gamma^t R(s_t, \as_t) \right] \nonumber \\
    &\phantom{{}={}} \text{(see \Cref{thm:lemma})} \nonumber \\
    &= \sum_{k,t} \gamma^t \Exp_\tau \left[ \nablai\log\policy_i(a_{i,k}; h_{i,k}) R(s_t, \as_t) \right] \nonumber \\
    &= \cancel{ \sum_{k,t<k} \gamma^t \Exp_\tau \left[ \nablai\log\policy_i(a_{i,k}; h_{i,k}) R(s_t, \as_t) \right] } + \sum_{k,t\ge k} \gamma^t \Exp_\tau \left[ \nablai\log\policy_i(a_{i,k}; h_{i,k}) R(s_t, \as_t) \right] \nonumber \\
    &= \sum_{k,t\ge k} \gamma^k \gamma^{t-k} \Exp_{\tau} \left[ \nablai\log\policy_i(a_{i,k}; h_{i,k}) R(s_t, \as_t) \right] \nonumber \\
    &= \sum_k \gamma^k \Exp_\tau \left[ \nablai\log\policy_i(a_{i,k}; h_{i,k}) \sum_{t\ge k} \gamma^{t-k} R(s_t, \as_t) \right] \nonumber \\
    &= \sum_k \gamma^k \Exp_{\hs_k, \as_k, s_{(k:)}, \as_{(k:)}} \left[ \nablai\log\policy_i(a_{i,k}; h_{i,k}) \sum_{t\ge k} \gamma^{t-k} R(s_t, \as_t) \right] \nonumber \\
    &= \sum_k \gamma^k \Exp_{\hs_k, \as_k} \left[ \nablai\log\policy_i(a_{i,k}; h_{i,k}) \Exp_{s_{(k:)}, \as_{(k:)}\mid \hs_k, \as_k} \left[ \sum_{t\ge k} \gamma^{t-k} R(s_t, \as_t) \right] \right] \nonumber \\
    &= \sum_k \gamma^k \Exp_{\hs_k, \as_k} \left[ \nablai\log\policy_i(a_{i,k}; h_{i,k}) \qpolicies(\hs_k, \as_k) \right] \nonumber \\
    &= \sum_k \gamma^k \Exp_\tau \left[ \nablai\log\policy_i(a_{i,k}; h_{i,k}) \qpolicies(\hs_k, \as_k) \right] \nonumber \\
    &= \Exp\left[ \sum_k \gamma^k \nablai\log\policy_i(a_{i,k}; h_{i,k}) \qpolicies(\hs_k, \as_k) \right] \nonumber \\
    &\phantom{{}={}} \text{(change of variable $k \to t$)} \nonumber \\
    &= \Exp\left[ \sum_t \gamma^t \qpolicies(\hs_t, \as_t) \nablai\log\policy_i(a_{i,t}; h_{i,t}) \right] \nonumber \\
    &= (1-\gamma) \Exp_{\hs,\as\sim\rho(\hs,\as)}\left[ \qpolicies(\hs, \as) \nablai\log\policy_i(a_i; h_i) \right] \,.
  \end{align}
  \endgroup
\end{proof}


\begin{theorem}[IACC-HS Policy Gradient]
\label{thm:pg:iacc-hs}
  The gradient of $J = \Exp\left[ \sum_t R(s_t, \as_t) \right]$ with respect to agent $i$'s parameters $\theta_i$ is
  \begin{align}
    \grad_{\hs,s} &= \Exp\left[ \sum_t \gamma^t \qpolicies(\hs_t, s_t, \as_t)
    \nablai\log\policy(a_{i,t}; h_{i,t}) \right] \nonumber \\
    &= (1-\gamma) \Exp_{\hs, s, \as\sim \rho(\hs, s, \as)} \left[ \qpolicies(\hs, s, \as) \nablai \log\policy_i(a_i; h_i, \theta_i) \right] \,.
  \end{align}
\end{theorem}

\begin{proof}
  \begingroup
  \allowdisplaybreaks
  \begin{align}
    \grad_{\hs,s}
    &= \nablai \Exp_\tau \left[ \sum_t \gamma^t R(s_t, \as_t) \right] \nonumber \\
    %
    &= \Exp_\tau \left[ \nablai\log\Pr(\tau) \sum_t \gamma^t R(s_t, \as_t) \right] \nonumber \\
    &\phantom{{}={}} \text{(note two independent time indices $k$ and $t$)} \nonumber \\
    &= \Exp_\tau \left[ \sum_k \nablai\log\policy_i(a_{i,k}; h_{i,k}) \sum_t \gamma^t R(s_t, \as_t) \right] \nonumber \\
    &\phantom{{}={}} \text{(see \Cref{thm:lemma})} \nonumber \\
    &= \sum_{k,t} \gamma^t \Exp_\tau \left[ \nablai\log\policy_i(a_{i,k}; h_{i,k}) R(s_t, \as_t) \right] \nonumber \\
    &= \cancel{ \sum_{k,t<k} \gamma^t \Exp_\tau \left[ \nablai\log\policy_i(a_{i,k}; h_{i,k}) R(s_t, \as_t) \right] } + \sum_{k,t\ge k} \gamma^t \Exp_\tau \left[ \nablai\log\policy_i(a_{i,k}; h_{i,k}) R(s_t, \as_t) \right] \nonumber \\
    &= \sum_{k,t\ge k} \gamma^k \gamma^{t-k} \Exp_{\tau} \left[ \nablai\log\policy_i(a_{i,k}; h_{i,k}) R(s_t, \as_t) \right] \nonumber \\
    &= \sum_k \gamma^k \Exp_\tau \left[ \nablai\log\policy_i(a_{i,k}; h_{i,k}) \sum_{t\ge k} \gamma^{t-k} R(s_t, \as_t) \right] \nonumber \\
    &= \sum_k \gamma^k \Exp_{\hs_k, s_k, \as_k, s_{(k:)}, \as_{(k:)}} \left[ \nablai\log\policy_i(a_{i,k}; h_{i,k}) \sum_{t\ge k} \gamma^{t-k} R(s_t, \as_t) \right] \nonumber \\
    &= \sum_k \gamma^k \Exp_{\hs_k, s_k, \as_k} \left[ \nablai\log\policy_i(a_{i,k}; h_{i,k}) \Exp_{s_{(k:)}, \as_{(k:)}\mid \hs_k, s_k, \as_k} \left[ \sum_{t\ge k} \gamma^{t-k} R(s_t, \as_t) \right] \right] \nonumber \\
    &= \sum_k \gamma^k \Exp_{\hs_k, s_k, \as_k} \left[ \nablai\log\policy_i(a_{i,k}; h_{i,k}) \qpolicies(\hs_k, s_k, \as_k) \right] \nonumber \\
    &= \sum_k \gamma^k \Exp_\tau \left[ \nablai\log\policy_i(a_{i,k}; h_{i,k}) \qpolicies(\hs_k, s_k, \as_k) \right] \nonumber \\
    &= \Exp\left[ \sum_k \gamma^k \nablai\log\policy_i(a_{i,k}; h_{i,k}) \qpolicies(\hs_k, s_k, \as_k) \right] \nonumber \\
    &\phantom{{}={}} \text{(change of variable $k \to t$)} \nonumber \\
    &= \Exp\left[ \sum_t \gamma^t \qpolicies(\hs_t, s_t, \as_t) \nablai\log\policy_i(a_{i,t}; h_{i,t}) \right] \nonumber \\
    &= (1-\gamma) \Exp_{\hs,s,\as\sim\rho(\hs,s,\as)}\left[ \qpolicies(\hs, s, \as) \nablai\log\policy_i(a_i; h_i) \right] \,.
  \end{align}
  \endgroup
\end{proof}

We note that, though \Cref{thm:pg:iac,thm:pg:iacc-h,thm:pg:iacc-hs} are framed in the infinite-horizon control setting, they are equally applicable to the finite-horizon and the indefinite-horizon (a.k.a. episodic) control settings due to
\begin{enumerate*}[label=(\alph*)]
    \item history policies and history values being intrinsically timed policies and values, and
    \item being able to reframe finite-horizon and indefinite-horizon control problems into the infinite-horizon case via sink states.
\end{enumerate*}

\subsection{Convergence Properties}

In the rest of this section, we address the convergence properties of gradient descent algorithms that employ either the expected or the sample versions of the IAC, IAC-H, and IAC-HS gradients, and show that they converge to local optima in policy space under the following assumptions:
%
%
\begin{itemize}
\item Independently parameterized policy models $\policy_i$ with no parameter sharing, i.e., parameters $\theta_i$ exclusively determine the behavior of policy $\policy_i$ and do not affect the behavior of other policies $\policy_j$ where $j\neq i$ (this assumption is not strictly necessary given appropriate adjustments to the optimization algorithms, though it does simplify the discussion).
\item Parameterized policy models $\policy_i$ that are smooth and differetiable everywhere.
\item Identical starting conditions and parameter initialization.
\item A synchronous optimization setting, i.e., all policy gradients are computed/estimated first (using the same sample data, if applicable), and then updated at the same time.
\end{itemize}
For stochastic gradient descent based on sample gradients, we make the following additional assumption:
\begin{itemize}
\item For stochastic gradient descent, properly seeded random processes to avoid stochasticity being the cause of differing convergence values.
\end{itemize}

\Cref{pseudocode:gradient-descent,pseudocode:stochastic-gradient-descent} represent the optimization processes associated with the expected and sample gradients, respectively. Note the synchronous optimization process that strictly alternates between a gradient computation/estimation phase, and a gradient update phase.
The synchronous optimization assumption is necessary to avoid non-stationary issues that may arise if the agents were to update their own policies at different and/or non-predetermined times. With that assumption in place, we can view the optimization processes of all agents as a single optimization process over the joint parameter space of all agents (as also discussed in \cite{peshkin2000learning}), which is sufficient to associate to the policy gradient algorithms the same convergence guarantees of (stochastic) gradient descent when used to optimize a generic non-convex function.

The (stochastic) gradient descent applied to a generic non-convex function does not have the same strict theoretical convergence guarantees as when applied to a generic convex function due to the existence of multiple stationary points (saddle points, local maxima, and local minima)~\cite{NIPS2014_17e23e50,pmlr-v38-choromanska15,pmlr-v40-Ge15}.
Despite this notable theoretical difference between the convex and non-convex case, in practice, (stochastic) gradient descent is often able to avoid saddle points and to find sufficiently performative local minima~\cite{pmlr-v38-choromanska15,Goodfellow-et-al-2016,robbins1951stochastic,kiefer1952stochastic}.


Finally, we note that gradient descent with expected gradients (\Cref{pseudocode:gradient-descent}) will naturally result in convergence to the same final policies for all gradient types, provided identical starting conditions and parameter initialization, simply due to the nature of all gradients being equal in expectation.
On the other hand, stochastic gradient descent with sample gradients (\Cref{pseudocode:stochastic-gradient-descent}) may result in convergence to different final policies for each gradient type, even provided identical starting conditions and parameter initialization.
Although the sample gradients $\samplegrad_i$, $\samplegrad_\hs$, and $\samplegrad_{\hs,s}$ are all unbiased and identical in expectation, each has a different variance profile, and the respective values $\qpolicies_i(h_i, a_i)$, $\qpolicies(\hs, \as)$, and $\qpolicies(\hs, s, \as)$ intrinsically integrate (or not) over different variables according to \Cref{thm:identity:qiha:qha,thm:identity:qha:qhsa}.

\subsection{Related Work on the Convergence of Actor-Critic Methods}

Previous work on the convergence of actor-critic methods have shown convergence in both single-agent and multi-agent settings under the average-reward setting, as opposed to the finite-horizon or episodic setting;
in addition, most works assume fully observability, which is considerably simpler compared to partial observability.

For the single agent case, early work~\cite{sutton2000policy,konda2000actor} showed that under average-reward setting, policy gradient methods converge to a local optimum with function approximation and linear models under mild assumptions.
However, convergence is not shown for the finite-horizon or episodic case, non-linear models, and partial observability.
More recent work~\cite{wu2020finite,tian2024convergence} extend the convergence result of actor-critic methods to non-linear deep neural networks (under some common mild assumptions) with a single training time scale (where critic and actor models are trained with the same frequency), whereas previous works assume two separate training time scales (with the critic moel being trained more frequently than the actor model).
Again, their convergence results also ignore the episodic setting and assume full observability.
\citeauthor{shen2023towards}~\citeyear{shen2023towards} established a finite-time local convergence of A3C for the general policy approximation and global convergence for softmax policy parameterization under i.i.d. sampling; they assume critics with bounded error, linear features, Lipschitz continuous and bounded policy gradient, as well as irreducible and aperiodic Markov chains under the average-reward setting.
Once again, the partially observable setting is not yet examined in this work.
In summary, to the best of our knowledge, the single-agent AC convergence results apply only in the fully observable MDP setting and are focused on average-reward formulation.
Those fully observable single-agent results do not extend directly into our problem formulation since we cannot leverage a stationary state distribution since we use action-observation histories rather than states, and histories cannot be revisited.
To derive the history distribution and the state distribution under partial observability, we give the solution in~\Cref{appendix_solutions} where we formally define the state distribution for Dec-POMDPs.

As for the multi-agent literature, \citeauthor{arrow1960stability}~\citeyear{arrow1960stability} show that for non-cooperative $n$-person games, under certain convexity assumptions on the shape of payoff functions, the gradient-descent process converges to an equilibrium point. This work provides theoretical insight, but does not consider the reinforcement learning setting.
\citeauthor{peshkin2000learning}~\citeyear{peshkin2000learning} imply that a gradient descent policy search algorithm converges to a local optimum with certain policy classes; however, they did not provide an accompanying proof of convergence in detail.
It should be noted that \citeauthor{peshkin2000learning} do assume partial observability and history-based local policies (through finite-state controllers), which is in line with our Dec-POMDP framework.
In this sense, our derivation of convergence follows directly after their work, by looking at the Actor-Critic framework specifically and providing the convergence result in greater detail.

More recent MARL research efforts propose IACC methods with a factorized centralized critic with convergence guarantees, such as DOP~\cite{wang2020dop}.
It is shown that DOP is able to achieve policy improvement with their linear value decomposition architecture; however, they do not explicitly show convergence, though it is implied by the policy improvement result.
We thus provide a convergence result for a set of generic Actor-Critic methods in greater detail.

In summary, convergence in the setting of average reward was regularly investigated in the literature of single-agent control; however, finite-horizon and episodic setting are often glossed over.
In the multi-agent literature,  convergence of policy gradient is often assumed or implied but not shown.
Our work filled the gap of convergence properties of actor-critic methods in both multi-agent finite-horizon and episodic settings, extending convergence results to generic policies. 

\begin{algorithm}[t]
\caption{Gradient Descent}
\begin{algorithmic}[1]
\label{pseudocode:gradient-descent}
\REQUIRE Individual actor models $\policy_i(a; h)$, parameterized by $\theta_i$
\REQUIRE Gradient function $g_i(\theta_i)$ for agent $i$ ($\nablai J$, $\grad_\hs$, or $\grad_{\hs,s}$)
\REQUIRE Stepsize schedule $\eta_k$
\FOR{$k \gets 1, 2, 3, \ldots$}
    \FOR{each agent $i$}
        \STATE $d\theta_i \gets g_i(\theta_i)$
        \hfill\COMMENT{Compute expected gradient}
    \ENDFOR
    \FOR{each agent $i$}
        \STATE $\theta_i \gets \theta_i - \eta_k \cdot d\theta_i$
        \hfill\COMMENT{Update parameters}
    \ENDFOR
\ENDFOR
\end{algorithmic}
\end{algorithm}

\begin{algorithm}[t]
\caption{Stochastic Gradient Descent}
\begin{algorithmic}[1]
\label{pseudocode:stochastic-gradient-descent}
\REQUIRE Individual actor models $\policy_i(a; h)$, parameterized by $\theta_i$
\REQUIRE Sample gradient function $g_i(\theta_i; \tau)$ for agent $i$ ($\samplegrad_i$, $\samplegrad_\hs$, or $\samplegrad_{\hs,s}$)
\REQUIRE Stepsize schedule $\eta_k$
\FOR{$k \gets 1, 2, 3, \ldots$}
    \STATE $\tau \gets (s_0, \os_0, \as_0, r_0, s_1, \os_1, \as_1, r_1, \ldots, s_{T-1}, \os_{T-1}, \as_{T-1}, r_{T-1})$
    \hfill\COMMENT{Sample episode}
    \FOR{each agent $i$}
        \STATE $d\theta_i \gets g_i(\theta_i; \tau)$
        \hfill\COMMENT{Compute sample gradient}
    \ENDFOR
    \FOR{each agent $i$}
        \STATE $\theta_i \gets \theta_i - \eta_k \cdot d\theta_i$
        \hfill\COMMENT{Update parameters}
    \ENDFOR
\ENDFOR
\end{algorithmic}
\end{algorithm}

\clearpage
\bibliographystyle{theapa}
\bibliography{sample}
    
\end{document}